\documentclass[11pt]{article}

\usepackage[a4paper,margin=1in]{geometry}
\usepackage{times}
\usepackage{microtype}
\usepackage{authblk}

\usepackage[numbers,sort&compress]{natbib}
\usepackage{graphicx}
\usepackage{booktabs}
\usepackage{subcaption}

\usepackage[accsupp]{axessibility}  

\usepackage{graphicx}
\graphicspath{{images/}}

\usepackage{hyperref}

\usepackage{orcidlink}

\usepackage{acronym}
\acrodef{IMSLP}{International Music Score Library Project}

\usepackage{makecell} 
\usepackage{tablefootnote} 
\usepackage{multirow}
\usepackage[dvipsnames,table]{xcolor}

\newcommand{\musvit}{\textsc{MuSViT}}
\newcommand{\musvitl}{\textsc{MuSViT}$_{\text{Light}}$}

\newcommand{\Capitan}{\emph{Capitan}}
\newcommand{\Guatemala}{\emph{Guatemala}}
\newcommand{\FMT}{\emph{FMT}}
\newcommand{\ILS}{\emph{Il Lauro Secco}}
\newcommand{\Malaga}{\emph{AMDC}}
\newcommand{\Mozarteum}{\emph{Mozarteum}}
\newcommand{\Polish}{\emph{Polish Digital Scores}}
\newcommand{\FreeScores}{\emph{FreeScores}}	
\newcommand{\PlayIt}{\emph{Can I Play It?}}
\newcommand{\PianoStreet}{\emph{PianoStreet}}	
\newcommand{\DeepScores}{\emph{DeepScoresV2}}

\newcommand{\Dedit}{\ensuremath{D^{\text{edit}}}}
\newcommand{\Dhist}{\ensuremath{D^{\text{his}}}}
\newcommand{\Demb}{\ensuremath{D^{\text{emb}}}}
\newcommand{\PearsonED}{\ensuremath{\rho_p^{\text{ed}}}}
\newcommand{\PearsonHist}{\ensuremath{\rho_p^{\text{h}}}}
\newcommand{\SpearmanED}{\ensuremath{\rho_s^{\text{ed}}}}
\newcommand{\SpearmanHist}{\ensuremath{\rho_s^{\text{h}}}}

\begin{document}\sloppy


\title{MuSViT: A Foundation Vision Model\\for Sheet Music Representation}

\author{Carlos Penarrubia}
\author{Antonio Rios-Vila}
\author{Eliseo Fuentes-Martinez}
\author{Juan C. Martinez-Sevilla}
\author{Francisco J. Castellanos}
\author{María Alfaro-Contreras}
\author{Jorge Calvo-Zaragoza}

\affil{Pattern Recognition and Artificial Intelligence Group, University of Alicante, Spain}


\maketitle

\begin{figure}[!ht]
    \centering
    \includegraphics[width=0.94\linewidth, trim = 0 15 0 60]{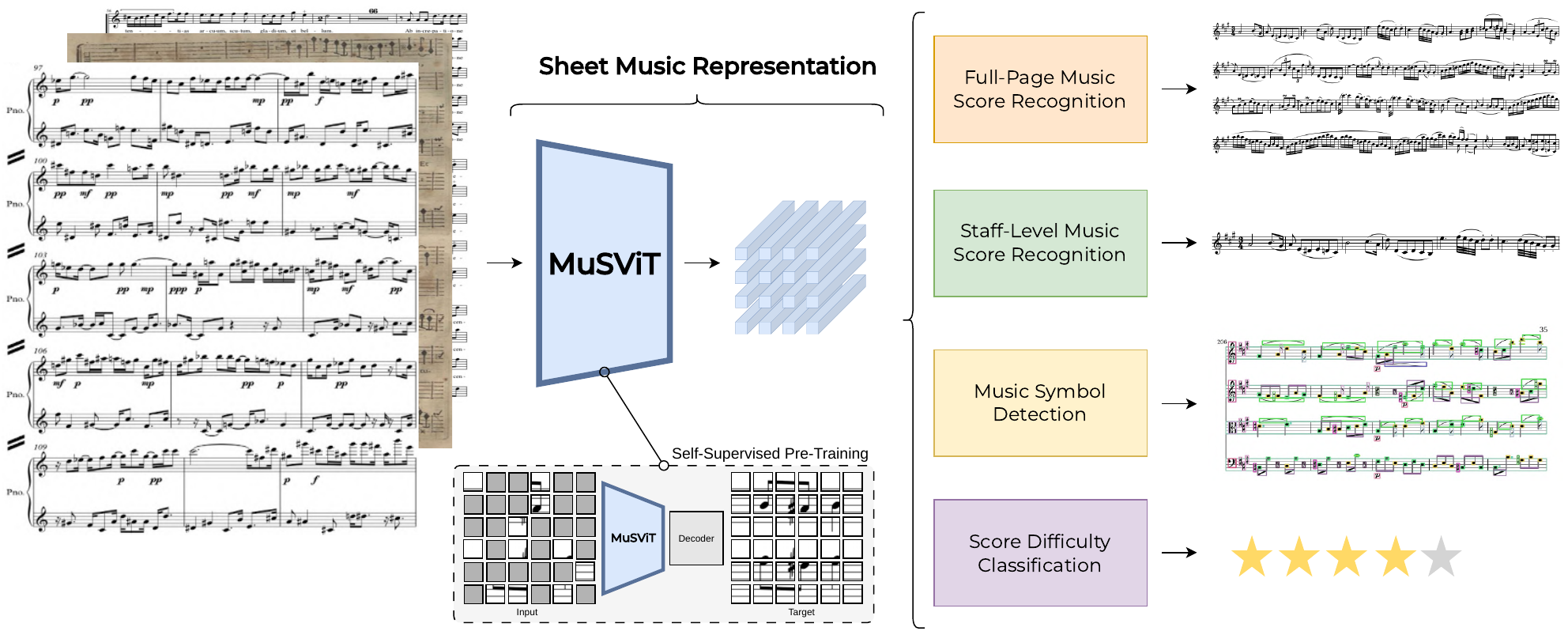}
    \caption{\textbf{Overview of \musvit{}.} \musvit{} is pre-trained on diverse sheet music pages using Masked Autoencoders: patches are randomly masked and the model learns to reconstruct the missing content from the remaining visible context. We evaluate the generality of the learned representations by probing the encoder across four diverse downstream tasks: full-page and staff-level music score recognition, music symbol detection, and score difficulty classification. Pre-trained models, code, and reproducibility resources are available through the project page: \url{https://grfia.dlsi.ua.es/musvit}.}
    \label{fig:graphical_abstract}
\end{figure}

\begin{figure}[!ht]
    \centering
    \includegraphics[width=0.94\linewidth, trim = 0 15 0 60]{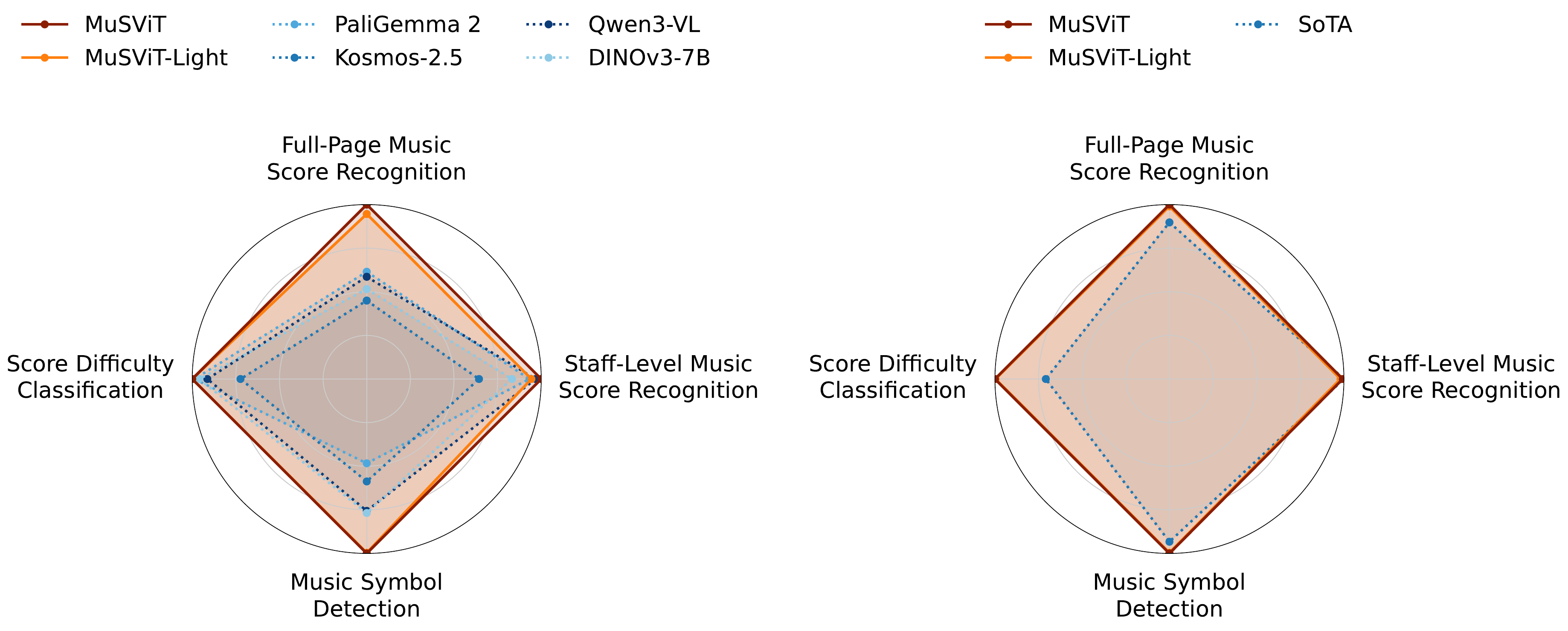}
    \caption{\musvit{} performance across four downstream tasks. \textbf{Left:} \emph{Linear probing} (frozen encoder)---\musvit{} (solid) consistently outperforms general-purpose vision encoders (dashed), demonstrating superior representation quality. \textbf{Right:} \emph{Fine-tuning}---\musvit{} generally outperforms state-of-the-art methods (SoTA). Axes represent normalized performance on each task (higher is better); see Section~\ref{sec:experiments} for detailed results.}
    \label{fig:spider_overview}
\end{figure}

\begin{abstract}
Foundation models have transformed vision and language processing by providing rich, reusable representations that transfer across diverse tasks. Sheet music, as a visual encoding of musical language, lacks such a strong domain-specific backbone. We introduce \musvit{} (\textbf{Mu}sic \textbf{S}core \textbf{Vi}sion \textbf{T}ransformer): the first foundation vision model for sheet music representation---a ViT encoder pre-trained via Masked Autoencoders on 9.7 million pages from the \acf{IMSLP}. To handle the complexity of real-world scores, we adopt a two-stage curriculum: a synthetic warm-up on typeset scores followed by large-scale training on the full \acs{IMSLP} corpus. We evaluate \musvit{} on four downstream tasks---full-page and staff-level music score recognition, music symbol detection, and score difficulty classification---under two scenarios: linear probing (frozen encoder) and fine-tuning. Under linear probing, \musvit{} consistently outperforms modern vision encoders, revealing that general-purpose representations, regardless of scale, fall systematically short on the structured symbolic properties of musical notation. Under fine-tuning, \musvit{} generally improves upon task-specific state-of-the-art methods. An additional embedding-transcription consistency analysis reveals that \musvit{} encodes symbolic musical structure directly in its representation space---unlike other encoders, whose embeddings do not correlate with music notation content. These results establish \musvit{} as a foundation backbone for sheet music understanding.
\end{abstract}



  
\section{Introduction}
\label{sec:introduction}

Music represents an essential part of human tradition and a valuable element of cultural heritage. Over the centuries, musical compositions have been transmitted both orally and in written form, with music encoded in documents known as scores (or sheet music). Today, vast collections of these scores are preserved in libraries and archives, and large-scale digitization initiatives have made millions of pages available online.

However, the majority of this material exists only as raw page images. Due to the high cost of manual transcription, most scores have not been converted into structured digital formats that enable indexing, retrieval, or automated analysis. As a result, despite the growing availability of digitized music documents, much of this cultural treasure remains effectively inaccessible and underutilized.

Traditional approaches to automated music document analysis---most notably in the field of Optical Music Recognition (OMR)---have focused on task-specific systems, employing end-to-end models or multi-stage pipelines trained on limited annotated datasets~\cite{Calvo-Zaragoza:ACM:2020}. These systems often remain brittle, with performance degrading sharply when confronted with unfamiliar notation styles, engraving conventions, or visual artifacts that differ from the training data~\cite{Tuggener:TISMIR:2024}. The extreme heterogeneity of real-world music scores continues to limit the generalization capabilities of these specialized models.

Recent advances in deep learning have demonstrated the value of foundation models for \emph{representation learning} from large-scale unlabeled data. In computer vision, Vision Transformers (ViTs) pre-trained via self-supervised learning---such as DINO~\cite{Caron:CVPR:2021,Oquab:Dinov2:2024,Simeoni:Dinov3:2025}---learn transferable visual representations that can serve as general-purpose backbones for diverse downstream tasks. Language-image pre-training approaches such as CLIP~\cite{Radford:ICML:2021} and SigLIP~\cite{Zhai:ICCV:2023}, together with vision-language models such as PaliGemma~\cite{Beyer:Paligemma:2024,Steiner:Paligemma2:2024} and Qwen-VL~\cite{Bai:Qwen:2023,Wang:Qwen2:2024,Bai:Qwen3:2025}, extend these principles to multimodal understanding. 

A similar trend is observed in the audio domain: general-purpose models such as Qwen-Audio~\cite{Chu:QwenAudio:2023,Chu:QwenAudio:2024} and Audio Flamingo~\cite{Kong:ICML:2024,Kong:ICML:2025,Kong:NeurIPS:2025} address a wide range of audio tasks, while music-specific foundation models~\cite{Yizhi:ICLR:2024,Qingqing:ISMIR:2022} demonstrate that domain-specific pre-training consistently outperforms general-purpose alternatives on music audio tasks. Despite this progress across text, vision, and audio, to the best of our knowledge, \emph{no foundation model exists for sheet music}---a highly structured visual domain with unique symbolic properties.

In this paper, we introduce \musvit{} (\textbf{Mu}sic \textbf{S}core \textbf{Vi}sion \textbf{T}ransformer), a ViT encoder pre-trained on 9.7 million sheet music pages from the \ac{IMSLP}\footnote{\href{https://imslp.org/}{https://imslp.org/}} using Masked Image Modeling (MIM)~\cite{Hondru:IJCV:2025}. \musvit{} learns rich visual representations by reconstructing masked regions of full-page sheet music images, thus capturing the structural and symbolic properties of music notation without requiring any annotated data. An overview of \musvit{} is shown in Fig.~\ref{fig:graphical_abstract}.

To assess the quality of the learned representations, we evaluate \musvit{} on four diverse downstream tasks from the related literature: \emph{full-page music score recognition} (page-level transcription)~\cite{RiosVila:IJCV:2026}, \emph{staff-level music score recognition} (staff-level transcription)~\cite{Martinez:ISMIR:2024}, \emph{music symbol detection} (dense object localization)~\cite{Luo:ESWA:2024}, and \emph{score difficulty classification} (document-level estimation)~\cite{Ramoneda:ISMIR:2023}. Following standard practice for evaluating foundation models~\cite{Awais:TPAMI:2025,Radford:ICML:2021,Chu:QwenAudio:2023, Kong:ICML:2024,Beyer:Paligemma:2024,Yizhi:ICLR:2024}, we adopt two complementary protocols: \emph{linear probing}, where the encoder is frozen and only lightweight task-specific heads are trained, which directly measures the quality of the extracted representations against other pre-trained vision encoders; and \emph{fine-tuning}, where the encoder is trained jointly with a downstream head, which measures transfer-learning capacity and enables a fair comparison against state-of-the-art task-specific methods. Our results show that \musvit{} achieves strong performance in both scenarios (see Fig.~\ref{fig:spider_overview}), highlighting the importance of music-specific representations for sheet music tasks.

Beyond task performance, an embedding-transcription consistency analysis demonstrates that \musvit{} representations align more closely with symbolic musical content than those of general-purpose vision encoders. This provides direct evidence that domain-specific pre-training captures musically meaningful structures and semantics.

The contributions of this work are summarized as follows:
\begin{enumerate}
    \item We introduce and publicly release \musvit{}, the first vision foundation model for sheet music representation. \musvit{} is a ViT encoder pre-trained on 9.7 million pages of sheet music collected from the \ac{IMSLP} via MIM.\footnote{The model, pre-training code, and evaluation scripts are available through the project page: \url{https://grfia.dlsi.ua.es/musvit}.}

    \item We evaluate \musvit{} on four representative downstream tasks---full-page music score recognition, staff-level music score recognition, music symbol detection, and score difficulty classification---under two scenarios: \emph{linear probing} (frozen encoder) and \emph{fine-tuning}.
    
    \begin{enumerate}
        \item Under \emph{linear probing}, \musvit{} consistently outperforms general-purpose pre-trained vision encoders, suggesting that sheet music analysis requires representations that general computer vision models---regardless of scale or pre-training diversity---do not capture well.
       
        \item Under \emph{fine-tuning}, \musvit{} surpasses task-specific state-of-the-art methods on three of the four tasks and matches the best results on the fourth, confirming that music-specific pre-training provides a strong initialization for task-specific adaptation.    
    \end{enumerate}
    
    \item Our embedding-transcription consistency analysis provides direct evidence that \musvit{} representations correlate with symbolic musical content, in contrast to those of general-purpose vision models.
\end{enumerate}

By publicly releasing \musvit{}, we aim to provide the research community with a foundation representation layer that accelerates progress across a wide range of sheet music applications.

\section{MuSViT: Music Score Vision Transformer}
\label{sec:methodology}

\musvit{} (\textbf{Mu}sic \textbf{S}core \textbf{Vi}sion \textbf{T}ransformer) is a self-supervised Vision Transformer (ViT) designed to learn meaningful visual representations for sheet music images. These serve as a general backbone that can be reused across diverse downstream tasks. This section describes \musvit{} in detail, covering the pre-training strategy, the encoder architecture, and the training data.

\subsection{Pre-Training Strategy}
\label{subsec:pretraining_strategy}

Sheet music differs fundamentally from natural images: it is a visual encoding of a symbolic language, where every element---such as noteheads, stems, accidentals, or rests---has a precise semantic meaning in terms of duration and pitch, following strict music notation rules. This structured nature has two important implications for the design of a ViT-based pre-training strategy. First, musical symbols are small and dense, requiring fine-grained patches such that each patch captures only elementary notational content---i.e. a single notehead, stem, or accidental---rather than spanning entire notes or measures, which would cause distinct symbols to blend within a single token. Second, the rigid spatial grammar of music notation makes local context highly informative: the identity and position of neighboring symbols strongly constrain what can appear in a masked region.

\begin{figure}[ht!]
  \centering
  \includegraphics[width=0.9\linewidth, trim = 0 15 0 60]{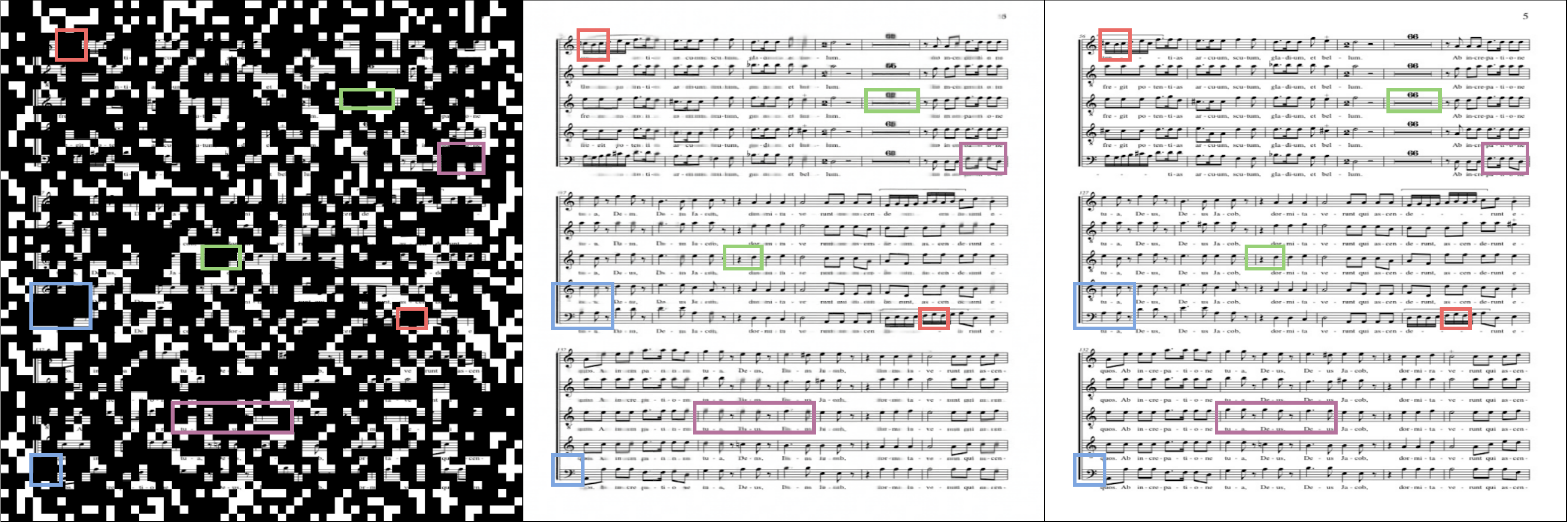}
  \caption{\musvit{} reconstruction example. \textbf{Left:} Masked music score image. \textbf{Middle:} \musvit{} reconstruction. \textbf{Right:} The original sheet music. The coloured rectangles highlight different reconstruction components: {\color{red} \emph{Red}} indicates pitch sequence reconstruction (staff position); {\color{cyan} \emph{Blue}} represents clef reconstruction; {\color{ForestGreen} \emph{Green}} marks rest reconstruction; and {\color{violet} \emph{Purple}} denotes musical note sequence reconstruction (musical symbol aligned with staff position).}
  \label{fig:reconstructions_in_detail}
\end{figure}

Among MIM approaches~\cite{Hondru:IJCV:2025}, we adopt Masked Autoencoders (MAE)~\cite{He:CVPR:2022} for pre-training. MAE learns visual representations by reconstructing randomly masked portions of the input image. With a high masking ratio, entire measures and long sequences of symbols are occluded at once. To reconstruct these missing regions, the encoder cannot simply interpolate textures as it might for natural images; instead, it must infer which symbols should appear (duration), where they should be placed vertically on the staff (pitch), and how they fit into the surrounding sequence (context). In other words, {MAE forces the model to learn the visual language of music notation without any supervision}. This is the main reason we choose MAE: its high masking ratio naturally aligns with the structured, symbolic nature of sheet music, requiring the encoder to develop a deep understanding of musical notation in order to solve the reconstruction task. A qualitative reconstruction example is shown in Fig.~\ref{fig:reconstructions_in_detail}.

Additionally, MAE reconstruction objective operates directly on pixel values and requires no external tokenizer or pre-trained teacher. This helps avoid domain mismatch, since no foundation model for sheet music currently exists. Its asymmetric encoder-decoder design, where the encoder processes only visible patches, also makes training efficient despite the long patch sequences that result from the fine-grained patch sizes required by our domain.

We now describe the pre-training procedure. Given an RGB input image $\mathbf{x} \in \mathbb{R}^{3 \times H \times W}$, we split it into a grid of non-overlapping $P \times P$ patches, yielding $N = \frac{H}{P} \times \frac{W}{P}$ patches per image. 
We then follow the two-stage curriculum learning strategy, with varying masking ratios:
\begin{itemize}
    \item \textbf{Stage~1 (Synthetic warm-up).} We initialize training on the \DeepScores{} dataset~\cite{Tuggener:DeepScoresv2:2020}, a synthetic corpus with lower visual variability and cleaner notation. This stage is not merely beneficial but necessary since we observed that training directly on real-world sheet music results in highly unstable and unreliable convergence, with the model predicting average pixel values rather than structured content. Therefore, starting on synthetic data allows the encoder to learn basic structures before encountering the high complexity of real-world scores. We use a masking ratio of 50\% to provide more visible context and ease convergence. Training operates on random $512 \times 512$ crops with patch size $P = 16$, yielding $N = 32 \times 32 = 1{,}024$ patches per crop, ensuring each patch captures elementary notational content (e.g., a single notehead or accidental). 
    
    \item \textbf{Stage~2 (Real-world adaptation).} We transfer to the full \ac{IMSLP} corpus, exposing the model to the diversity of real-world sheet music. We increase the masking ratio to 70\% and move from crops to full pages resized to $1{,}024 \times 1{,}024$ pixels, maintaining patch size $P = 16$ and yielding $N = 64 \times 64 = 4{,}096$ patches per image. This transition introduces full-page context---including global layout structure, multi-system arrangements, and page-level spatial relationships---while the consistent patch size ensures that the local features learned in Stage~1 remain applicable.
\end{itemize}

An ablation comparing the proposed curriculum against single-stage MAE training directly on IMSLP confirms that this two-stage design is necessary: without the synthetic warm-up, the encoder suffers dimensional collapse~\cite{jing2021understanding,zhang2022mask}---the decoder predicts average patches rather than music-notation structures. A detailed analysis of the singular value spectrum and effective rank is provided in the ``Supplementary Material''.


Unlike distillation-based approaches such as DINO~\cite{Caron:CVPR:2021,Oquab:Dinov2:2024}, which require aggressive augmentation (including Gaussian noise that can easily erase staff lines---destroying pitch information, a well-known challenge in music document processing~\cite{Fujinaga:IGI:2004}), the MAE reconstruction objective naturally encourages robust learning with minimal augmentation. We apply only random resized cropping to preserve the structural integrity of musical notation. Full pre-training details (learning rate, batch size, epochs, warm-up schedule, weight decay) are provided in the ``Supplementary Material''.

\subsection{Architecture}
\label{subsec:architecture}

The \musvit{} encoder employs a ViT architecture~\cite{Dosovitskiy:ICLR:2021}, comprising 12 Transformer layers with approximately 85M parameters. Each $P \times P$ patch from the input image is flattened and linearly projected into a $d = 768$-dimensional embedding vector. Following MAE design~\cite{He:CVPR:2022} and prior work on music document recognition~\cite{RiosVila:IJCV:2026}, we use 2D sinusoidal positional encodings (PE) rather than the standard 1D PE, as 2D PE explicitly encodes vertical position and helps the model identify elements along the height of the document---information that 1D PE forces the model to learn implicitly. During pre-training, only the visible (unmasked) patches are processed by the encoder, as described in Section~\ref{subsec:pretraining_strategy}.

Adhering to MAE design philosophy~\cite{He:CVPR:2022}, a lightweight decoder is used exclusively during pre-training and discarded afterward. The encoded visible patch embeddings are projected to the decoder dimension, after which learnable mask tokens are inserted into the sequence at the positions of missing patches. As with the encoder, 2D PE is applied to the full sequence to maintain spatial correspondence. Each output at a masked position is linearly projected to reconstruct the patch pixel values. The training loss is the mean squared error (MSE) between predicted and target pixels, applied only to masked patches. Full architecture details are provided in the ``Supplementary Material''.

In addition to \musvit{}, we train a lightweight variant, \musvitl{}, which shares the same 12-layer Transformer architecture but uses a smaller embedding dimension ($d = 384$), totalling approximately 25M parameters. \musvitl{} targets limited computational resources scenarios, while \musvit{} is intended for maximum performance.

\subsection{IMSLP training data}
\label{subsec:training_data}

\begin{figure}[t]
    \centering
    \includegraphics[width=1.0\linewidth, trim = 0 15 0 60]{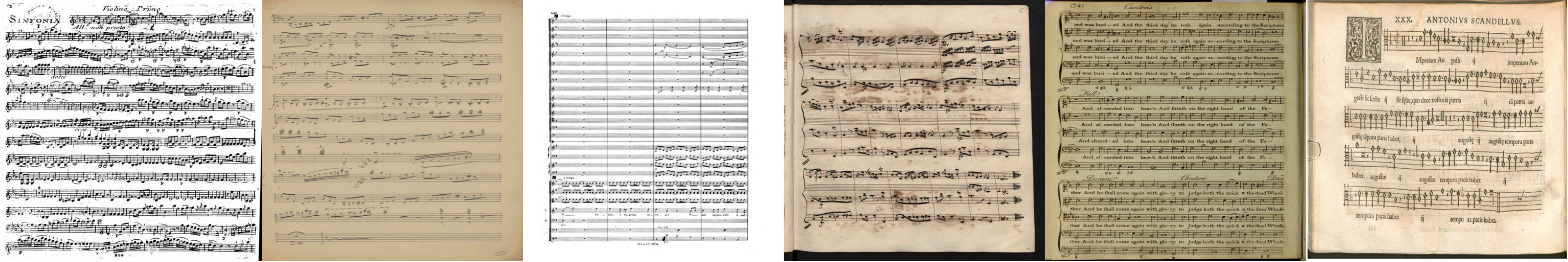}
    \caption{Representative pages from the \ac{IMSLP} pre-training corpus, illustrating its visual diversity. The collection spans historical periods, notation systems, engraving styles, and musical textures.}
    \label{fig:imslp_examples}
\end{figure}

\musvit{} is pre-trained on a large-scale corpus of public-domain sheet music pages collected from the \acf{IMSLP}.\footnote{All data used for pre-training is sourced from publicly available scans that are in the public domain or permissively licensed under the \ac{IMSLP} terms of use.} The \ac{IMSLP} hosts hundreds of thousands of scanned music scores spanning multiple centuries, composers, and engraving styles. Unlike synthetic or homogeneous corpora, this collection captures a wide spectrum of real-world sheet music variability---from monophonic vocal lines to dense orchestral scores, across both modern typeset and historical engraving conventions---providing the visual diversity necessary for learning generalizable representations. Specifically, we collect 9.7 million pages drawn from around 400,000 distinct musical works. Each page is rendered as an RGB image. Representative samples are shown in Fig.~\ref{fig:imslp_examples}; additional examples are provided in the ``Supplementary Material''.

\section{Evaluation on Downstream Tasks}
\label{sec:experiments}

Sheet music analysis encompasses a wide range of perceptual and structural challenges, from localizing individual symbols to transcribing entire pages and estimating high-level score properties. We probe \musvit{} representations across four tasks that collectively cover this spectrum: \emph{full-page music score recognition}~\cite{RiosVila:IJCV:2026}, \emph{staff-level music score recognition}~\cite{Martinez:ISMIR:2024}, \emph{music symbol detection}~\cite{Luo:ESWA:2024}, and \emph{score difficulty classification}~\cite{Ramoneda:ISMIR:2023}. We believe that performance across this diverse set of tasks serves as a reliable proxy for the generality of \musvit{} representations for sheet music analysis. Table~\ref{tab:task_summary} summarizes the datasets and metrics. For each task, we consider two evaluation scenarios:

\begin{enumerate}
    \item \textbf{Linear probing.} The \musvit{} encoder remains frozen and only lightweight, task-specific heads are trained on top of the extracted representations. This protocol isolates the intrinsic quality of the learned features and provides a controlled comparison against other pre-trained vision encoders, all of them evaluated under identical frozen conditions.
    \item \textbf{Fine-tuning.} The \musvit{} encoder is unfrozen and trained jointly with the task-specific head, allowing the representations to adapt to each downstream objective. Since each task has its own established training and evaluation protocols, we follow the respective protocols in each case to ensure a fully comparable evaluation against state-of-the-art methods. Full training details are provided in the ``Supplementary Material''.
\end{enumerate}

\begin{table}[!ht]
    \centering
    \setlength{\abovecaptionskip}{1pt}   
    \setlength{\belowcaptionskip}{1pt}   
    \caption{Overview of the four downstream tasks used to evaluate \musvit{}. Each task targets a different level of sheet music understanding---from sequence-level transcription (full-page and staff-level recognition) to dense spatial localization (symbol detection) and document-level semantics (difficulty classification). We report the datasets and primary evaluation metrics for each task.}
    \label{tab:task_summary}
    \setlength{\tabcolsep}{8pt}
    \renewcommand{\arraystretch}{1.6} 
    \resizebox{\columnwidth}{!}{%
        \begin{tabular}{p{3.5cm} p{4.cm} l c}
        \toprule[1.5pt]
        \textbf{Task} & \textbf{Description}                                  & \textbf{Datasets}                                                                                                                                                         & \textbf{Metric} (\%) \\ \cmidrule(lr){1-4}
        
        Full-Page Music Score Recognition   
            & Transcribe full music score pages                     & {\Mozarteum{}\tablefootnote{\href{https://dme.mozarteum.at/movi/en}{https://dme.mozarteum.at/movi/en}}, \Polish{}\tablefootnote{\href{https://github.com/pl-wnifc/humdrum-polish-scores}{https://github.com/pl-wnifc/humdrum-polish-scores}}}       
            & {SER $\downarrow$} \\
        
        {Staff-Level Music Score Recognition} 
        & Transcribe single staff-level images 
        & {\Capitan{}~\cite{Calvo-Zaragoza:PRL:2019}, \Guatemala{}~\cite{Thomae:DLfM:2022}, \ILS{}~\cite{Parada-Cabaleiro:ISMIR:2019}, \Malaga{}~\cite{Madueno:DLfM:2021}, \FMT{}~\cite{Rios-Vila:Applied-Sciences:2021}} 
        & SER $\downarrow$ \\
        
        {Music Symbol Detection}              
        & Localize and classify music symbols
        & {\DeepScores{}~\cite{Tuggener:DeepScoresv2:2020}}
        & {mAP, w-mAP $\uparrow$} \\
                                        
        {Score Difficulty Classification}     
        & Estimate performance difficulty from score images     
        & {\FreeScores{}~\cite{Ramoneda:ISMIR:2023}, \PlayIt{}~\cite{Ramoneda:ISMIR:2023}, \PianoStreet{}~\cite{Ramoneda:ISMIR:2023}}                                                                                                                         & $Acc_{0}, Acc_{1} \uparrow$ \\
          
        \bottomrule[1.5pt]
        \end{tabular}
    }
\end{table}


Together, these two scenarios disentangle the contribution of pre-trained representations (linear probing) from the model capacity when allowed to specialize (fine-tuning), offering a comprehensive view of \musvit{} as both a general-purpose foundation layer and a competitive task-specific model. Due to space constraints, we here report results averaged across datasets for all tasks; per-dataset breakdowns are available in the ``Supplementary Material''.

We compare \musvit{} against four general-purpose vision encoders representing complementary pre-training paradigms: (i) \textbf{DINOv3-7B}~\cite{Simeoni:Dinov3:2025}, latest large-scale self-supervised vision model; (ii) \textbf{Qwen3-VL}~\cite{Bai:Qwen3:2025}, state-of-the-art vision-language model pre-trained on large-scale image--text corpora; (iii) \textbf{PaliGemma~2}~\cite{Steiner:Paligemma2:2024}, vision-language model that combines a SigLIP vision encoder pre-trained on diverse language--image pairs with a language model; and (iv) \textbf{Kosmos-2.5}~\cite{Tengchao:Kosmos25:2025}, document-oriented multimodal model designed for structured visual inputs such as forms and documents. Only the vision encoder component is used for evaluation. To our knowledge, none of these encoders has been exposed to sheet music during pre-training, making them direct controls for measuring the value of music-specific visual representations.

Figure~\ref{fig:spider_overview} provides a high-level summary of results across all four tasks and both evaluation scenarios. Under \emph{linear probing}, \musvit{} achieves the largest enclosed area in the radar chart, outperforming all general-purpose vision encoders across every task. Under \emph{fine-tuning}, \musvit{} matches or surpasses task-specific state-of-the-art methods. The following subsections present the detailed results for each task.\footnote{Additionally, we fine-tuned all general-purpose models on two representative tasks to verify that the gap is not an artifact of the frozen evaluation protocol; \musvit{} remains superior even against fine-tuned four baselines while being 5--82$\times$ smaller in parameters and 16--260$\times$ more efficient in GFLOPs. These results and detailed computational cost comparison are provided in the ``Supplementary Material''.}

\subsection{Full-Page Music Score Recognition}
\label{subsec:full_page_omr}

Full-page music score recognition is the task of transcribing an entire score image into its corresponding symbol-level sequence, considering the correct reading order. This task has classically been approached using a two-stage pipeline---where a layout analysis step first segments individual staves, which are then transcribed independently---but recent work has enabled end-to-end models that process the full page in a single pass~\cite{RiosVila:IJCV:2026}. This end-to-end formulation is a particularly demanding test of representation quality: the encoder must capture both fine-grained notational detail at the symbol level and the global spatial organization of the page.

\begin{table}[t]
    \centering
    \setlength{\abovecaptionskip}{1pt}   
    \setlength{\belowcaptionskip}{1pt}   
    \caption{Average Symbol Error Rate (SER \%, $\downarrow$) across all {full-page recognition} datasets under the \emph{linear probing} scenario (frozen encoder). Best result is in \textbf{bold}; second best is \underline{underlined}.}
    \label{tab:omr_full_page_avg_results_lp}
    \setlength{\tabcolsep}{12pt} 
    \renewcommand{\arraystretch}{1.5} 
    \resizebox{\columnwidth}{!}{%
        \begin{tabular}{lcccc >{\columncolor[gray]{0.9}}c >{\columncolor[gray]{0.9}}c}
            \toprule[1pt]
                            & PaliGemma 2~\cite{Steiner:Paligemma2:2024}    & Kosmos-2.5~\cite{Tengchao:Kosmos25:2025}  & Qwen3-VL~\cite{Bai:Qwen3:2025}        & DINOv3-7B~\cite{Simeoni:Dinov3:2025}   & \musvit{}        & \musvitl{} \\ \cmidrule(lr){1-7}
            \textbf{Avg.}   & 48.6    & 62.4   & 51.0  & 56.9   & \textbf{16.4}    & \underline{20.9}          \\ 
            \bottomrule[1pt]
        \end{tabular}
    }
\end{table}

\begin{table}[t]
    \centering
    \setlength{\abovecaptionskip}{1pt}   
    \setlength{\belowcaptionskip}{1pt}   
    \caption{Average Symbol Error Rate (SER \%, $\downarrow$) across all {full-page recognition} datasets under the \emph{fine-tuning} scenario. Best result is in \textbf{bold}; second best is \underline{underlined}.}
    \label{tab:omr_full_page_avg_results_ft}
    \setlength{\tabcolsep}{12pt} 
    \renewcommand{\arraystretch}{1.5} 
    \resizebox{0.6\columnwidth}{!}{%
        \begin{tabular}{lc >{\columncolor[gray]{0.9}}c >{\columncolor[gray]{0.9}}c}
            \toprule[1pt]
                            & State-of-the-art~\cite{RiosVila:IJCV:2026}    & \musvit{}           & \musvitl{}  \\ \cmidrule(lr){1-4}
            \textbf{Avg.}   & 20.0                                          & \textbf{10.9}       & \underline{11.8}   \\ 
            \bottomrule[1pt]
        \end{tabular}
    }
\end{table}


We evaluate on two datasets of scanned pianoform scores: \Mozarteum{} (101 pages, printed CWMN) and \Polish{} (117 pages, printed CWMN). We adopt an autoregressive Transformer decoder~\cite{RiosVila:IJCV:2026} on top of the \musvit{} encoder, which attends autoregressively to the patch embeddings to generate the output sequence. Performance is measured using the Symbol Error Rate (SER), defined as the normalized edit distance between predicted and ground-truth sequences~\cite{RiosVila:IJCV:2026}.

Table~\ref{tab:omr_full_page_avg_results_lp} reveals a striking gap under linear probing: all general-purpose encoders produce SER values in the 48--62\% range, while \musvit{} achieves 16.4\%---more than 2.5$\times$ lower than the best baseline (PaliGemma~2, 48.6\%). Remarkably, \musvit{} already outperforms the state-of-the-art performance~\cite{RiosVila:IJCV:2026} (20.0\%). These results show that \musvit{} simultaneously encodes fine-grained symbol-level detail and the global reading-order structure spanning the entire page, whereas no general-purpose encoder provides representations sufficient for this dual requirement.


Table~\ref{tab:omr_full_page_avg_results_ft} shows that under fine-tuning \musvit{} achieves 10.9\% average SER, outperforming the state of the art~\cite{RiosVila:IJCV:2026} by 9.1 points. Per-dataset results (see ``Supplementary Material'') show that this improvement is consistent across both corpora, with a particularly large gain on \Polish{} where the prior system struggled most, suggesting that pre-training on large-scale diverse scores yields a more robust initialization than task-specific supervision alone.


\subsection{Staff-Level Music Score Recognition}
\label{subsec:staff_line_omr}

Staff-level music score recognition shares the same transcription objective as full-page recognition but operates on individually segmented staff images rather than complete pages, requiring a prior layout analysis step. Although this intermediate segmentation makes the transcription pipeline less end-to-end, staff-level recognition is among the most established benchmarks in music transcription research, providing a complementary evaluation at finer spatial granularity.

\begin{table}[t]
    \centering
    \setlength{\abovecaptionskip}{1pt}   
    \setlength{\belowcaptionskip}{1pt}   
    \caption{Average Symbol Error Rate (SER \%, $\downarrow$) across all {staff-level recognition} datasets under the \emph{linear probing} scenario (frozen encoder). Best result is in \textbf{bold}; second best is \underline{underlined}.}
    \label{tab:omr_staff_line_avg_results_lp}
    \setlength{\tabcolsep}{12pt} 
    \renewcommand{\arraystretch}{1.5} 
    \resizebox{\columnwidth}{!}{%
        \begin{tabular}{lcccc >{\columncolor[gray]{0.9}}c >{\columncolor[gray]{0.9}}c}
            \toprule[1pt]
                            & PaliGemma 2~\cite{Steiner:Paligemma2:2024}    & Kosmos-2.5~\cite{Tengchao:Kosmos25:2025}  & Qwen3-VL~\cite{Bai:Qwen3:2025}    & DINOv3-7B~\cite{Simeoni:Dinov3:2025}      & \musvit{}             & \musvitl{} \\ \cmidrule(lr){1-7}
            \textbf{Avg.}   & 23.9                                          & 47.5                                  & \underline{21.0}                             & 32.1                                                                                & \textbf{18.4}         &  23.0              \\ 
            \bottomrule[1pt]
        \end{tabular}
    }
\end{table}


\begin{table}[t]
    \centering
    \setlength{\abovecaptionskip}{1pt}   
    \setlength{\belowcaptionskip}{1pt}   
    \caption{Average Symbol Error Rate (SER \%, $\downarrow$) across all {staff-level recognition} datasets under the \emph{fine-tuning} scenario. Best result is in \textbf{bold}; second best is \underline{underlined}.}
    \label{tab:omr_staff_line_avg_results_ft}
    \setlength{\tabcolsep}{12pt} 
    \renewcommand{\arraystretch}{1.5} 
    \resizebox{0.6\columnwidth}{!}{%
        \begin{tabular}{lc >{\columncolor[gray]{0.9}}c >{\columncolor[gray]{0.9}}c}
            \toprule[1pt]
                            & State-of-the-art~\cite{Martinez:ISMIR:2024}   & \musvit{}         & \musvitl{}\\ \cmidrule(lr){1-4}
            \textbf{Avg.}   & \textbf{8.0}                                  & \underline{8.6}   & 9.8             \\ 
            \bottomrule[1pt]
        \end{tabular}
    }
\end{table}


We evaluate \musvit{} on five corpora spanning both historical and modern notation, as well as printed and handwritten engraving styles: \Capitan{} (828 staves, handwritten mensural)~\cite{Calvo-Zaragoza:PRL:2019}, \Guatemala{} (3,263 staves, handwritten mensural)~\cite{Thomae:DLfM:2022}, \ILS{} (1,136 staves, printed mensural)~\cite{Parada-Cabaleiro:ISMIR:2019}, \Malaga{} (308 staves, printed CWMN)~\cite{Madueno:DLfM:2021}, and \FMT{} (1,305 staves, handwritten CWMN)~\cite{Rios-Vila:Applied-Sciences:2021}. We follow the recurrent neural network-based recognition head of~\cite{Martinez:ISMIR:2024} and report SER values.

Table~\ref{tab:omr_staff_line_avg_results_lp} shows that under linear probing \musvit{} achieves 18.4\% average SER---the best result among all encoders, 2.6 points ahead of Qwen3-VL (21.0\%). Per-dataset results (see ``Supplementary Material'') follow the same trend. The gap over DINOv3-7B (32.1\%) and Kosmos-2.5 (47.5\%) is large, confirming that neither scale nor linguistic pre-training substitutes for music-specific visual representations.

Table~\ref{tab:omr_staff_line_avg_results_ft} shows that under fine-tuning \musvit{} reaches 8.6\% average SER, within 0.6 points of the task-specific state of the art~\cite{Martinez:ISMIR:2024} (8.0\%). This narrow gap---smaller than the advantage seen in full-page recognition---is consistent with the reduced spatial complexity of the staff-level setting: because each input contains a single row of notation, the encoder does not need to simultaneously manage global page layout and local symbol detail, leaving less room for music-specific pre-training to provide an additional edge.

\subsection{Music Symbol Detection}
\label{subsec:detection}

Music symbol detection is the task of simultaneously localizing and classifying all individual notation elements within a score image, producing a set of bounding boxes with class labels for every visible symbol (noteheads, stems, accidentals, rests, clefs, etc). Unlike the transcription tasks above, this is a dense spatial task: it requires the model to reason about 2D locations rather than reading order.

\begin{table}[t]
    \centering
    \setlength{\abovecaptionskip}{1pt}   
    \setlength{\belowcaptionskip}{1pt}   
    \caption{mAP and weighted mAP ($\uparrow$) on the \DeepScores{} dataset for {music symbol detection} under the \emph{linear probing} scenario (frozen encoder). Best results are in \textbf{bold}; second best are \underline{underlined}.}
    \label{tab:symbol_detection_lp}
    \setlength{\tabcolsep}{12pt} 
    \renewcommand{\arraystretch}{1.5} 
    \resizebox{\columnwidth}{!}{%
        \begin{tabular}{lcccc >{\columncolor[gray]{0.9}}c >{\columncolor[gray]{0.9}}c}
            \toprule[1pt]
                            & PaliGemma 2~\cite{Steiner:Paligemma2:2024} & Kosmos-2.5~\cite{Tengchao:Kosmos25:2025}  & Qwen3-VL~\cite{Bai:Qwen3:2025}    & DINOv3-7B~\cite{Simeoni:Dinov3:2025}      & \musvit{}             & \musvitl{} \\ \cmidrule(lr){1-7}
            \textbf{mAP}           & 31.7      & 42.7     & 67.1     & 70.4     & \textbf{79.7}    & \underline{79.1}              \\ 
            \textbf{w-mAP}         & 39.0      & 47.4     & 61.0     & 62.0     & \textbf{80.7}    & \underline{80.4}              \\ 
            \bottomrule[1pt]
        \end{tabular}
    }
\end{table}


We evaluate on \DeepScores{}~\cite{Tuggener:DeepScoresv2:2020} (1,714 scores, 135 symbol classes) using a Faster R-CNN~\cite{Ren:NeurIPS:2015}, reporting mAP and w-mAP for linear probing. Concerning fine-tuning, we consider mAP$_{50}$ to mirror the current state of the art~\cite{Luo:ESWA:2024}; full implementation details are in the ``Supplementary Material''.

\begin{table}[t]
    \centering
    \setlength{\abovecaptionskip}{1pt}   
    \setlength{\belowcaptionskip}{1pt}   
    \caption{mAP$_{50}$ (\%, $\uparrow$) on the \DeepScores{} dataset for {music symbol detection} under the \emph{fine-tuning} scenario. Best results are in \textbf{bold}; second best are \underline{underlined}.}
    \label{tab:symbol_detection_ft}
    \setlength{\tabcolsep}{12pt} 
    \renewcommand{\arraystretch}{1.5} 
    \resizebox{0.6\columnwidth}{!}{%
        \begin{tabular}{lc >{\columncolor[gray]{0.9}}c >{\columncolor[gray]{0.9}}c}
            \toprule[1pt]
                            & State-of-the-art~\cite{Luo:ESWA:2024}    & \musvit{}           & \musvitl{} \\ \cmidrule(lr){1-4}
            \textbf{mAP$_{\mathbf{50}}$}   & 90.5                                         & \textbf{97.0}       &  \underline{96.6}  \\ 
            \bottomrule[1pt]
        \end{tabular}
    }
\end{table}


Table~\ref{tab:symbol_detection_lp} shows that, under linear probing, \musvit{} achieves 79.7\% mAP and 80.7\% w-mAP---both best among all encoders---with \musvitl{} within 0.6 and 0.3 points, respectively. The gap over DINOv3-7B is substantial on both metrics---9.3 points in mAP (70.4\%) and 18.7 in w-mAP (62.0\%). Vision-language models trail far behind. The large disparity between vision-only and language-aligned encoders suggests that language-aligned training actively degrades spatial precision on symbol-dense layouts.

Fine-tuning yields the most decisive gains in the entire evaluation as shown in Table~\ref{tab:symbol_detection_ft}: both \musvit{} variants surpass 96\% mAP$_{50}$, outperforming the state of the art~\cite{Luo:ESWA:2024} (90.5\%) by more than 6 points. This margin is doubly notable: \musvit{} uses a Faster R-CNN detector, considerably lighter than the Transformer-based architecture of the state of the art, and the encoder is only parameter-efficiently adapted via LoRA rather than fully fine-tuned (see ``Supplementary Material'')---both factors favour the baseline. The performance gap therefore reflects the quality of the pre-trained representations rather than a more powerful detection head or a more intense training.

\subsection{Score Difficulty Classification}
\label{subsec:difficulty}

Score difficulty classification aims at estimating the performance difficulty of a musical piece directly from its sheet music images~\cite{Ramoneda:ISMIR:2023}. Prior research has primarily addressed this task using symbolic music representations, which offer high interpretability but are limited by their scarce availability and dependence on successful prior transcription steps~\cite{Ramoneda:ISMIR:2024}. In contrast, the image-based formulation operates directly on raw score pages, bypassing the need for intermediate transcription and enabling application to any digitized score. Unlike recognition or detection, this task demands representations that aggregate holistic page-level properties rather than resolving individual symbols or their spatial arrangement.

\begin{table}[t]
    \centering
    \setlength{\abovecaptionskip}{1pt}   
    \setlength{\belowcaptionskip}{1pt}   
    \caption{Average $Acc_{0}$ and $Acc_{1}$ across all {score difficulty classification} datasets under the \emph{linear probing} scenario (frozen encoder). Best results are in \textbf{bold}; second best are \underline{underlined}.}
    \label{tab:diff_avg_results_lp}
    \setlength{\tabcolsep}{12pt} 
    \renewcommand{\arraystretch}{1.5} 
    \resizebox{\columnwidth}{!}{%
        \begin{tabular}{lcccc >{\columncolor[gray]{0.9}}c >{\columncolor[gray]{0.9}}c}
            \toprule[1pt]
                            & PaliGemma 2~\cite{Steiner:Paligemma2:2024} & Kosmos-2.5~\cite{Tengchao:Kosmos25:2025}  & Qwen3-VL~\cite{Bai:Qwen3:2025}    & DINOv3-7B~\cite{Simeoni:Dinov3:2025}      & \musvit{}             & \musvitl{} \\ \cmidrule(lr){1-7}
            \textbf{Avg. $\mathbf{Acc_{0}}$}   & 46.8    & 34.3   & 43.3   & 45.5  & \textbf{47.4}   &  \textbf{47.4}    \\ 
            \textbf{Avg. $\mathbf{Acc_{1}}$}   & 83.9   & 69.6   & 83.1   & 83.9  & \textbf{87.1}   & \underline{84.4}     \\ 
            \bottomrule[1pt]
        \end{tabular}
    }
\end{table}


\begin{table}[t]
    \centering
    \setlength{\abovecaptionskip}{1pt}   
    \setlength{\belowcaptionskip}{1pt}   
    \caption{Average $Acc_{0}$ and $Acc_{1}$ (\%, $\uparrow$) across all {score difficulty classification} datasets under the \emph{fine-tuning} scenario. Best results are in \textbf{bold}; second best are \underline{underlined}.}
    \label{tab:diff_avg_results_ft}
    \setlength{\tabcolsep}{12pt} 
    \renewcommand{\arraystretch}{1.5} 
    \resizebox{0.6\columnwidth}{!}{%
        \begin{tabular}{lc >{\columncolor[gray]{0.9}}c >{\columncolor[gray]{0.9}}c}
            \toprule[1pt]
                            & State-of-the-art~\cite{Ramoneda:ISMIR:2023}   & \musvit{}         & \musvitl{}\\ \cmidrule(lr){1-4}
            \textbf{Avg.} $\mathbf{Acc_{0}}$   & 38.4                                & \textbf{54.2}  & \underline{54.0}            \\ 
            \textbf{Avg.} $\mathbf{Acc_{1}}$   & 84.3                                &\textbf{89.3}   & \underline{88.9}             \\ 
            \bottomrule[1pt]
        \end{tabular}
    }
\end{table}


We evaluate \musvit{} on three piano corpora: \FreeScores{} (4,193 pieces, 5 difficulty levels)~\cite{Ramoneda:ISMIR:2023}, \PlayIt{} (652 pieces, 9 levels)~\cite{Ramoneda:ISMIR:2023}, and \PianoStreet{} (2,816 pieces, 9 levels)~\cite{Ramoneda:ISMIR:2023}. We report $Acc_{0}$ (exact match) and $Acc_{1}$ (within one adjacent level)~\cite{Ramoneda:ISMIR:2023}. We also mirror the evaluation protocol of~\cite{Ramoneda:ISMIR:2023}, which compares simpler and more complex classification heads and reports best results with the latter. Therefore, for {linear probing}, we adopt the simpler head: per-page patch embeddings are mean-pooled into a document-level embedding and passed to an MLP classifier on top of the frozen encoder; whereas for {fine-tuning}, we adopt a more expressive head: a GRU-based recurrent network that processes the sequence of page embeddings sequentially.

Table~\ref{tab:diff_avg_results_lp} show that, under linear probing, \musvit{} achieves 47.4\% $Acc_{0}$ and 87.1\% $Acc_{1}$, the best results among all encoders. Notably, \musvitl{} ties \musvit{} on $Acc_{0}$, suggesting that even the lightweight variant captures the global page-level properties relevant to difficulty estimation. Both \musvit{} variants already surpass the current state of the art~\cite{Ramoneda:ISMIR:2023} (38.4\% $Acc_{0}$, 84.3\% $Acc_{1}$) by 9.0 and 2.8 points respectively. The relatively small gap between \musvit{} and the best general-purpose encoder (PaliGemma~2, 46.8\% $Acc_{0}$) is consistent with the holistic nature of this task, where page-level visual statistics alone already provide a useful difficulty signal.

In the fine-tuning scenario, reflected by Table~\ref{tab:diff_avg_results_ft}, \musvit{} achieves 54.2\% $Acc_{0}$ and 89.3\% $Acc_{1}$, outperforming the state of the art~\cite{Ramoneda:ISMIR:2023} by 15.8 and 5.0 points, respectively. \musvitl{} matches these results closely (54.0\% $Acc_{0}$, 88.9\% $Acc_{1}$), with a margin of only 0.2 and 0.4 points. This represents the smallest gap between \musvit{} variants observed across all tasks, further reinforcing that difficulty estimation is well served by holistic representations rather than fine-grained encoder capacity.


\section{Embedding-Transcription Consistency Analysis}
\label{sec:analysis}

Although downstream task performance provides evidence of representation quality, it does not directly reveal whether the learned embeddings capture music-notation semantics. To address this question, we perform an embedding-transcription consistency analysis that quantifies how strongly the visual embedding space aligns with symbolic musical content (i.e., transcription). The hypothesis is simple: if \musvit{} learns music-aware representations, then pairs of score images with similar transcriptions should produce similar embeddings, and pairs with different transcriptions should be farther apart in the embedding space. We expect \musvit{} to exhibit stronger embedding–transcription alignment than general-purpose vision models, validating that music-specific pre-training captures symbolic musical structure rather than purely visual patterns.

We conduct this analysis on the \Mozarteum{} and \Polish{} datasets, for which both score images and symbolic transcriptions are available. For each image $i$, we extract an embedding vector $\mathbf{e}_i$ by flattening all token representations produced by the frozen encoder into a single vector. We compare \musvit{} against PaliGemma~2~\cite{Steiner:Paligemma2:2024}, Kosmos-2.5~\cite{Tengchao:Kosmos25:2025}, Qwen3-VL~\cite{Bai:Qwen3:2025}, and DINOv3-7B~\cite{Simeoni:Dinov3:2025}, all evaluated under identical frozen conditions.

To quantify similarity, we compute pairwise distances in two spaces. In the \emph{embedding space}, we use the Euclidean distance between embedding pairs, yielding $\Demb \in \mathbb{R}^{N \times N}$ with $D^{\text{emb}}_{ij} = \|\mathbf{e}_i - \mathbf{e}_j\|_2$. In the \emph{transcription space}, we compute two complementary measures: (1) \textbf{Edit distance} \Dedit{}, applying Levenshtein distance to transcription pairs, capturing differences in symbolic content and sequential ordering; and (2) \textbf{Histogram distance} \Dhist{}, computing distance between token frequency distributions, measuring compositional differences in notation while ignoring order.

To assess embedding–transcription consistency, we compute both Pearson ($\rho_p$) and Spearman ($\rho_s$) correlation~\cite{Benesty:NRSP:2009,Wissler:Science:1905} between corresponding entries of $\Demb$ and each transcription distance matrix, yielding four values per encoder: \PearsonED{}, \PearsonHist{}, \SpearmanED{}, and \SpearmanHist{}. A higher positive correlation indicates that the embedding space preserves symbolic similarity, i.e., transcriptionally similar images are placed closer in the embedding space.


Table~\ref{tab:correlation} reports the embedding--transcription correlation results. All general-purpose encoders yield low negative correlations across both distance measures and both correlation metrics, meaning their embedding spaces are not merely uncorrelated with musical content, but they are slightly {anti-correlated}: images with similar content might be placed {farther apart} in the embedding space. Because these encoders are optimized for visual appearance, they can be misled by surface-level cues that do not necessarily reflect the underlying musical structure. Therefore, visually similar pages may encode different musical content, and vice versa. In contrast, \musvit{} variants yield consistently high positive correlations, confirming that music-specific models are necessary for the encoder to capture symbolic musical structure. The larger gap on histogram distance suggests that \musvit{} models are particularly effective at capturing the distribution of music notation. We further complement this research with two analyses in the ``Supplementary Material'': an attention heat map that visualizes which image regions \musvit{} focuses on, and a k-curve analysis that examines local embedding structure across neighborhood sizes.

\begin{table}[t]
    \centering
    \setlength{\abovecaptionskip}{1pt}   
    \setlength{\belowcaptionskip}{1pt}   
    \caption{Pearson and Spearman correlation between pairwise embedding distances and pairwise transcription distances (edit and histogram) for each encoder. Higher positive values indicate stronger alignment between the visual embedding space and symbolic musical content. Best results are in \textbf{bold}; second best are \underline{underlined}.}
    \label{tab:correlation}
    \setlength{\tabcolsep}{6pt}
    \renewcommand{\arraystretch}{1.5}
    \resizebox{\columnwidth}{!}{%

    
    \begin{tabular}{ll cccc >{\columncolor[gray]{0.95}}c >{\columncolor[gray]{0.95}}c}
        \toprule[1.5pt]
        \multicolumn{2}{l}{} & 
            PaliGemma 2 & 
            Kosmos-2.5 & 
            Qwen3-VL & 
            DINOv3-7B & 
            \musvit{} & 
            \musvitl{} 
            \\ \midrule
        
        \multirow{2}{*}{\makecell[c]{\textbf{Pearson}\\\textbf{Correlation}}}  
            & \PearsonED{}   
                &  -0.110
                & -0.080 
                & -0.041 
                & -0.080 
                & \underline{0.606} 
                & \textbf{0.618} \\
                & \PearsonHist{}  
                & -0.127
                & -0.130 
                & -0.052 
                & -0.100  
                & \textbf{0.665}    
                & \underline{0.646} 
                \\ \midrule
        
        \multirow{2}{*}{\makecell[c]{\textbf{Spearman}\\\textbf{Correlation}}} & \SpearmanED{}   & -0.120 & -0.114 & -0.009 & -0.135 & \textbf{0.691}    & \underline{0.658} \\
                & \SpearmanHist{} & -0.132 & -0.153 & -0.009 & -0.152 & \textbf{0.714}    & \underline{0.662} \\
        
        \bottomrule[1.5pt]
    \end{tabular}
    }
\end{table}

\section{Conclusions}
\label{sec:conclusions}

We introduce \musvit{}, the first foundation vision model for sheet music representation: a ViT pre-trained via MAE on 9.7 million public-domain score pages from the \ac{IMSLP} through a two-stage curriculum progressing from synthetic to real-world data. As a result, \musvit{} representations are grounded in the structure of music notation. Across four downstream tasks, \musvit{} generally outperforms other vision encoders under linear probing and task-specific state-of-the-art systems under fine-tuning. This advantage spans tasks with fundamentally different demands---from spatially precise symbol detection, which requires dense per-symbol localization, to classification-oriented score difficulty estimation, which requires global semantic understanding---demonstrating that the learned representations are broadly useful rather than narrowly specialized. Underlying this performance gap is a more fundamental result: sheet music is a genuinely distinct visual domain that general-purpose models fail to represent adequately. Our embedding-transcription consistency analysis shows that general-purpose encoders produce representations that do not correlate with musical content, while \musvit{} encodes symbolic musical structure directly in its embedding space. Therefore, a domain-specific model is not merely beneficial for sheet music representation but necessary.


\section*{Acknowledgements}
The authors gratefully acknowledge Edward Guo, on behalf of IMSLP/Petrucci Music Library, for providing access to the data used to train the models.

This publication is part of the LEMUR project PID2023-148259NB-I00, funded by MICIU/AEI/10.13039/501100011033 and by ERDF/EU. The first author is supported by the University of Alicante through the FPU Program (UAFPU22-19). The third author is supported by a predoctoral contract associated with the LEMUR project. The fourth author is supported by a predoctoral contract from grant CISEJI/2023/9 ``Programa para el apoyo a personas investigadoras con talento (Plan GenT) de la Generalitat Valenciana''.

%
%
\bibliographystyle{abbrvnat}
\bibliography{main}

\clearpage
\appendix

\begin{appendix}

\setcounter{table}{10}  
\setcounter{figure}{4}  

\section{Supplementary Material}

\noindent We provide additional details and results in this supplementary material, organized as follows:
\begin{itemize}
    \item \textbf{Section~\ref{sup-sec:glossary}} — Terminology glossary for non-music readers.
    \item \textbf{Section~\ref{sup-sec:reconstructions}} — \musvit{} reconstruction examples.
    \item \textbf{Section~\ref{sup-sec:pretraining}} — Pre-training hyperparameters.
    \item \textbf{Section~\ref{sup-sec:curriculum_ablation}} — Two-stage curriculum ablation.
    \item \textbf{Section~\ref{sup-sec:architecture}} — Detailed architecture specifications for each \musvit{} variant.
    \item \textbf{Section~\ref{sup-sec:imslp}} — Additional IMSLP training data examples.
    \item \textbf{Section~\ref{sup-sec:datasets}} — Downstream task datasets and representative examples.
    \item \textbf{Section~\ref{sup-sec:encoders}} — General-purpose encoder model specifications for reproducibility.
    \item \textbf{Section~\ref{sup-sec:experiments}} — Fine-tuning protocols and per-dataset results for each downstream task.
    \item \textbf{Section~\ref{sup-sec:ft_general}} — Fine-tuning general-purpose models and computational cost.
    \item \textbf{Section~\ref{sup-sec:representation_analyses}} — Supplementary representation analyses.
\end{itemize}

\newpage

\subsection{Terminology Glossary}
\label{sup-sec:glossary}

Table~\ref{tab:glossary} provides definitions of key music notation and document analysis terms used throughout the paper, intended to assist readers less familiar with the music domain.

\begin{table}[ht]
    \centering
    \setlength{\abovecaptionskip}{1pt}
    \setlength{\belowcaptionskip}{1pt}
    \caption{Glossary of music notation and document analysis terminology.}
    \label{tab:glossary}
    \setlength{\tabcolsep}{6pt}
    \renewcommand{\arraystretch}{1.4}
    \resizebox{\columnwidth}{!}{%
        \begin{tabular}{lp{10cm}}
            \toprule[1pt]
            \textbf{Term} & \textbf{Definition} \\ \cmidrule(lr){1-2}
            Staff & A set of five horizontal lines on which musical notes are placed; vertical position on the staff encodes pitch. \\
            Clef & A symbol placed at the beginning of a staff that defines the pitch range (e.g., treble clef, bass clef). \\
            Notehead & The oval part of a musical note; its position on the staff indicates pitch. \\
            Stem & The vertical line attached to a notehead, used in combination with the notehead shape to indicate note duration. \\
            Beam & A horizontal or slanted bar connecting stems of consecutive notes, indicating rhythmic grouping. \\
            Accidental & A symbol (\(\sharp\), \(\flat\), \(\natural\)) that modifies the pitch of a note. \\
            Rest & A symbol indicating a period of silence of a specified duration. \\
            Measure (bar) & A segment of music bounded by vertical bar lines, containing a fixed number of beats. \\
            CWMN & Common Western Music Notation---the standard staff-based notation system used in Western music since approximately the 17th century. \\
            Mensural notation & A predecessor of CWMN used roughly from the 13th to the 16th century, with different conventions for rhythm and note shapes. \\
            Pianoform & A score layout for keyboard instruments using two staves (treble and bass) connected by a brace. \\
            OMR & Optical Music Recognition---the task of automatically converting images of music scores into machine-readable symbolic formats. \\
            SER & Symbol Error Rate---the normalized edit distance between predicted and ground-truth symbol sequences; lower is better. \\
            \bottomrule[1pt]
        \end{tabular}
    }
\end{table}

\subsection{MuSViT Reconstruction Examples}
\label{sup-sec:reconstructions}

\begin{figure}[ht!]
  \centering
  \begin{subfigure}{0.32\linewidth}
    \centering
    \includegraphics[width=\linewidth]{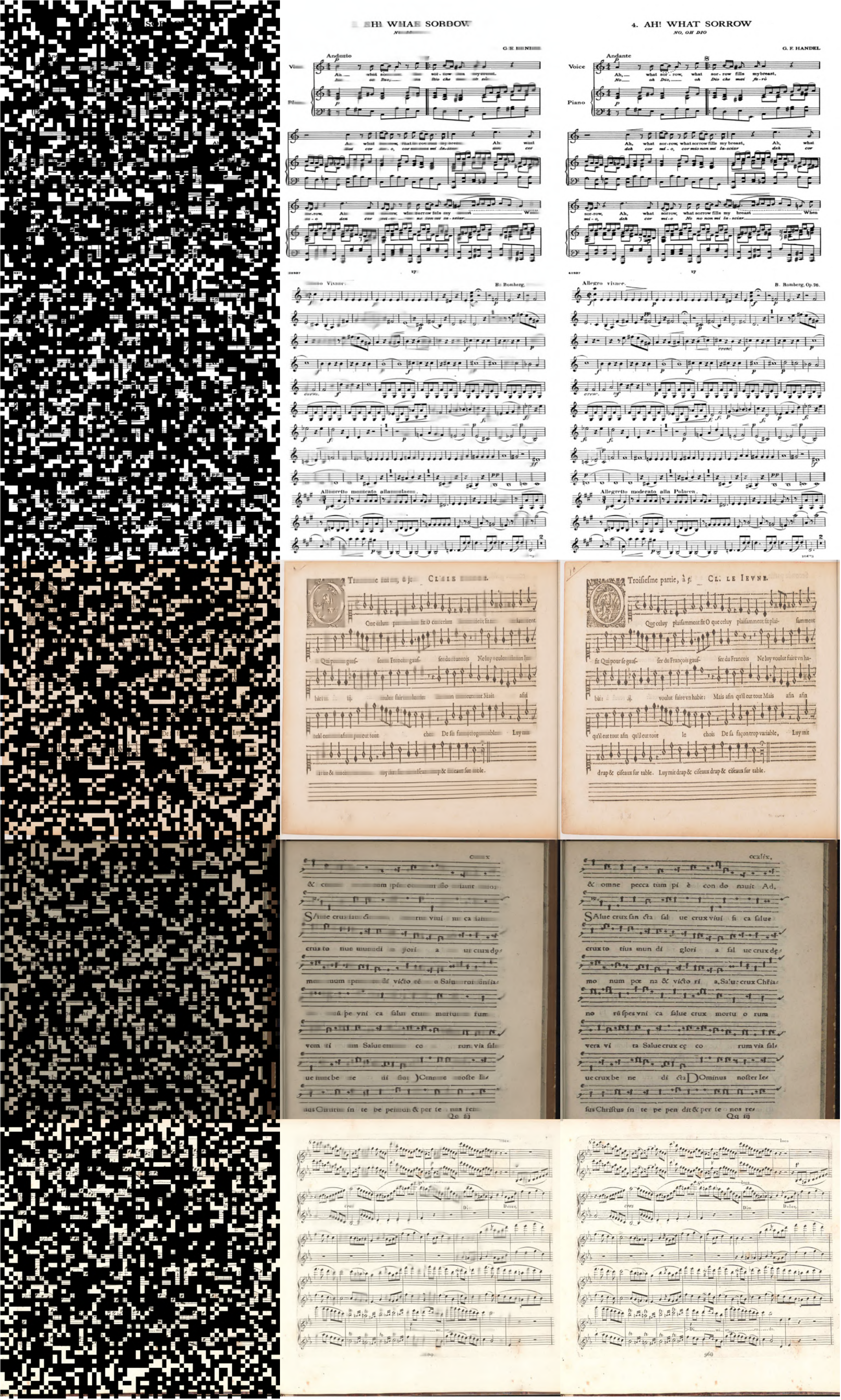}
  \end{subfigure}
  \hfill
  \begin{subfigure}{0.32\linewidth}
    \centering
    \includegraphics[width=\linewidth]{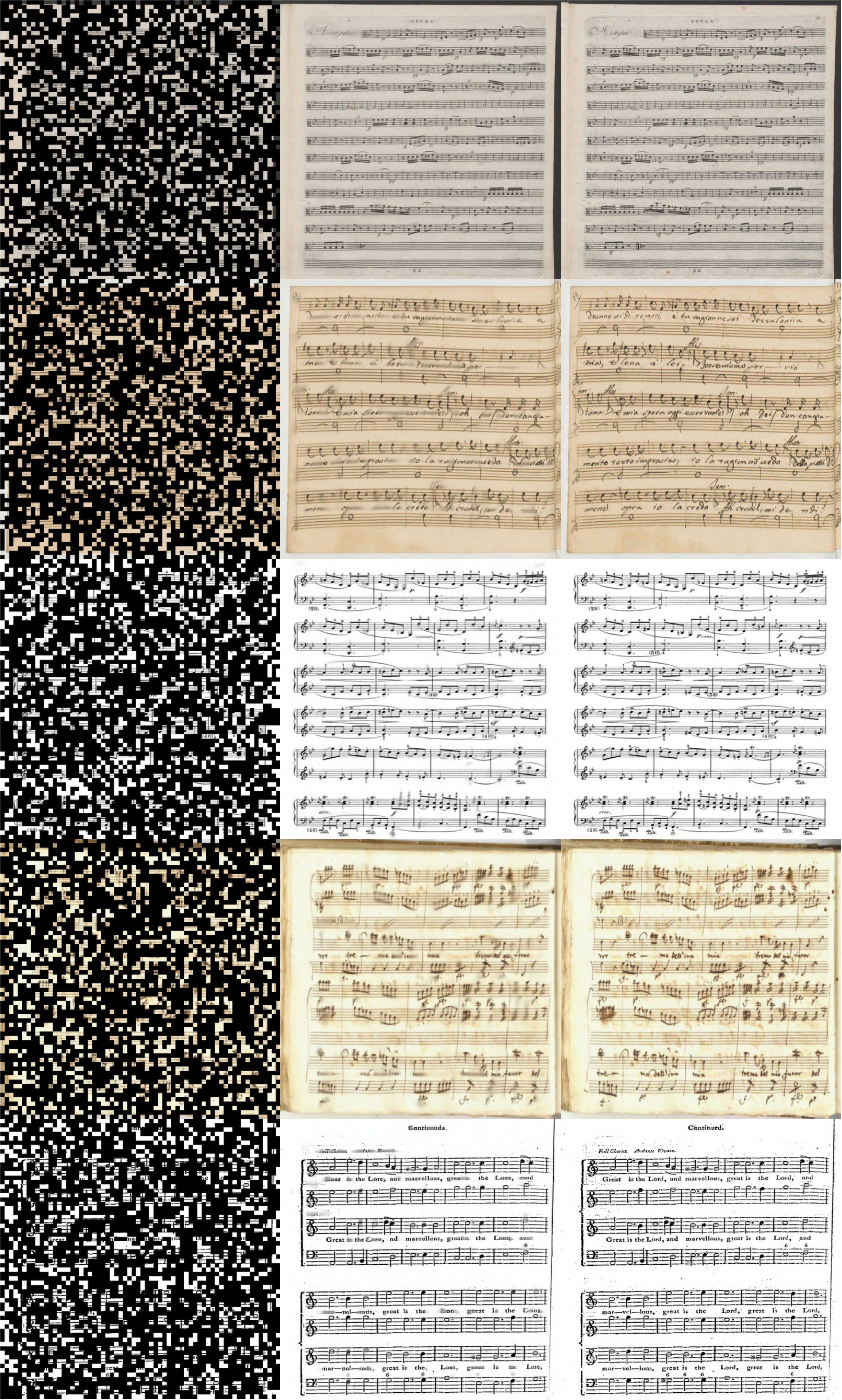}
  \end{subfigure}
  \hfill
  \begin{subfigure}{0.32\linewidth}
    \centering
    \includegraphics[width=\linewidth]{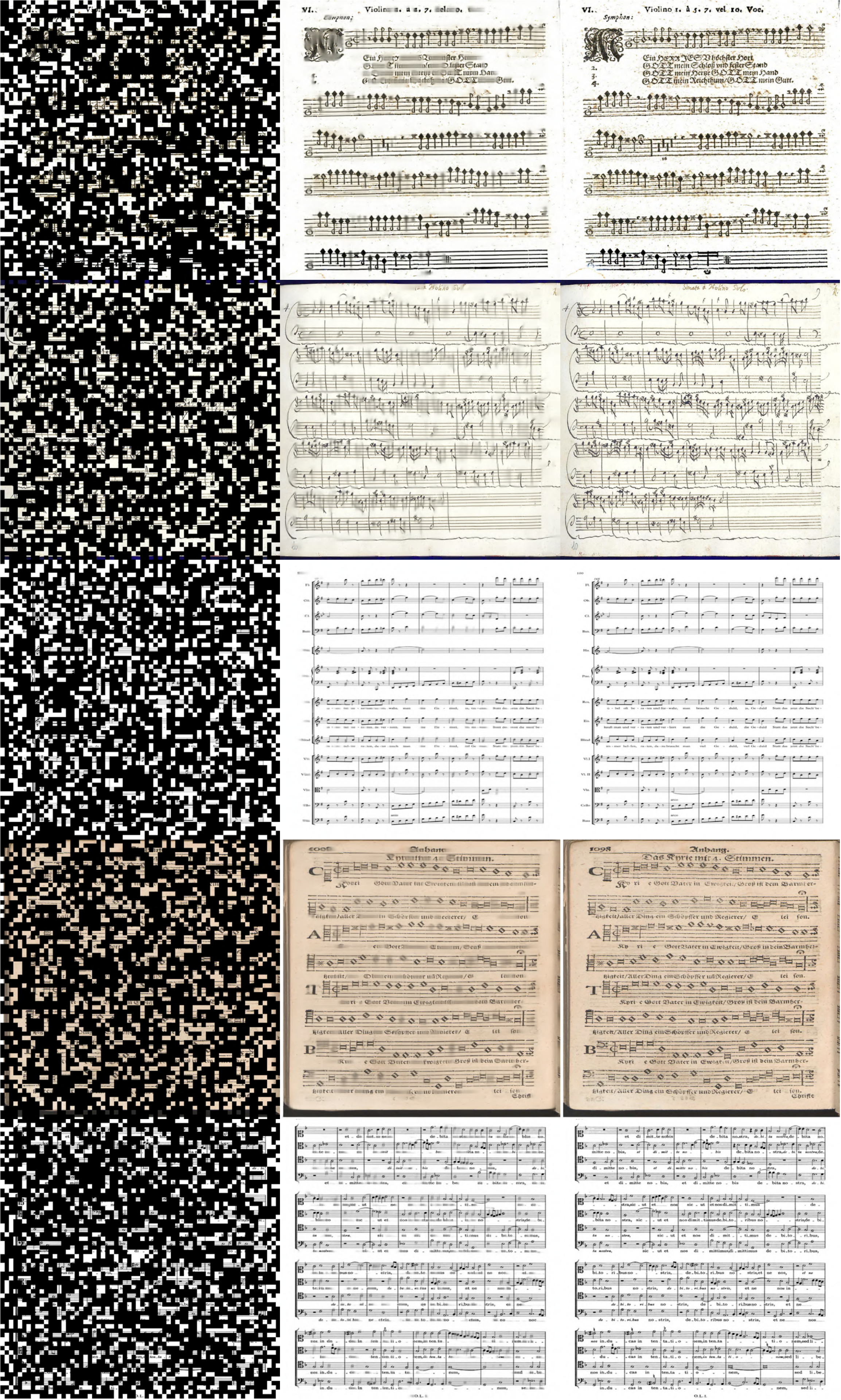}
  \end{subfigure}
  \caption{\musvit{} reconstruction examples. Each example row shows the masked input with 70\% of patches removed (left), the \musvit{} reconstruction (middle), and the original sheet music region (right).}
  \label{fig:reconstructions}
\end{figure}

Figure~\ref{fig:reconstructions} shows additional qualitative reconstruction examples produced by \musvit{}. Each panel shows three images: the masked input (70\% of patches removed), the \musvit{} reconstruction, and the original score region. Despite the heavy masking that occludes entire measures, \musvit{} accurately recovers fine-grained notational elements---noteheads, stems, accidentals, and staff lines---demonstrating that the encoder has internalized the visual grammar of music notation. This qualitative evidence supports the hypothesis that the pre-training objective forces learning of genuine musical structure rather than low-level texture statistics.

\subsection{Pre-Training Details}
\label{sup-sec:pretraining}

Pre-training follows a two-stage curriculum. Stage~1 uses synthetic crops from \DeepScores{}~\cite{Tuggener:DeepScoresv2:2020} at $512\times512$ resolution (1,024 patches per image, masking ratio 50\%), serving as a structured warm-up before exposure to real-world data. Stage~2 uses full pages from \ac{IMSLP} at $1024\times1024$ resolution (4,096 patches per image, masking ratio 70\%), adapting the model to the full complexity of scanned music documents. The higher masking ratio in Stage~2 forces the encoder to reason over long-range musical context, occluding entire measures and requiring inference of pitch, duration, and sequential structure from sparse visible regions. Table~\ref{tab:pretraining} lists the hyperparameters for each stage.

\begin{table}[ht!]
    \centering
    \caption{Pre-training hyperparameters for each curriculum stage.}
    \label{tab:pretraining}
    \setlength{\tabcolsep}{12pt}
    \renewcommand{\arraystretch}{1.3}
    \begin{tabular}{lcc}
        \toprule[1.5pt]
        \textbf{Hyperparameter} & \textbf{Stage 1 (Synthetic)} & \textbf{Stage 2 (Real-world)} \\ \midrule
        Dataset                 & \DeepScores{}                 & \ac{IMSLP} \\
        Input resolution        & $512 \times 512$              & $1024 \times 1024$ \\
        Patch size $P$          & 16                            & 16 \\
        Patches per image $N$   & 1,024                         & 4,096 \\
        Masking ratio           & 50\%                          & 70\% \\ \cmidrule(lr){1-3}
        Optimizer               & AdamW                 & AdamW \\
        Base Learning rate           & 0.0001                 & 0.00015 \\
        Learning rate Scheduler & None  & Cosine    \\
        Warm-up epochs          & 0                 & 0.1 \\
        Batch size              & 128                 & 70 \\
        Epochs                  & 30                 & 4 \\
        Weight decay            & 0                 & 0.001 \\
        \bottomrule[1.5pt]
    \end{tabular}
\end{table}


\subsection{Two-Stage Curriculum Ablation}
\label{sup-sec:curriculum_ablation}

\begin{figure}[ht]
  \centering
  \begin{subfigure}{0.49\linewidth}
    \centering
    \includegraphics[width=\linewidth]{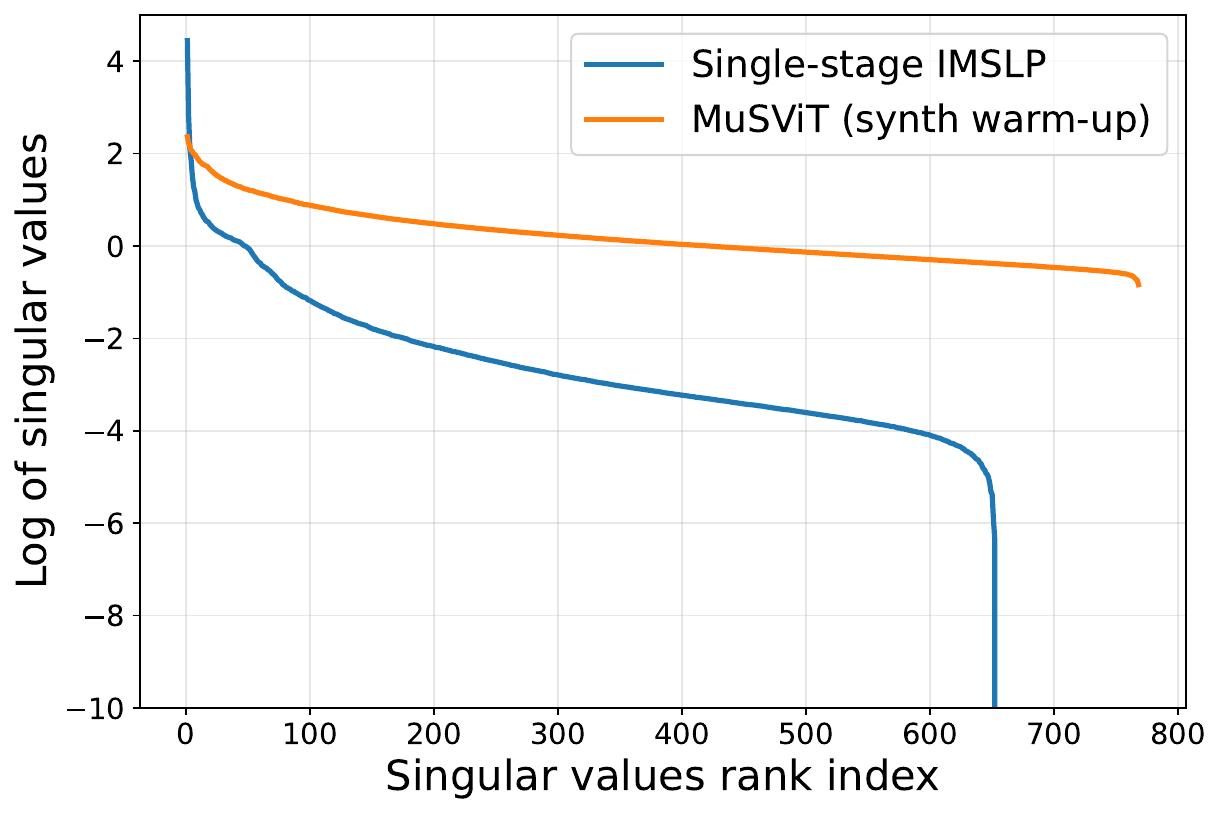}
    \caption{Singular value spectrum}
    \label{fig:singular_values}
  \end{subfigure}
  \hfill
  \begin{subfigure}{0.49\linewidth}
    \centering
    \includegraphics[width=\linewidth]{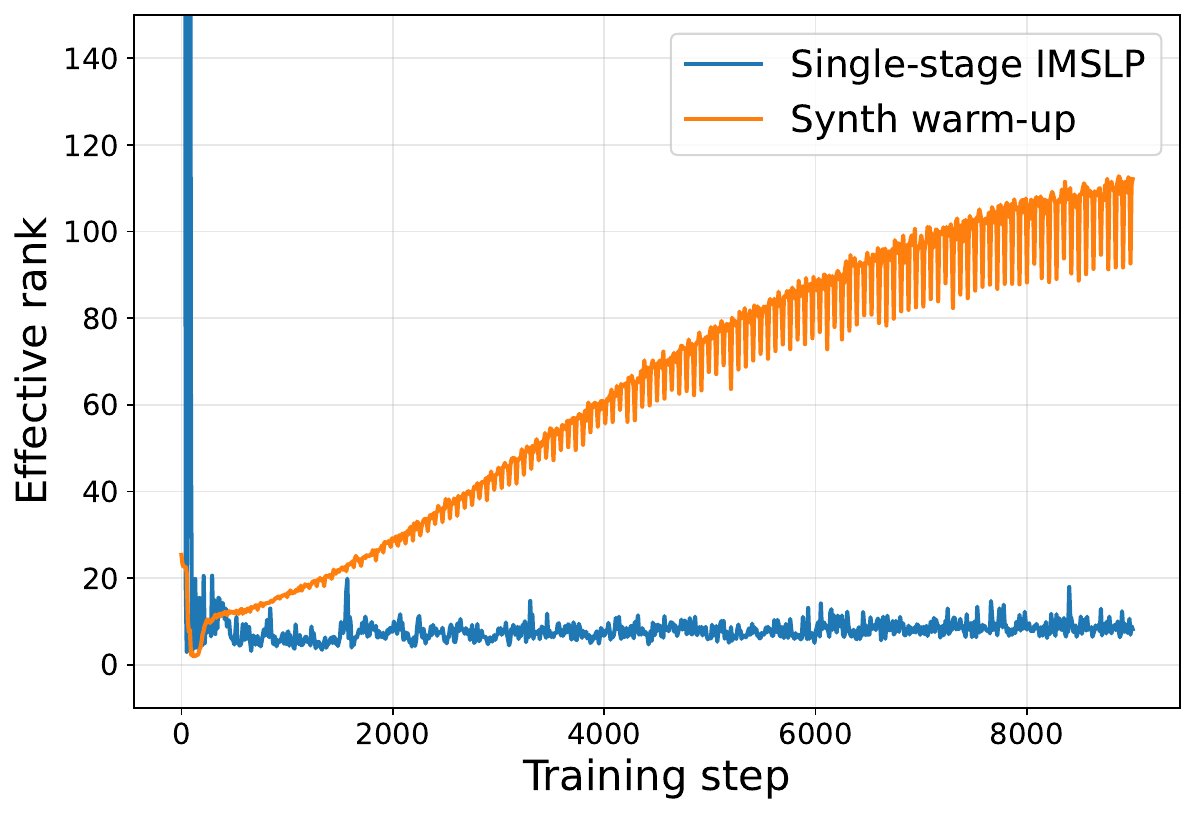}
    \caption{Effective rank over training}
    \label{fig:effective_rank}
  \end{subfigure}
  \caption{Comparison between single-stage IMSLP training and the proposed two-stage curriculum. Single-stage training exhibits dimensional collapse, whereas the curriculum maintains a flatter spectrum and higher effective rank.}
  \label{fig:two_stage}
\end{figure}

To validate the necessity of the two-stage curriculum described in Section~2.1 of the main paper, we compare it against single-stage MAE training directly on IMSLP. As shown in Fig.~\ref{fig:two_stage}, single-stage training leads to \emph{dimensional collapse}~\cite{jing2021understanding,zhang2022mask,kim2025taxonomy}: despite reducing reconstruction loss, the encoder concentrates most variance in a few dimensions, and its effective rank drops early during training. The decoder resorts to predicting average patches rather than music-notation structures. In contrast, the proposed curriculum maintains a substantially flatter singular value spectrum and higher effective rank throughout training, indicating better use of the 768-dimensional embedding space. This confirms that real-world sheet music is too heterogeneous for MAE to converge stably from scratch without first learning elementary notation structure on synthetic data.

\subsection{Architecture Details}
\label{sup-sec:architecture}

Table~\ref{tab:architecture} summarizes the architectural configurations of \musvit{} and \musvitl{}. Both variants follow the standard ViT design~\cite{Dosovitskiy:ICLR:2021}. The key difference is the embedding dimension: \musvit{} uses $d=768$ while \musvitl{} uses $d=384$, reducing the parameter count from approximately 85M to 25M. Both models use the same patch size ($P=16$) and number of layers (12), maintaining comparable depth while reducing width.

\begin{table}[ht!]
    \centering
    \caption{Architecture configurations of \musvit{} and \musvitl{}.}
    \label{tab:architecture}
    \setlength{\tabcolsep}{12pt} 
    \renewcommand{\arraystretch}{1.3}
    \begin{tabular}{lcc}
        \toprule[1.5pt]
        \textbf{Hyperparameter}     & \musvit{}         & \musvitl{} \\ \midrule
        Patch size $P$              & 16                & 16 \\
        Embedding dim $d$           & 768               & 384 \\
        Transformer layers          & 12                & 12 \\
        Attention heads             & 12                & 6 \\
        Positional encoding         & 2D Sinusoidal     & 2D Sinusoidal \\
        Encoder parameters          & $\sim$85M         & $\sim$25M \\ \cmidrule(lr){1-3}
        Decoder layers              & 8                 & 8\\
        Decoder dim                 & 512               & 384 \\
        Decoder parameters          & $\sim$16M         & $\sim$8M \\
        \bottomrule[1.5pt]
    \end{tabular}
\end{table}

\newpage

\subsection{IMSLP Training Data}
\label{sup-sec:imslp}

\begin{figure}[t]
    \centering
    \includegraphics[width=1.0\linewidth]{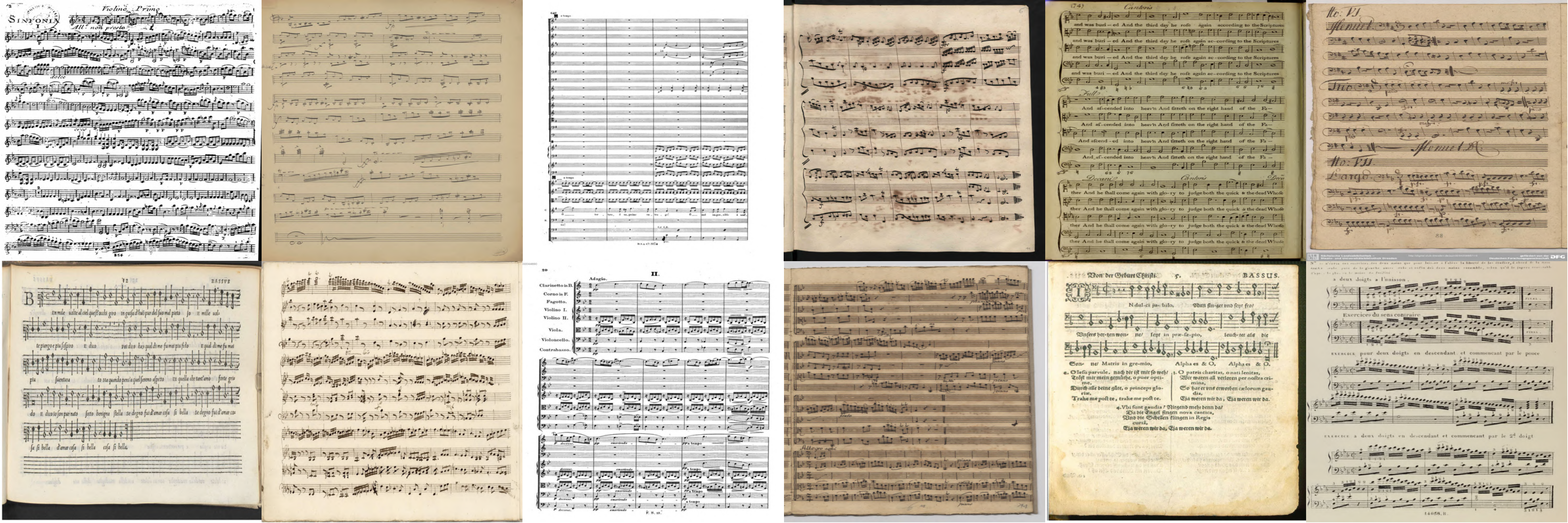}
    \includegraphics[width=1.0\linewidth]{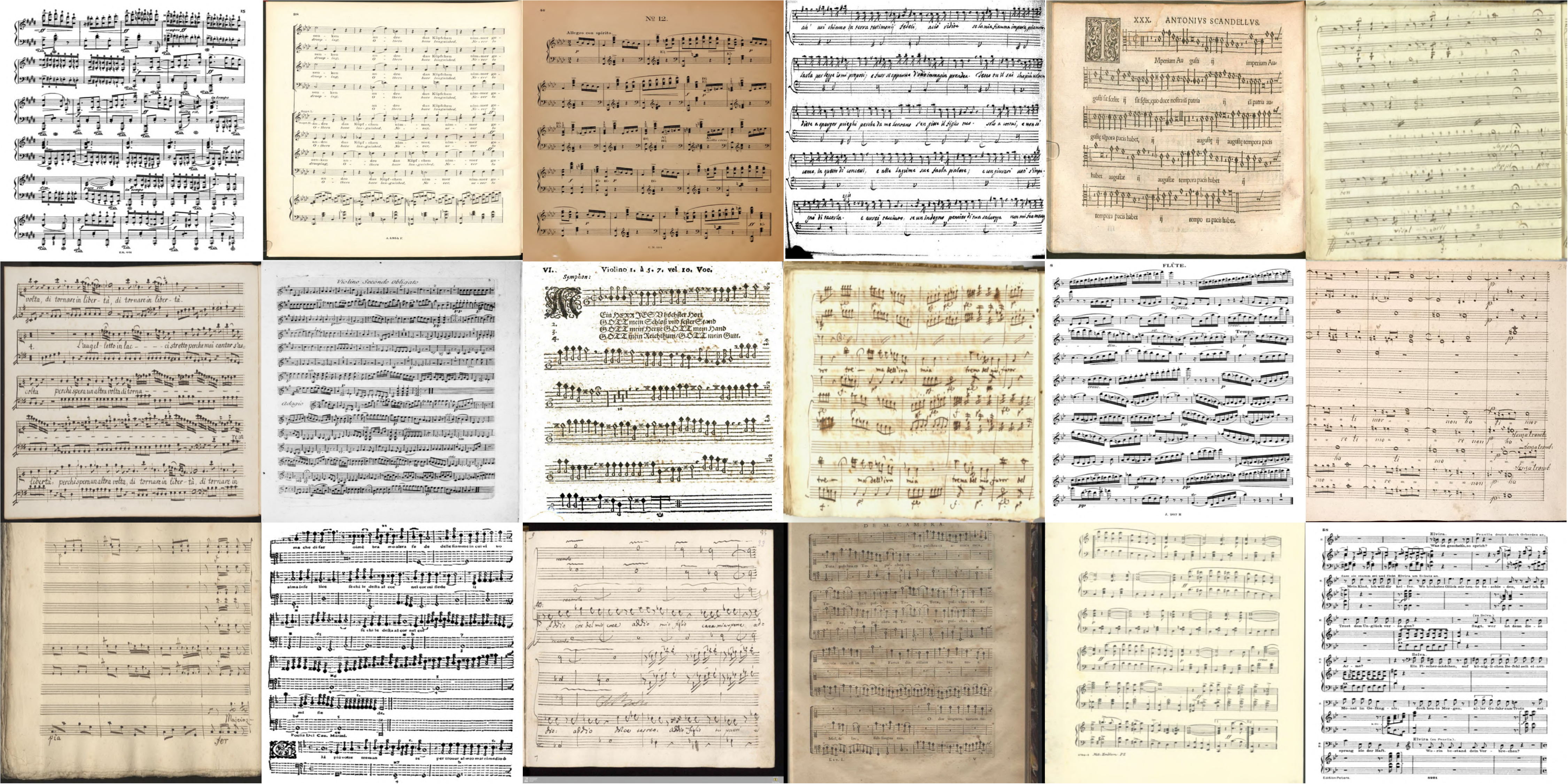}
    \caption{Representative pages from the \ac{IMSLP} pre-training corpus. The collection spans historical periods, notation systems (mensural, CWMN), engraving styles (handwritten, typeset), and musical textures (monophonic, polyphonic).}
    \label{fig:imslp_examples_supp}
\end{figure}

Figure~\ref{fig:imslp_examples_supp} shows representative pages drawn from the \ac{IMSLP} corpus, illustrating the breadth of visual variability in the collection.

\newpage

\subsection{Downstream Task Datasets}
\label{sup-sec:datasets}

\begin{figure}[ht!]
  \centering
  \begin{subfigure}{\linewidth}
    \centering
    \begin{subfigure}{0.24\linewidth}
      \includegraphics[width=\linewidth]{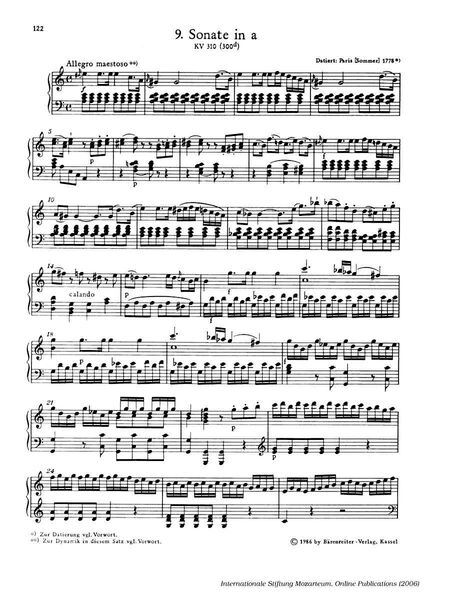}
    \end{subfigure}
    \hfill
    \begin{subfigure}{0.24\linewidth}
      \includegraphics[width=\linewidth]{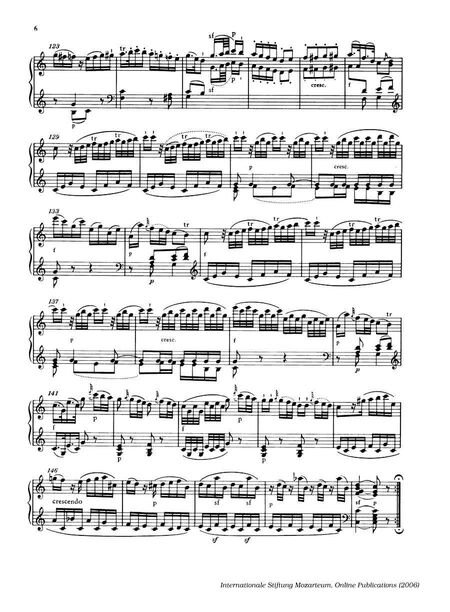}
    \end{subfigure}
    \hfill
    \begin{subfigure}{0.24\linewidth}
      \includegraphics[width=\linewidth]{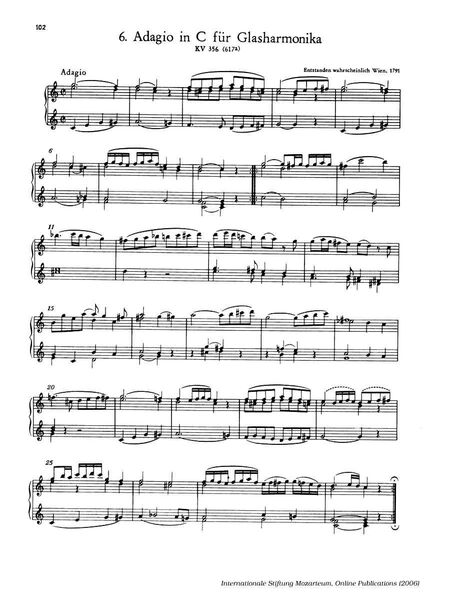}
    \end{subfigure}
    \hfill
    \begin{subfigure}{0.24\linewidth}
      \includegraphics[width=\linewidth]{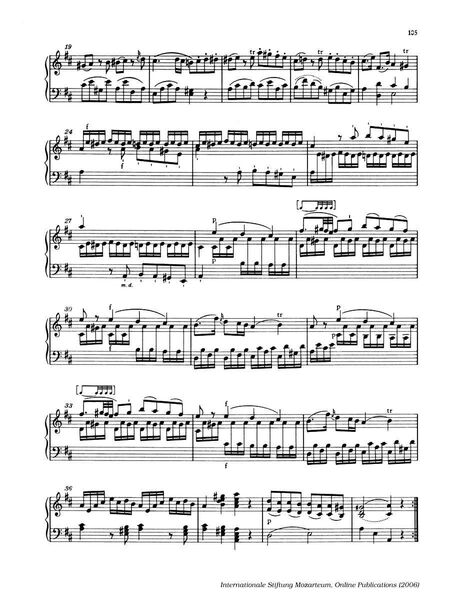}
    \end{subfigure}
    \caption{\Mozarteum{} (101 pages, printed CWMN)}
    \label{fig:dataset_mozarteum}
  \end{subfigure}
  \begin{subfigure}{\linewidth}
    \centering
    \begin{subfigure}{0.24\linewidth}
      \includegraphics[width=\linewidth]{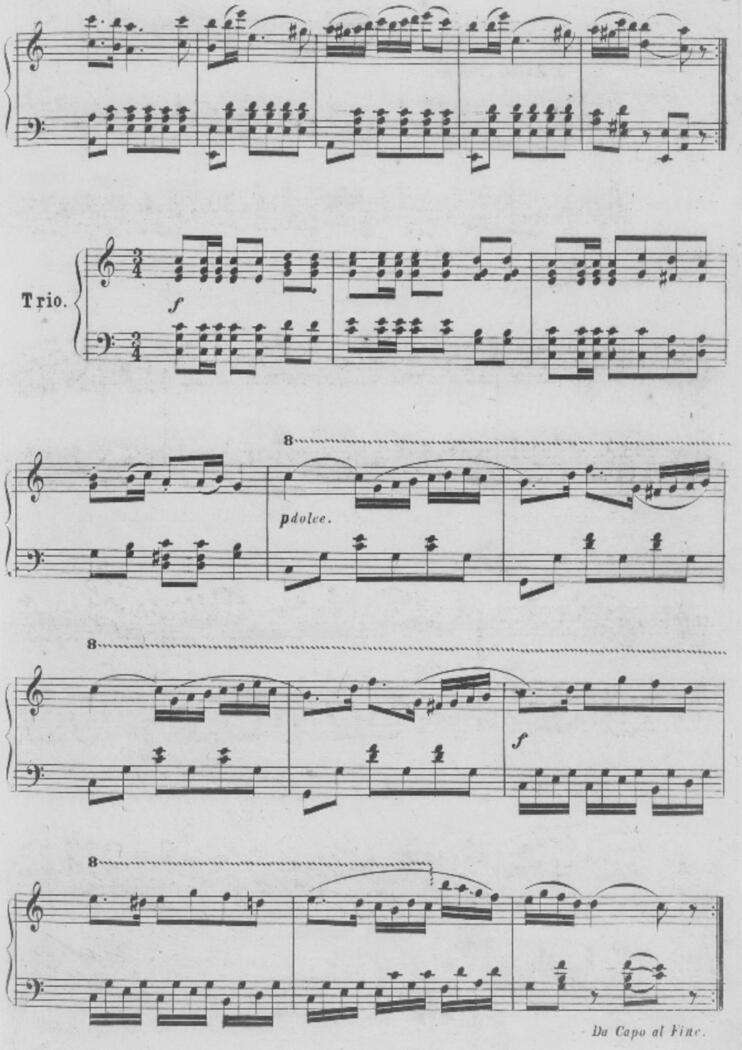}
    \end{subfigure}
    \hfill
    \begin{subfigure}{0.24\linewidth}
      \includegraphics[width=\linewidth]{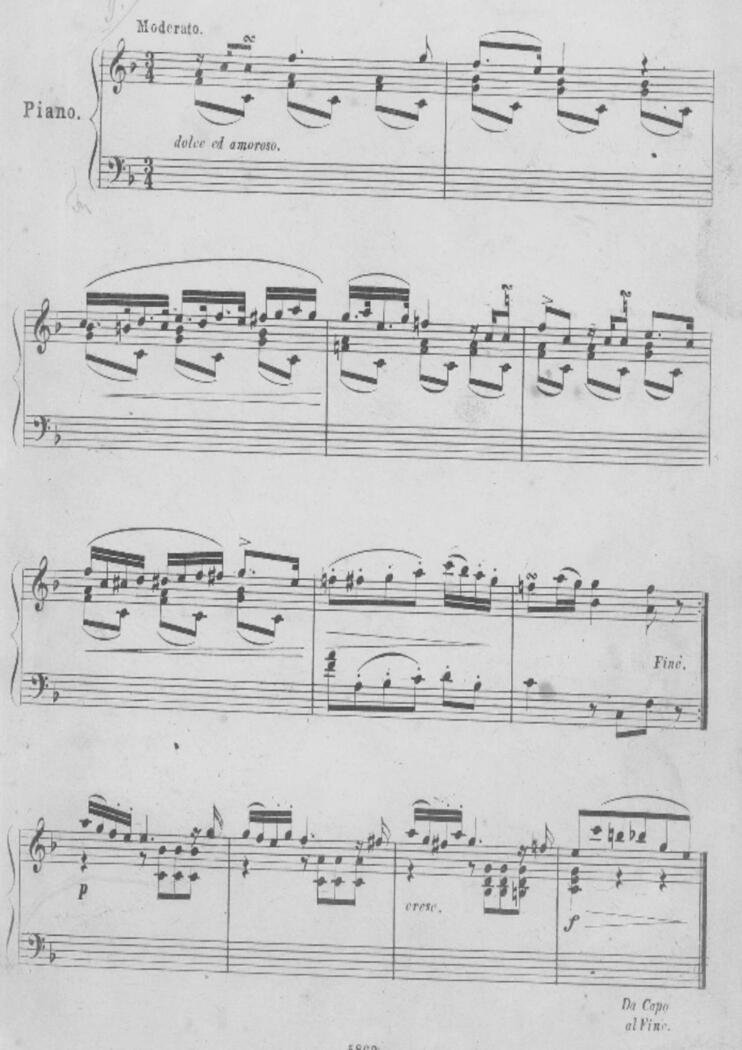}
    \end{subfigure}
    \hfill
    \begin{subfigure}{0.24\linewidth}
      \includegraphics[width=\linewidth]{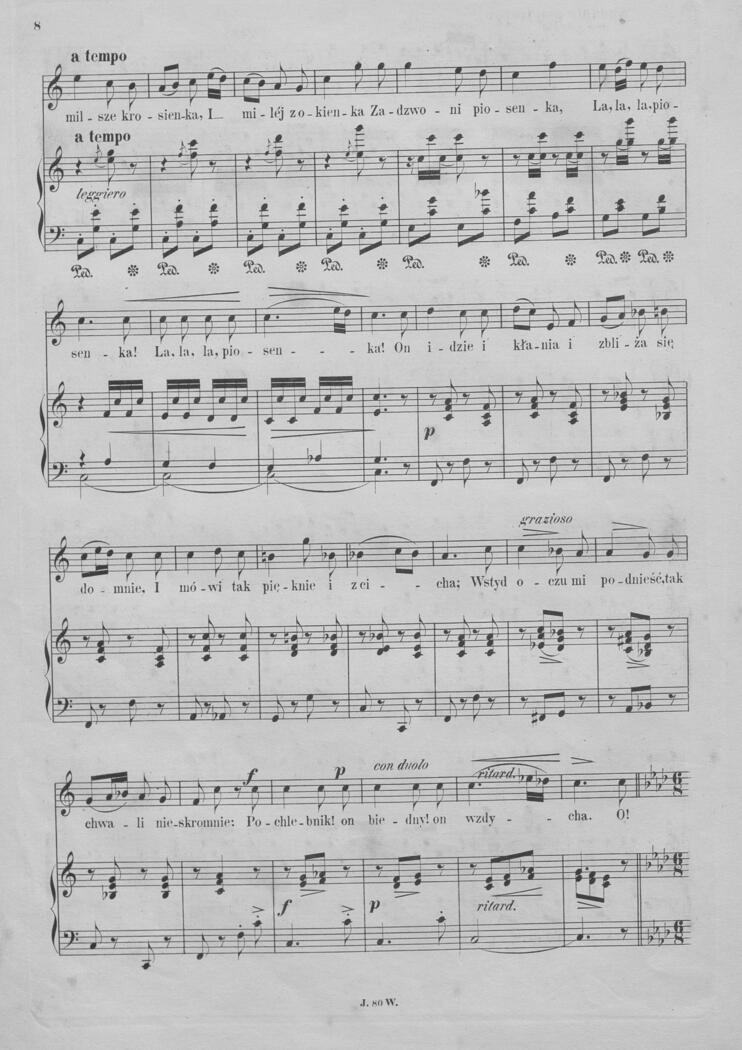}
    \end{subfigure}
    \hfill
    \begin{subfigure}{0.24\linewidth}
      \includegraphics[width=\linewidth]{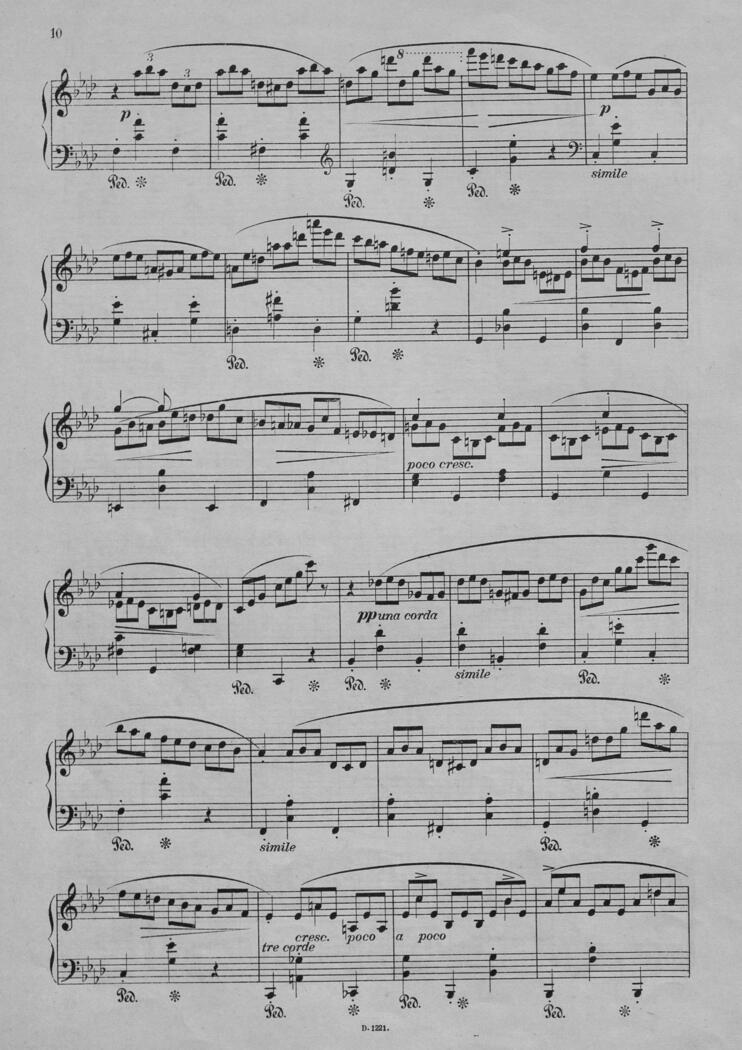}
    \end{subfigure}
    \caption{\Polish{} (117 pages, printed CWMN)}
    \label{fig:dataset_polish}
  \end{subfigure}
  \caption{Representative pages from the two full-page recognition datasets.}
  \label{fig:datasets_full_page}
\end{figure}

\begin{figure}[ht!]
  \centering
  \begin{subfigure}{\linewidth}
    \centering
    \includegraphics[width=0.55\textwidth]{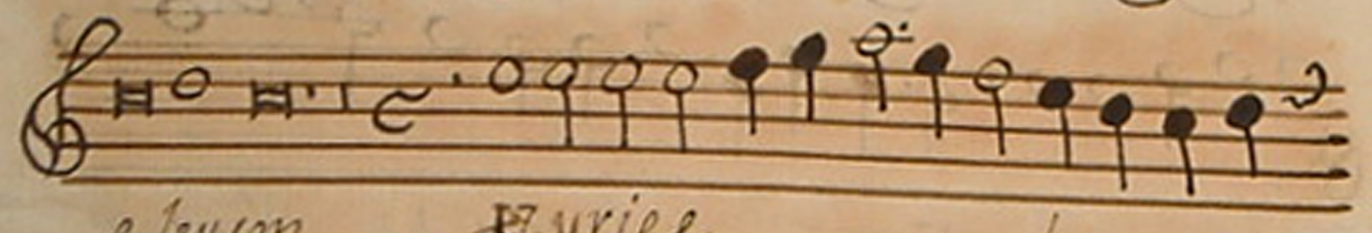}
    \caption{\Capitan{} (828 staves, handwritten mensural)}
    \label{fig:capitan}
  \end{subfigure}
  \begin{subfigure}{\linewidth}
    \centering
    \includegraphics[width=0.6\textwidth]{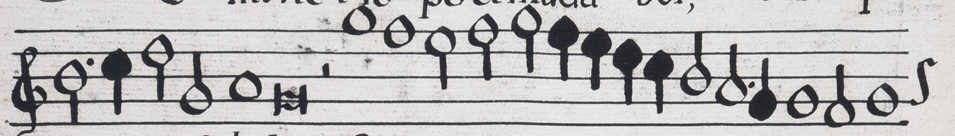}
    \caption{\Guatemala{} (3,263 staves, handwritten mensural)}
    \label{fig:guatemala}
  \end{subfigure}
  \begin{subfigure}{\linewidth}
    \centering
    \includegraphics[width=0.65\textwidth]{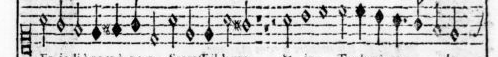}
    \caption{\ILS{} (1,136 staves, printed mensural)}
    \label{fig:ils}
  \end{subfigure}
  \begin{subfigure}{\linewidth}
    \centering
    \includegraphics[width=0.6\textwidth]{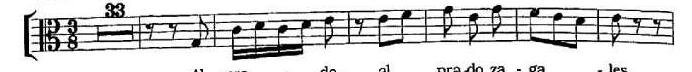}
    \caption{\Malaga{} (308 staves, printed CWMN)}
    \label{fig:malaga}
  \end{subfigure}
  \begin{subfigure}{\linewidth}
    \centering
    \includegraphics[width=0.6\textwidth]{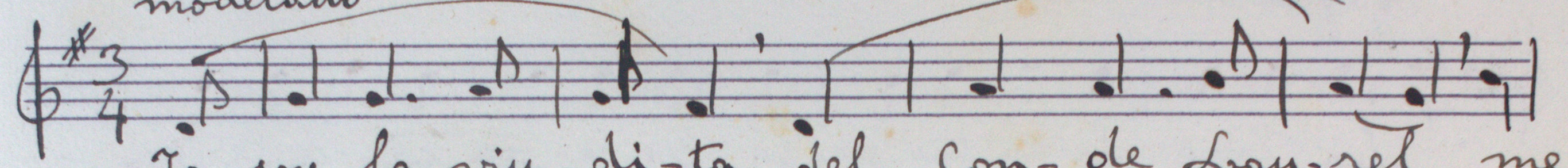}
    \caption{\FMT{} (1,305 staves, handwritten CWMN)}
    \label{fig:fmt}
  \end{subfigure}
  \caption{Representative staff images from the five staff-level recognition corpora.}
  \label{fig:datasets_staff}
\end{figure}

\begin{figure}[ht!]
  \centering
  \begin{subfigure}{0.24\linewidth}
    \centering
    \includegraphics[width=\linewidth]{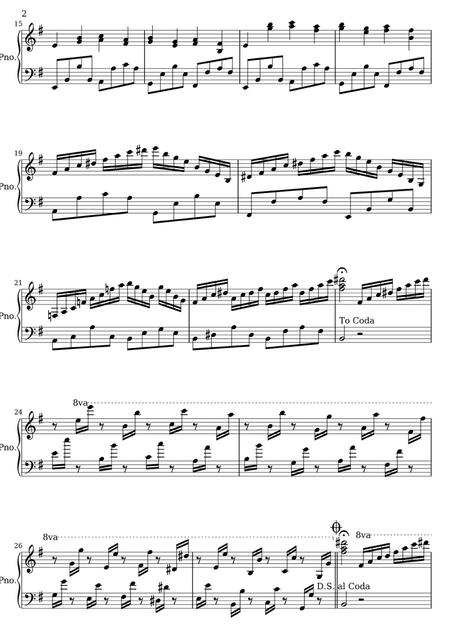}
  \end{subfigure}
  \hfill
  \begin{subfigure}{0.24\linewidth}
    \centering
    \includegraphics[width=\linewidth]{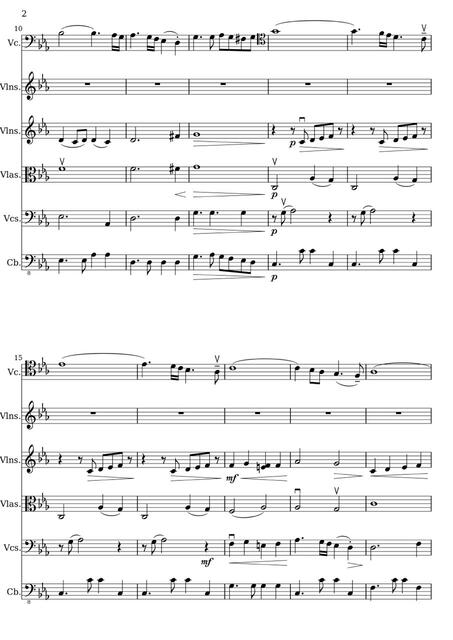}
  \end{subfigure}
  \hfill
  \begin{subfigure}{0.24\linewidth}
    \centering
    \includegraphics[width=\linewidth]{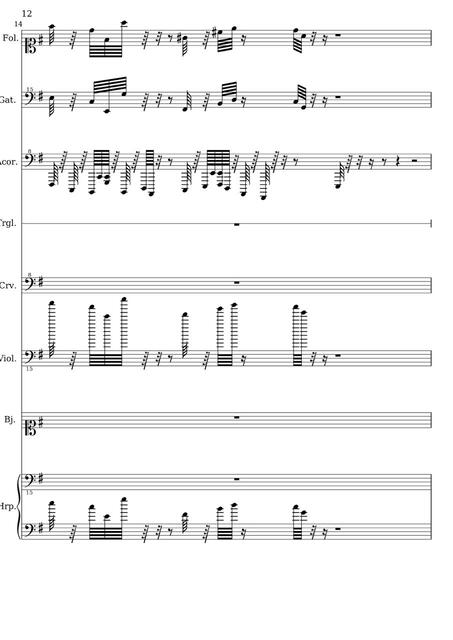}
  \end{subfigure}
  \hfill
  \begin{subfigure}{0.24\linewidth}
    \centering
    \includegraphics[width=\linewidth]{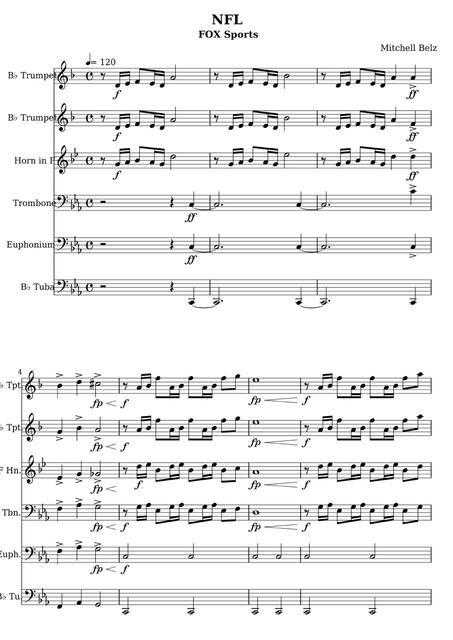}
  \end{subfigure}
  \caption{Representative annotated score pages from \DeepScores{} (1,714 scores, 135 symbol classes).}
  \label{fig:datasets_detection}
\end{figure}

\begin{figure}[ht!]
  \centering
  \begin{subfigure}{\linewidth}
    \centering
    \begin{subfigure}{0.24\linewidth}
      \includegraphics[width=\linewidth]{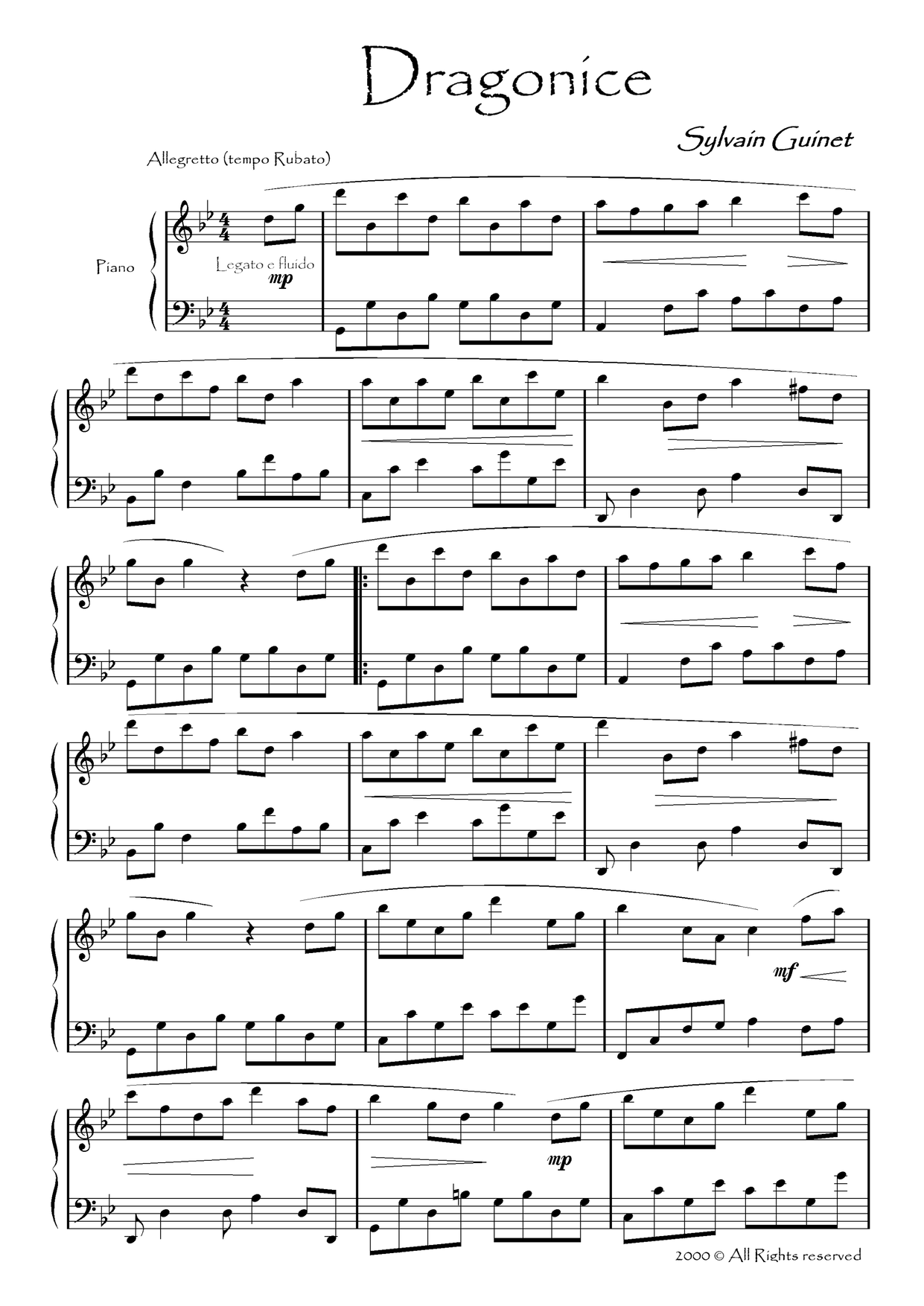}
    \end{subfigure}
    \hfill
    \begin{subfigure}{0.24\linewidth}
      \includegraphics[width=\linewidth]{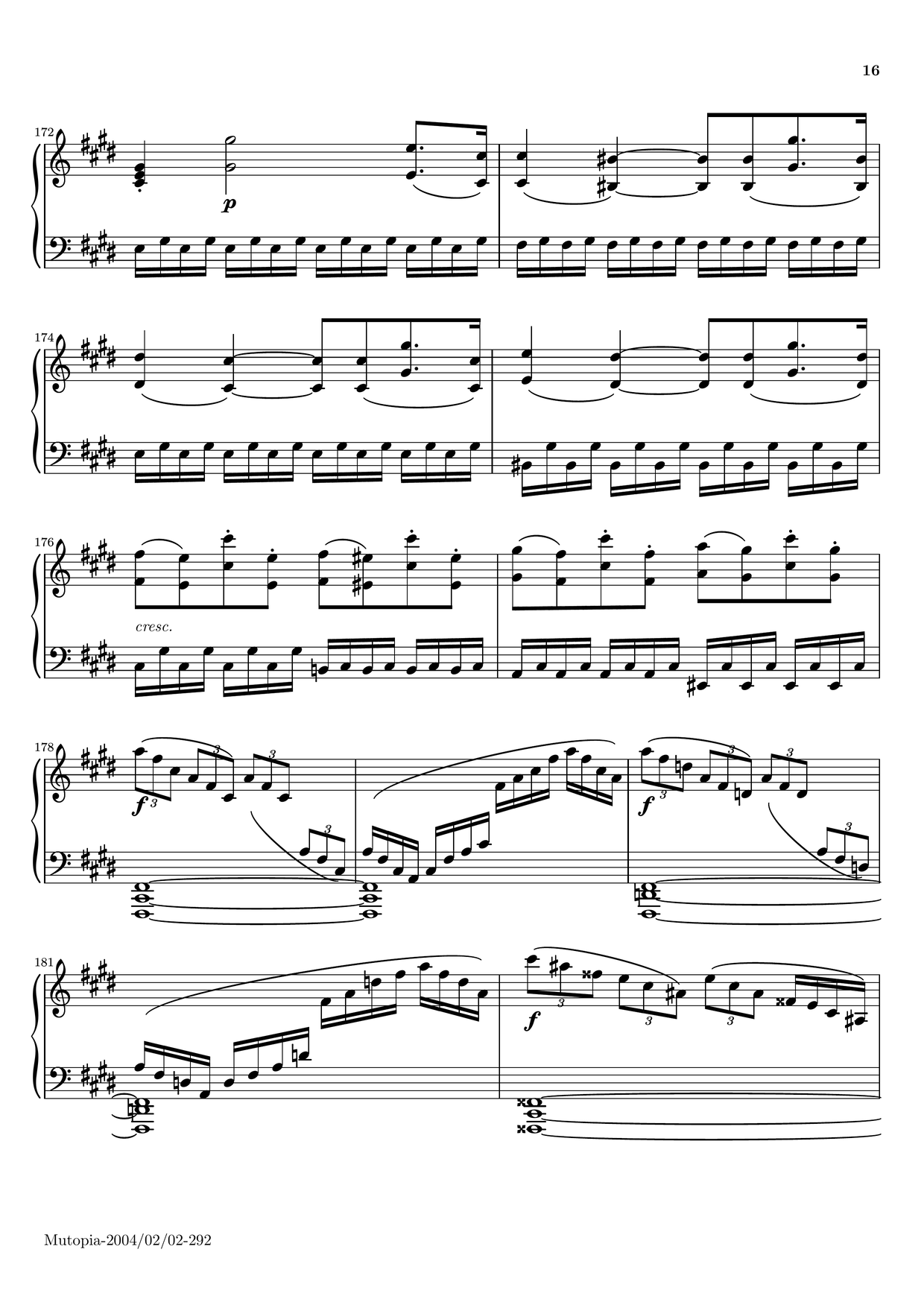}
    \end{subfigure}
    \hfill
    \begin{subfigure}{0.24\linewidth}
      \includegraphics[width=\linewidth]{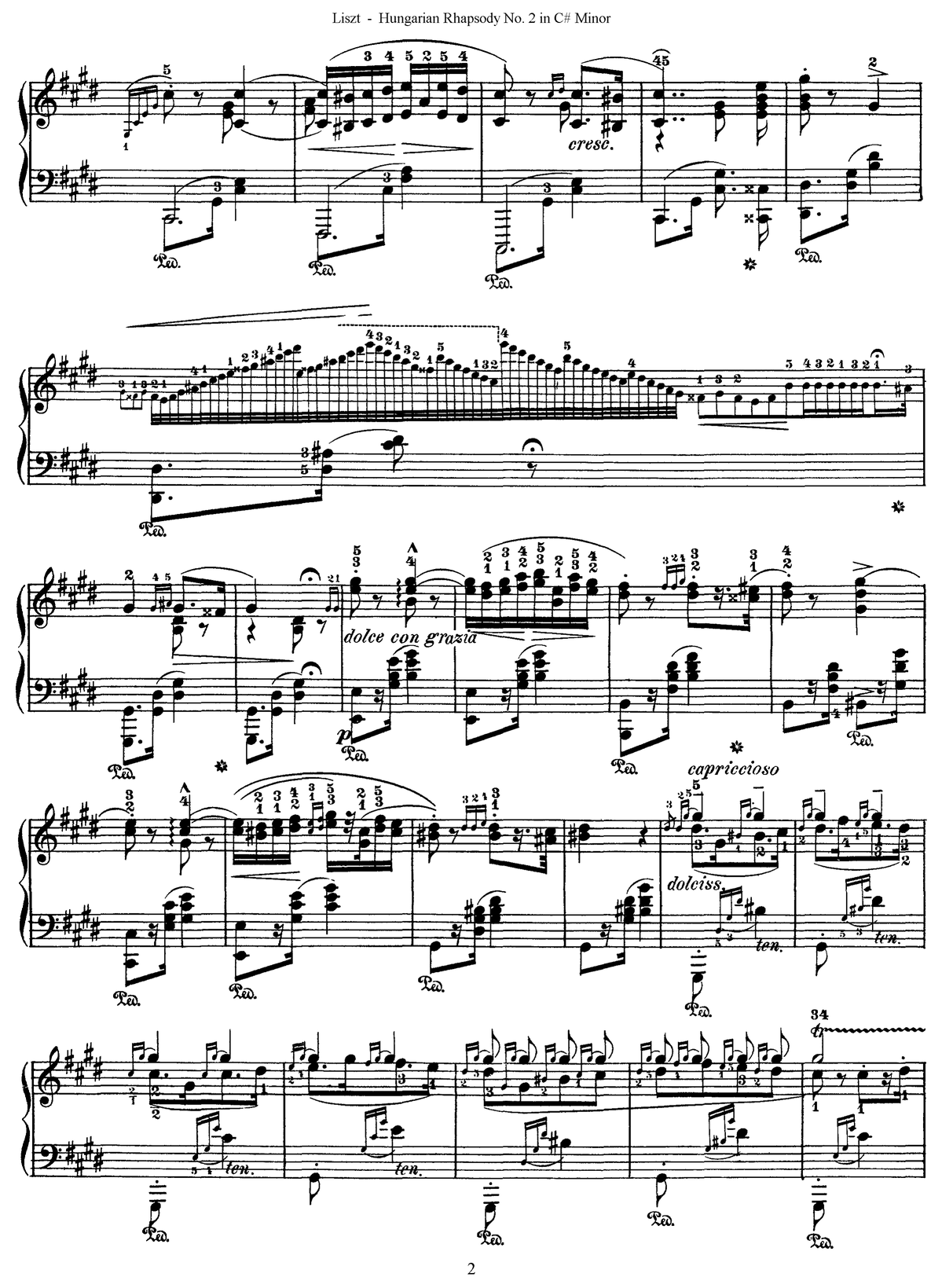}
    \end{subfigure}
    \hfill
    \begin{subfigure}{0.24\linewidth}
      \includegraphics[width=\linewidth]{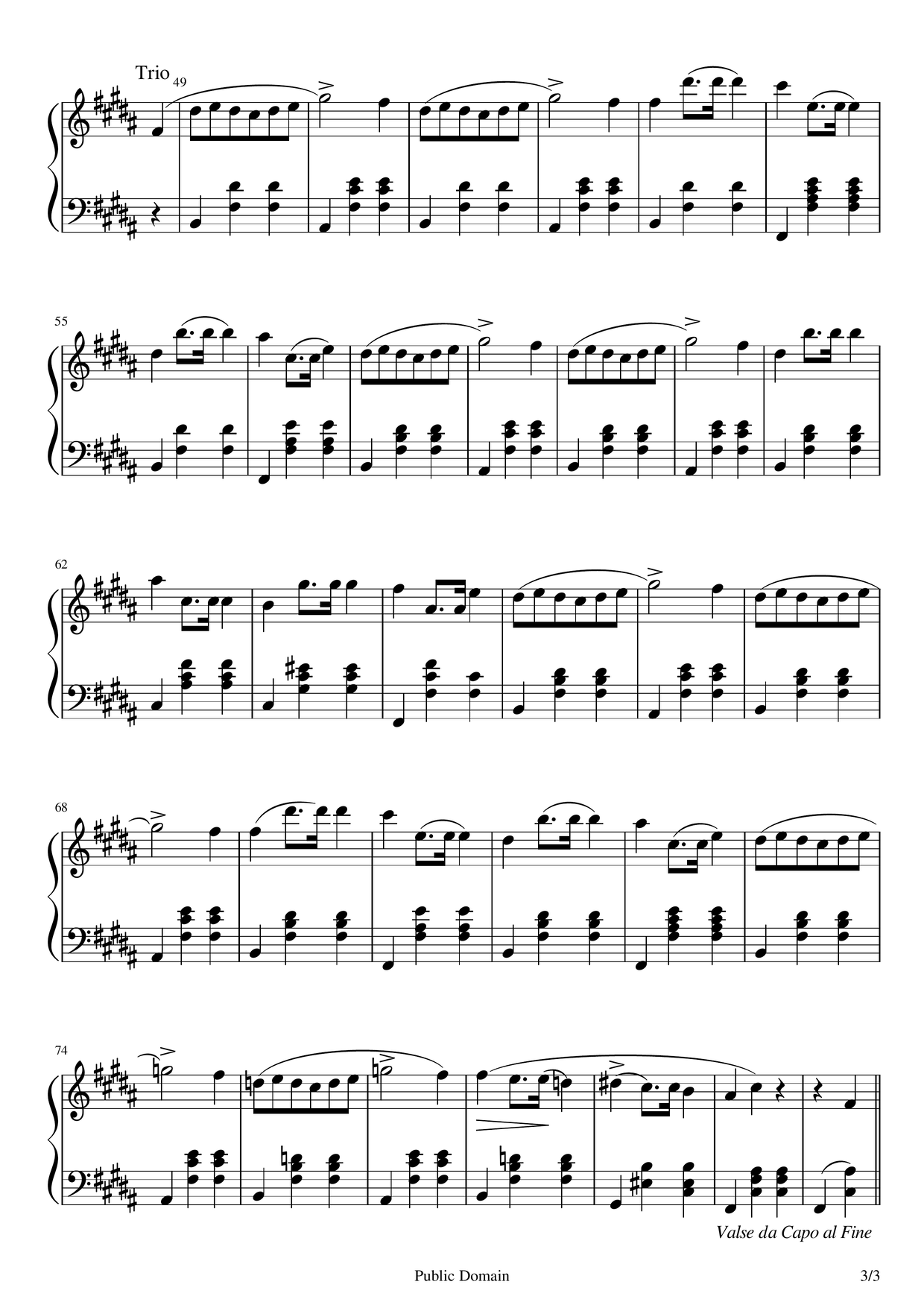}
    \end{subfigure}
    \hfill
    \caption{\FreeScores{} (4,193 pieces, 5 difficulty levels)}
    \label{fig:dataset_freescores}
  \end{subfigure}
  \begin{subfigure}{\linewidth}
    \centering
    \begin{subfigure}{0.24\linewidth}
      \includegraphics[width=\linewidth]{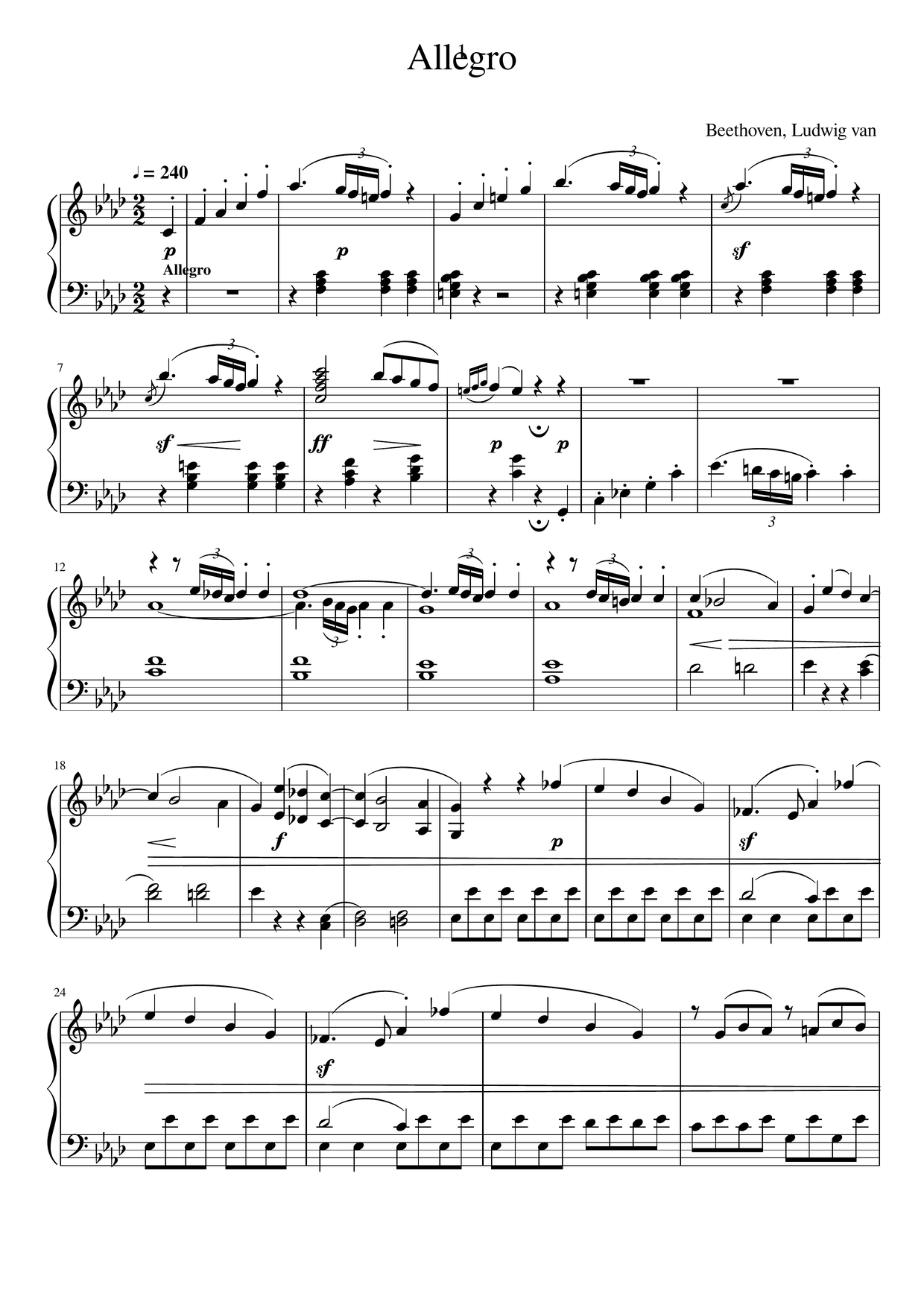}
    \end{subfigure}
    \hfill
    \begin{subfigure}{0.24\linewidth}
      \includegraphics[width=\linewidth]{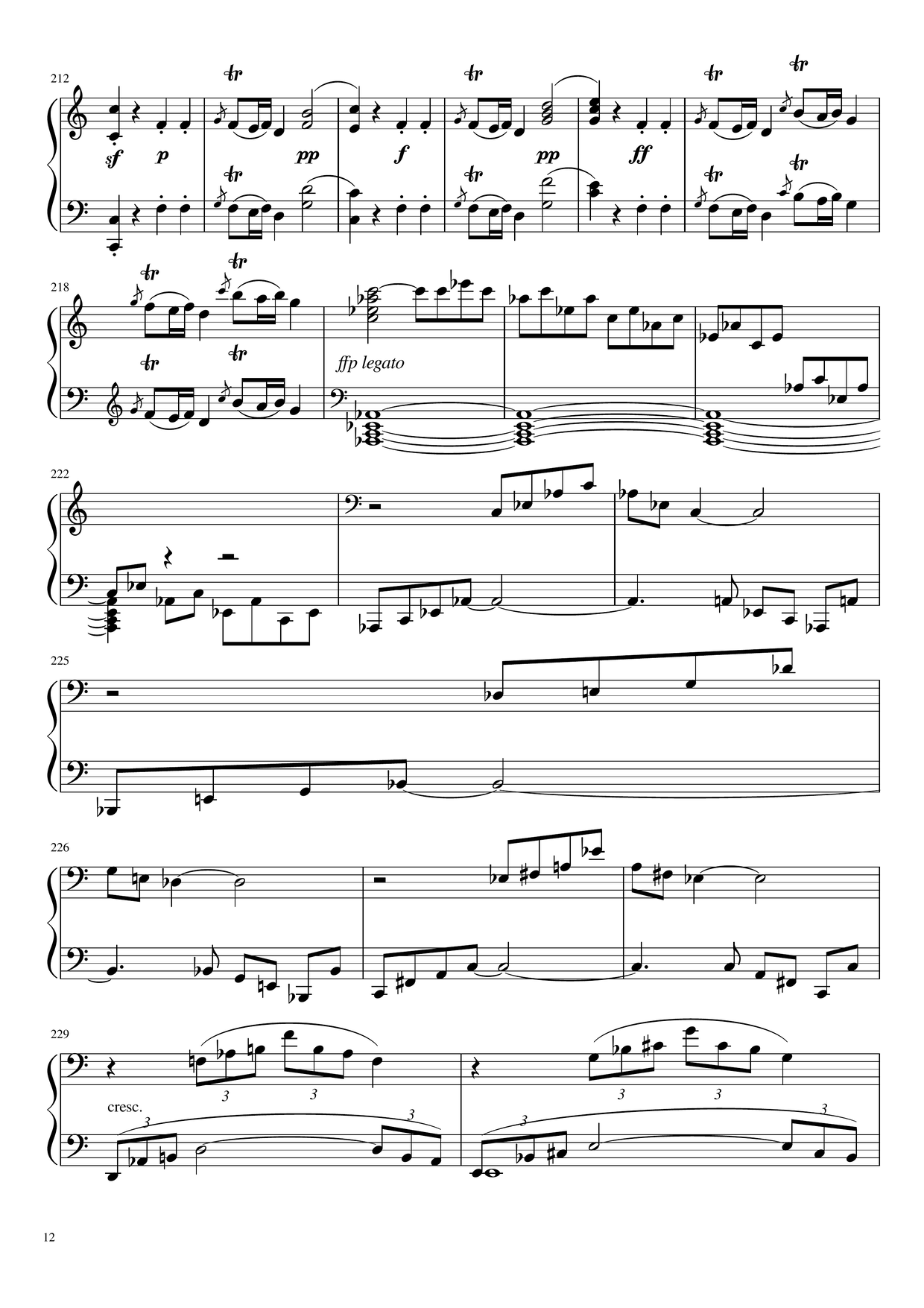}
    \end{subfigure}
    \hfill
    \begin{subfigure}{0.24\linewidth}
      \includegraphics[width=\linewidth]{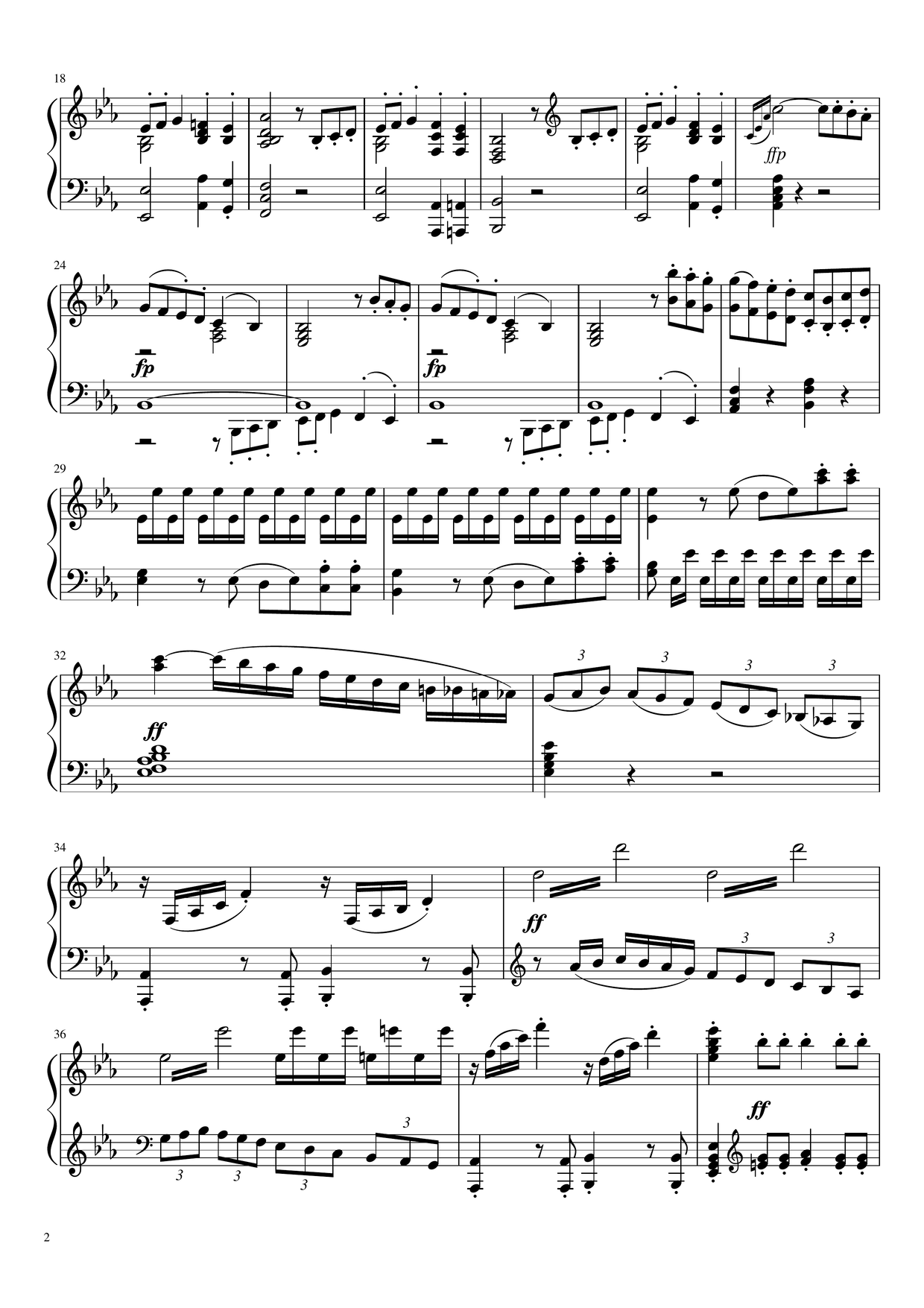}
    \end{subfigure}
    \hfill
    \begin{subfigure}{0.24\linewidth}
      \includegraphics[width=\linewidth]{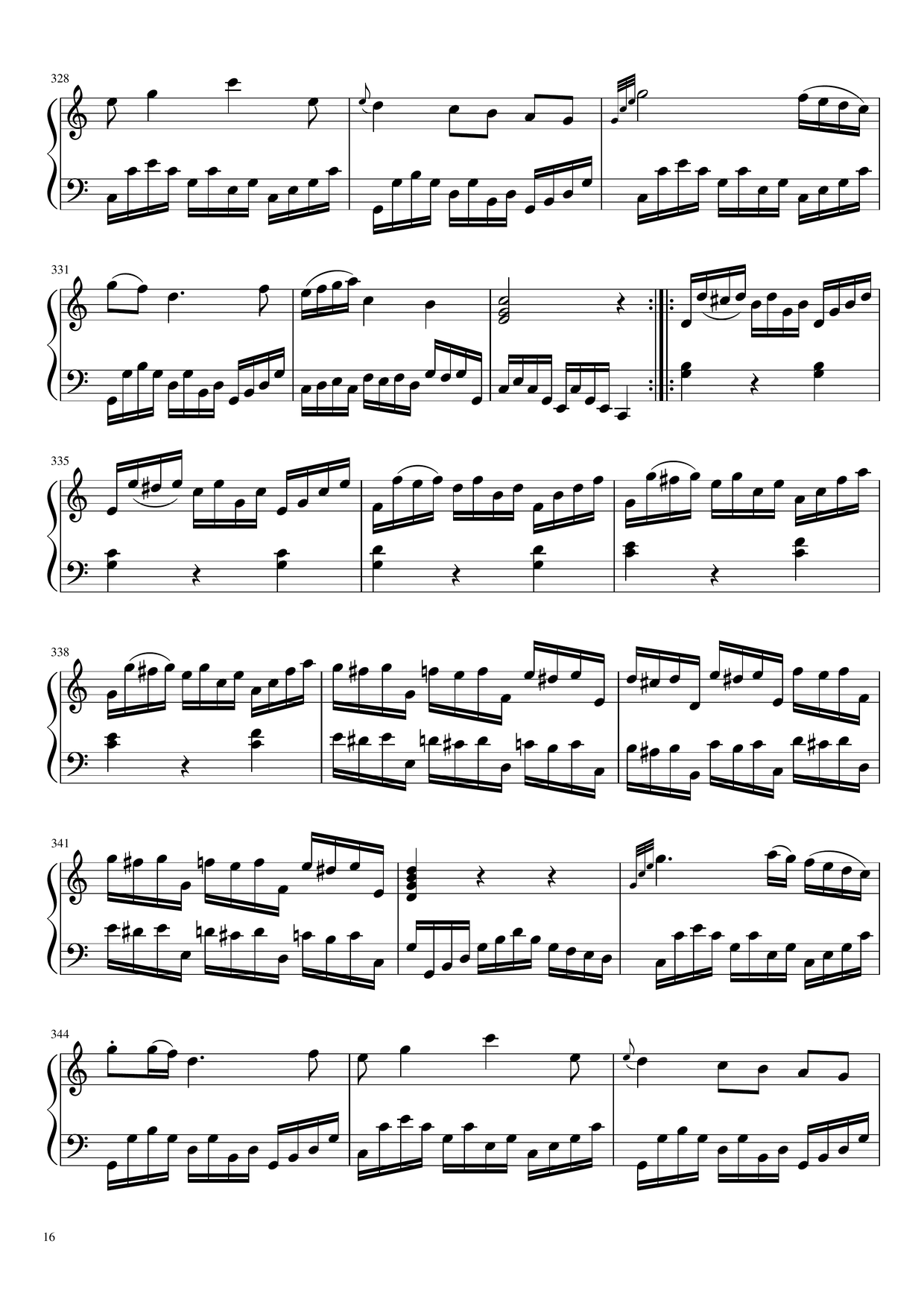}
    \end{subfigure}
    \hfill

    \caption{\PlayIt{} (652 pieces, 9 difficulty levels)}
    \label{fig:dataset_playit}
  \end{subfigure}
  \begin{subfigure}{\linewidth}
    \centering
    \begin{subfigure}{0.24\linewidth}
      \includegraphics[width=\linewidth]{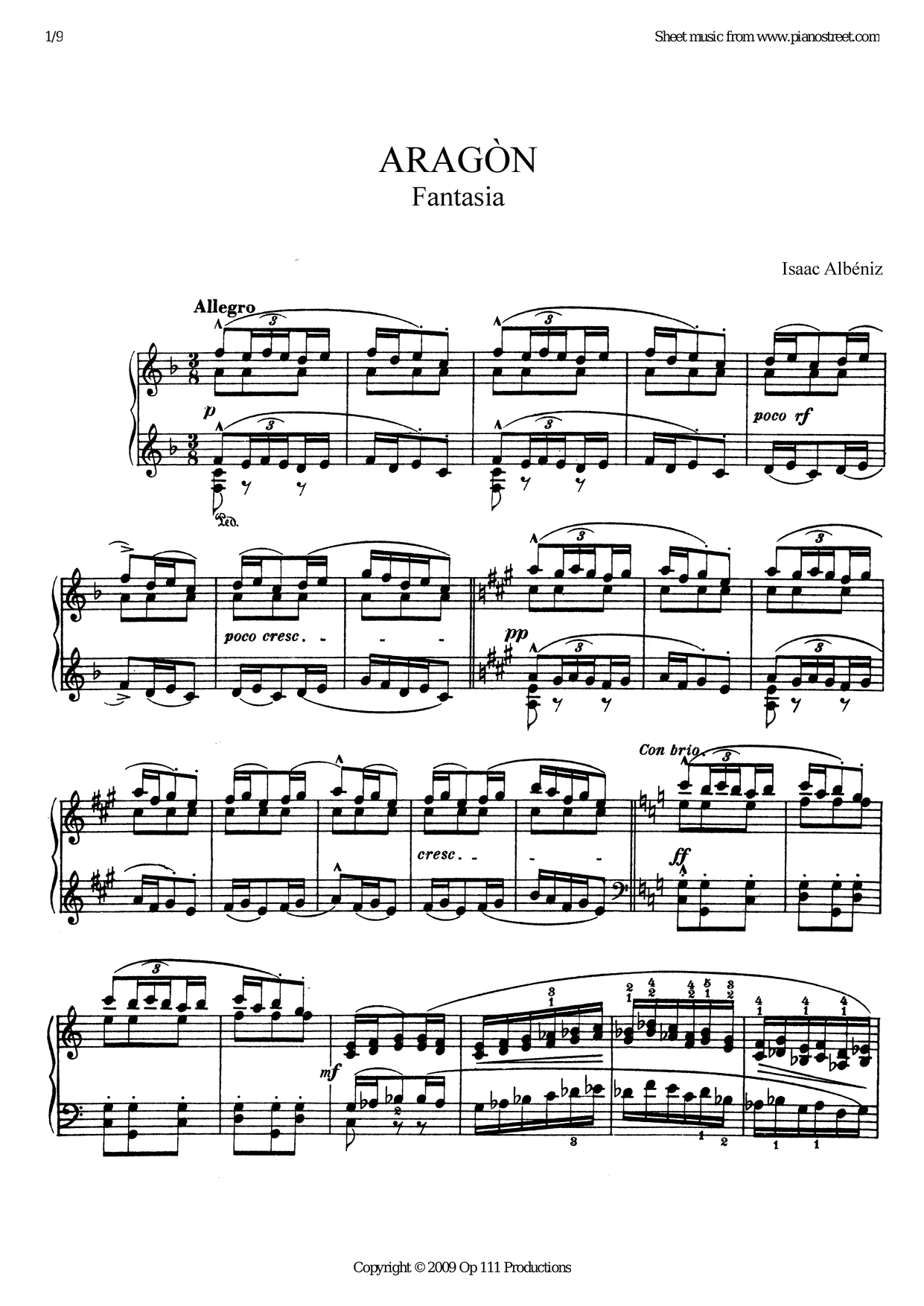}
    \end{subfigure}
    \hfill
    \begin{subfigure}{0.24\linewidth}
      \includegraphics[width=\linewidth]{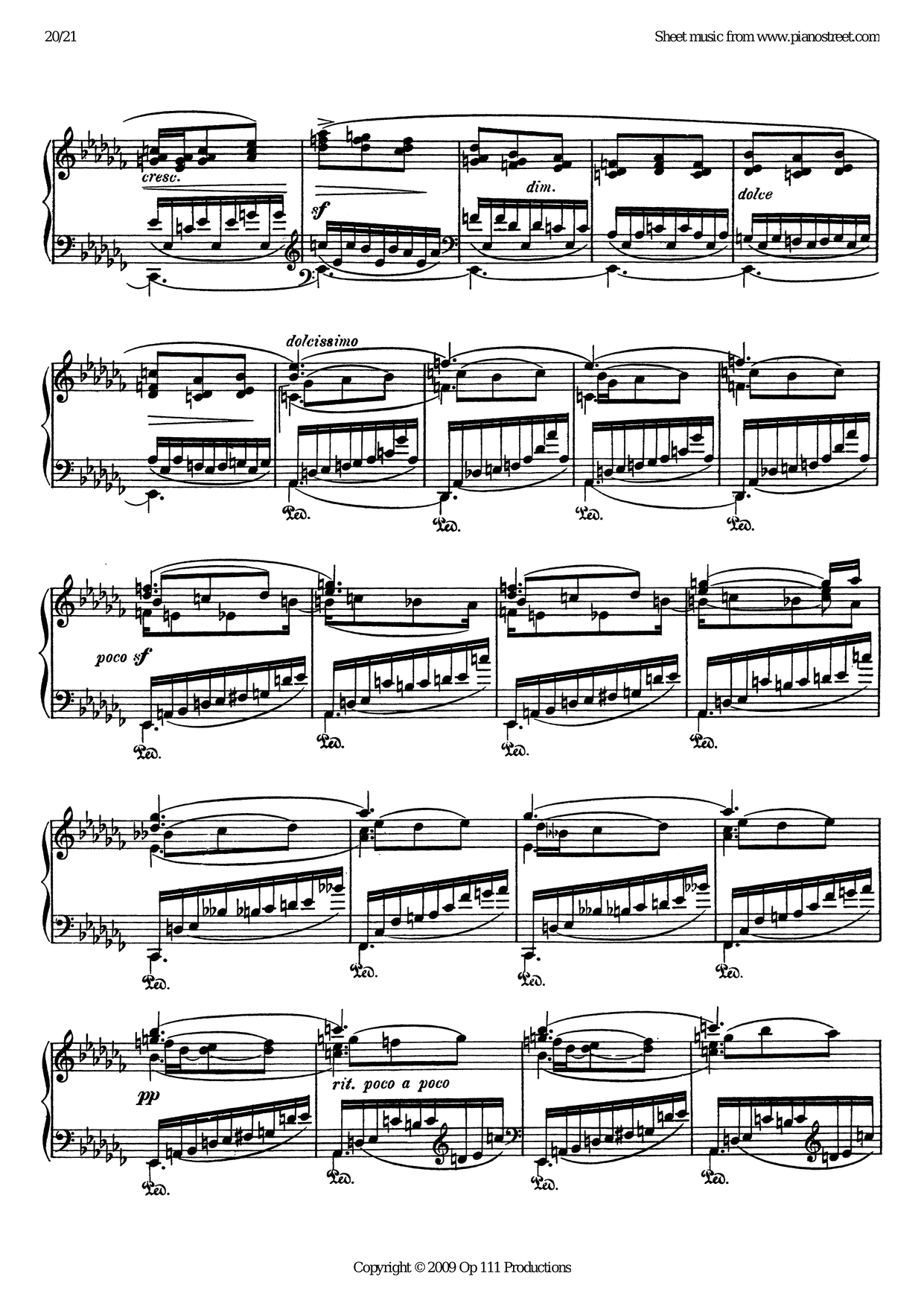}
    \end{subfigure}
    \hfill
    \begin{subfigure}{0.24\linewidth}
      \includegraphics[width=\linewidth]{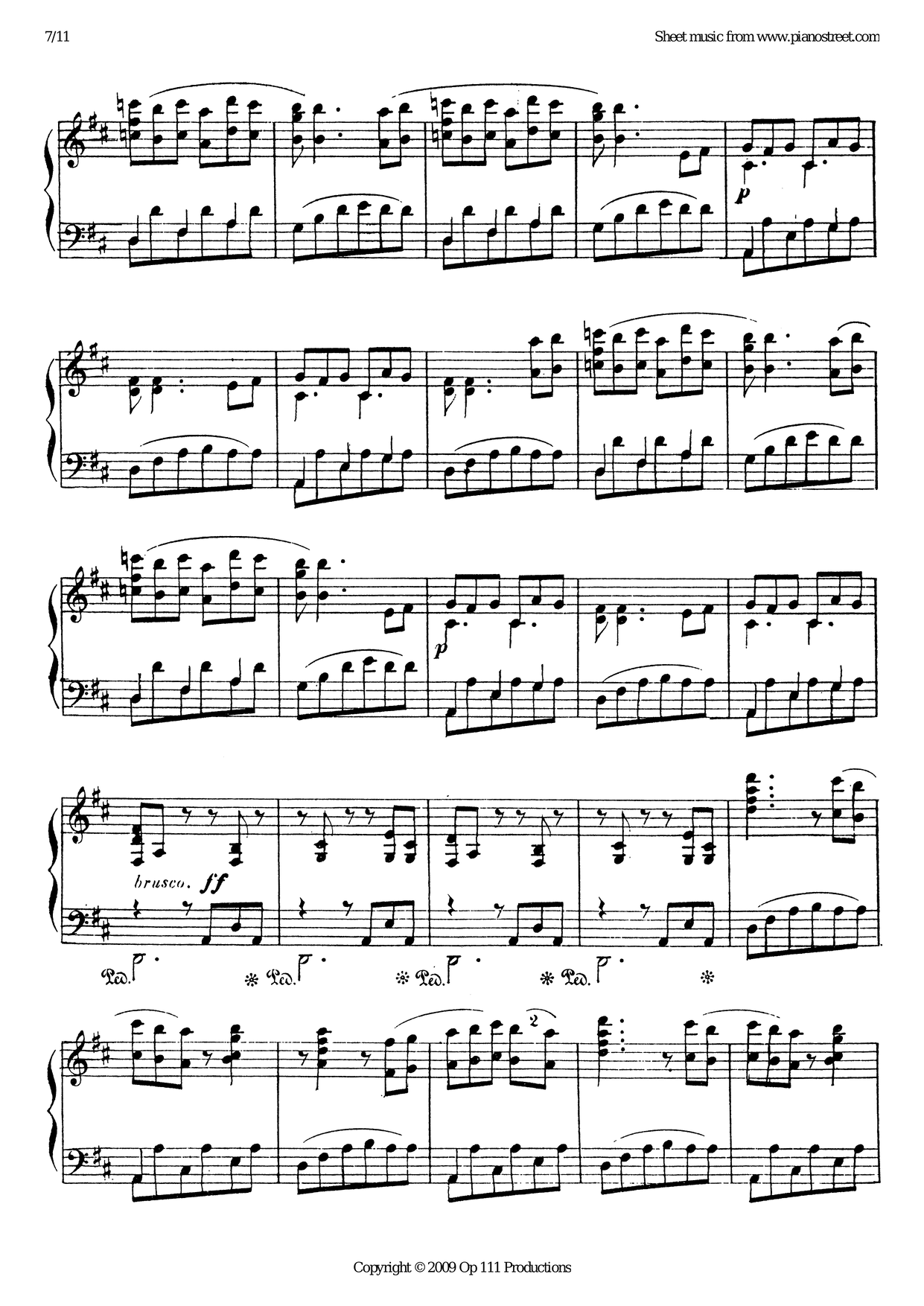}
    \end{subfigure}
    \hfill
    \begin{subfigure}{0.24\linewidth}
      \includegraphics[width=\linewidth]{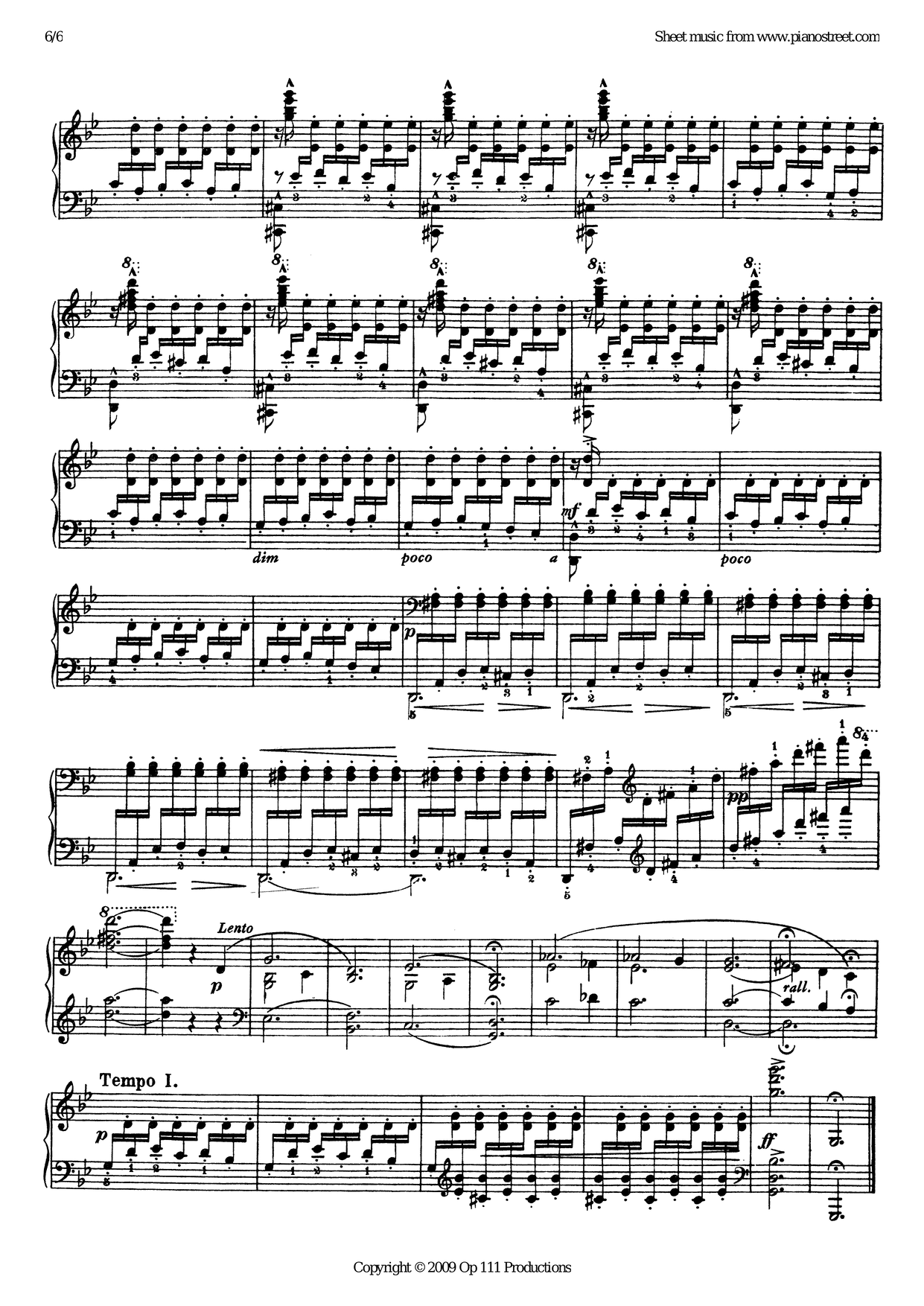}
    \end{subfigure}
    \hfill
    \caption{\PianoStreet{} (2,816 pieces, 9 difficulty levels)}
    \label{fig:dataset_pianostreet}
  \end{subfigure}
  \caption{Representative score pages from the three score difficulty classification datasets.}
  \label{fig:datasets_difficulty}
\end{figure}

We provide representative examples from each dataset used in the downstream evaluation, organized by task:
\begin{itemize}
    \item Figure~\ref{fig:datasets_full_page} --- Full-page music score recognition: \Mozarteum{} and \Polish{}.
    \item Figure~\ref{fig:datasets_staff} --- Staff-level music score recognition: \Capitan{}, \Guatemala{}, \ILS{}, \Malaga{}, and \FMT{}.
    \item Figure~\ref{fig:datasets_detection} --- Music symbol detection: \DeepScores{}.
    \item Figure~\ref{fig:datasets_difficulty} --- Score difficulty classification: \FreeScores{}, \PlayIt{}, and \PianoStreet{}.
\end{itemize}

\newpage

\subsection{General-Purpose Encoder Specifications}
\label{sup-sec:encoders}

For reproducibility, Table~\ref{tab:encoder_models} lists the exact model checkpoints used for all general-purpose encoder baselines, as hosted on the Hugging Face Model Hub.

\begin{table}[ht!]
    \centering
    \caption{Hugging Face model identifiers for all general-purpose encoders.}
    \label{tab:encoder_models}
    \setlength{\tabcolsep}{12pt} 
    \begin{tabular}{ll}
        \toprule[1pt]
        \textbf{Model} & \textbf{Hugging Face ID} \\
        \cmidrule(lr){1-2}
        PaliGemma~2~\cite{Steiner:Paligemma2:2024}   & google/paligemma2-3b-pt-896 \\
        Kosmos-2.5~\cite{Tengchao:Kosmos25:2025}      & microsoft/kosmos-2.5 \\
        Qwen3-VL~\cite{Bai:Qwen3:2025}                & Qwen/Qwen3-VL-8B-Instruct \\
        DINOv3-7B~\cite{Simeoni:Dinov3:2025}          & facebook/dinov3-vit7b16-pretrain-lvd1689m \\
        \bottomrule[1pt]
    \end{tabular}
\end{table}
\newpage

\subsection{Downstream Tasks}
\label{sup-sec:experiments}

This section provides implementation details and per-dataset results for each of the four downstream tasks evaluated in Section~3 of the main paper. For each task, we describe the task-specific head architecture and the linear probing and fine-tuning protocols, followed by a per-dataset breakdown of results that complements the dataset-averaged figures reported in the main paper.

\subsubsection{Full-Page Music Score Recognition}
\label{sup-subsec:full_page_omr}

We follow the state-of-the-art architecture in full-page music score recognition~\cite{RiosVila:IJCV:2026} and use an autoregressive Transformer decoder as the task-specific head. Given a score page $\mathbf{x} \in \mathbb{R}^{3 \times H \times W}$, the decoder attends to the patch embeddings and generates the output sequence left-to-right via cross-attention. Performance is measured by Symbol Error Rate (SER), defined as the normalized edit distance between predicted and ground-truth sequences.

Training follows a curriculum learning strategy adapted from~\cite{RiosVila:IJCV:2026}. We omit the system-level pre-training stage of the original procedure, as \musvit{} is already pre-trained on full-page score images. Training therefore begins with synthetic full-page scores generated from excerpts of the GrandStaff dataset~\cite{RiosVila:IJDAR:2023}, rendered using Verovio~\cite{Pugin:ISMIR:2014}, with the number of systems per page increasing incrementally. Once this stage converges, real pages from the target dataset are gradually introduced while retaining synthetic samples.

The two evaluation scenarios differ in how the encoder is treated during this final stage. In the \emph{linear probing} scenario, the encoder remains frozen throughout and only the decoder is updated during training. In the \emph{fine-tuning} scenario, the encoder is also frozen during the synthetic stage; however, once real pages are introduced, the encoder weights are unfrozen and the full model is jointly optimized.

Tables~\ref{tab:omr_full_page_datasets_lp} and~\ref{tab:omr_full_page_datasets_ft} report per-dataset results, which mirror the average trends shown in Tables~2 and~3 of the main paper. Under linear probing, the gap between \musvit{} and all general-purpose encoders is consistent across both corpora: \musvit{} achieves 17.6\% on \Mozarteum{} and 15.2\% on \Polish{}, while all baselines remain above 46\% on both datasets. Remarkably, both per-dataset linear probing results individually beat the state of the art~\cite{RiosVila:IJCV:2026} on \Polish{} (15.2\% vs.\ 25.8\%), and approach it closely on \Mozarteum{} (17.6\% vs.\ 14.1\%), without any task-specific training. Under fine-tuning, both \musvit{} variants substantially outperform the state of the art on both datasets. The improvement is especially pronounced on \Polish{} (SoTA~\cite{RiosVila:IJCV:2026}: 25.8\% $\to$ \musvit{}: 11.3\%), more than halving the error, compared to a more modest but still substantial gain on \Mozarteum{} (SoTA~\cite{RiosVila:IJCV:2026}: 14.1\% $\to$ \musvit{}: 10.5\%).

\begin{table}[t]
    \centering
    \caption{Per-dataset Symbol Error Rate (SER \%, $\downarrow$) for {full-page recognition} under the \emph{linear probing} scenario (frozen encoder). Best results are in \textbf{bold}; second best are \underline{underlined}.}
    \label{tab:omr_full_page_datasets_lp}
    \setlength{\tabcolsep}{10pt} 
    \renewcommand{\arraystretch}{1.3} 
    \resizebox{0.6\columnwidth}{!}{%
        \begin{tabular}{lcc}
        \toprule[1pt]
                                                    & \Mozarteum{}   & \Polish{}    \\ \cmidrule(lr){1-3}
        PaliGemma 2~\cite{Steiner:Paligemma2:2024}  & 46.2              & 50.9            \\
        Kosmos-2.5~\cite{Tengchao:Kosmos25:2025}    & 56.0              & 68.8            \\
        Qwen3-VL~\cite{Bai:Qwen3:2025}              & 48.3              & 53.7            \\
        DINOv3-7B~\cite{Simeoni:Dinov3:2025}        & 57.4              & 56.4            \\ \cmidrule(lr){1-3}
        \rowcolor[gray]{0.9} \musvit{}              & \textbf{17.6}     & \textbf{15.2}    \\ 
        \rowcolor[gray]{0.9} \musvitl{}       & \underline{22.6}  & \underline{19.3} \\ \bottomrule[1pt]
        \end{tabular}
    }
\end{table}

\begin{table}[t]
    \centering
    \caption{Per-dataset Symbol Error Rate (SER \%, $\downarrow$) for {full-page recognition} under the \emph{fine-tuning} scenario. Best results are in \textbf{bold}; second best are \underline{underlined}.}
    \label{tab:omr_full_page_datasets_ft}
    \setlength{\tabcolsep}{10pt}
    \renewcommand{\arraystretch}{1.3}
    \resizebox{0.6\columnwidth}{!}{%
        \begin{tabular}{lcc}
        \toprule[1pt]
                                                    & \Mozarteum{}      & \Polish{}    \\ \cmidrule(lr){1-3}
        State-of-the-art~\cite{RiosVila:IJCV:2026}  & 14.1              & 25.8       \\ \cmidrule(lr){1-3}
        \rowcolor[gray]{0.9} \musvit{}              & \textbf{10.5} & \textbf{11.3}         \\ 
        \rowcolor[gray]{0.9} \musvitl{}             & \underline{11.3}    & \underline{12.4}         \\ \bottomrule[1pt]
        \end{tabular}
    }
\end{table}

\subsubsection{Staff-Level Music Score Recognition}
\label{sup-subsec:staff_line_omr}

The task head follows the standard architecture of~\cite{Martinez:ISMIR:2024}: patch embeddings are reshaped to their 2D spatial grid and column-wise averaged to produce a 1D temporal sequence, which is processed by two stacked bidirectional LSTM layers; a fully connected layer with softmax outputs a posteriogram over the extended vocabulary $\Sigma' = \Sigma \cup \{\text{blank}\}$, and the final prediction is obtained via connectionist temporal classification (CTC) greedy decoding.

In the \emph{linear probing} scenario, the encoder remains frozen and only the BiLSTM layers and the CTC head are trained from scratch. In the \emph{fine-tuning} scenario, we apply LoRA adaptation~\cite{Hu:ICLR:2022} (rank 8) to the query, key, and value projections of all self-attention modules in the encoder, enabling parameter-efficient adaptation while keeping the remaining encoder weights frozen. The BiLSTM layers and CTC head are trained jointly with the LoRA parameters.

Under linear probing (see Table~\ref{tab:omr_staff_line_datasets_lp}), \musvit{} achieves the best SER on four of the five corpora: \Capitan{} (26.2\%), \Guatemala{} (7.7\%), \FMT{} (29.7\%), and \ILS{} (7.2\%). The only exception is \Malaga{}, where Qwen3-VL performs slightly better (17.4\% vs.\ 21.0\%). These results suggest that \musvit{} generalizes effectively across a wide range of musical notations and engraving styles, rather than being narrowly specialized to a single modern domain. Additional evidence of \musvit{} competence in CWMN comes from the full-page music score recognition results (see Section~\ref{sup-subsec:full_page_omr}), where both evaluation datasets consist of printed CWMN and \musvit{} substantially outperforms all general-purpose baselines.

Under fine-tuning (see Table~\ref{tab:omr_staff_line_datasets_ft}), the task-specific state of the art~\cite{Martinez:ISMIR:2024} retains an advantage on \Capitan{} (8.6\%), \FMT{} (9.0\%), and \ILS{} (2.7\%), where per-corpus training signal is strongest. In contrast, both \musvit{} variants surpass the state of the art on \Guatemala{} (\musvit{}: 2.0\%, \musvitl{}: 1.6\% vs.\ 2.2\%) and \Malaga{} (\musvit{}: 16.7\%, \musvitl{}: 15.0\% vs.\ 17.3\%), the two corpora where LoRA adaptation generalizes better from the pre-trained representations. Notably, the gap between \musvit{} and \musvitl{} remains small across all corpora, suggesting that the lightweight variant is a competitive alternative at substantially reduced parameter count.

\begin{table}[t]
    \centering
    \caption{Per-dataset Symbol Error Rate (SER \%, $\downarrow$) for {staff-level recognition} under the \emph{linear probing} scenario (frozen encoder). Best result are in \textbf{bold}; second best are \underline{underlined}.}
    \label{tab:omr_staff_line_datasets_lp}
    \setlength{\tabcolsep}{10pt} 
    \renewcommand{\arraystretch}{1.3} 
    \resizebox{0.8\columnwidth}{!}{%
        \begin{tabular}{lcccccc}
        \toprule[1pt]
                                                    & \Capitan{}            & \Guatemala{}      & \FMT{}            & \Malaga{}         & \ILS{} \\ \cmidrule(lr){1-6}
        PaliGemma 2~\cite{Steiner:Paligemma2:2024}  & 34.1                  & 13.6              & 36.4              & 23.1              & 12.5 \\
        Kosmos-2.5~\cite{Tengchao:Kosmos25:2025}    & 62.7                  & 26.2              & 55.8              & 58.3              & 34.6 \\
        Qwen3-VL~\cite{Bai:Qwen3:2025}              & \underline{32.9}      & 13.6              & \underline{31.1}  & \textbf{17.4}     & 9.9 \\
        DINOv3-7B~\cite{Simeoni:Dinov3:2025}        & 47.3                  & 19.7              & 40.1              & 35.9              & 17.6 \\ \cmidrule(lr){1-6}
        \rowcolor[gray]{0.9} \musvit{}              & \textbf{26.2}         & \textbf{7.7}      & \textbf{29.7}     & \underline{21.0}  & \textbf{7.2} \\ 
        \rowcolor[gray]{0.9} \musvitl{}       & 33.1                  & \underline{12.2}  & 32.4              & 28.5              & \underline{8.8} \\ \bottomrule[1pt]
        \end{tabular}
    }
\end{table}

\begin{table}[t]
    \centering
    \caption{Per-dataset Symbol Error Rate (SER \%, $\downarrow$) for {staff-level recognition} under the \emph{fine-tuning} scenario. Best results are in \textbf{bold}; second best are \underline{underlined}.}
    \label{tab:omr_staff_line_datasets_ft}
    \setlength{\tabcolsep}{10pt}
    \renewcommand{\arraystretch}{1.3}
    \resizebox{0.8\columnwidth}{!}{%
        \begin{tabular}{lcccccc}
        \toprule[1pt]
                                                        & \Capitan{}    & \Guatemala{}      & \FMT{}            & \Malaga{}         & \ILS{}    \\ \cmidrule(lr){1-6}
        State-of-the-art~\cite{Martinez:ISMIR:2024}     &\textbf{8.6}   & 2.2               & \textbf{9.0}      & 17.3              & \textbf{2.7}      \\ \cmidrule(lr){1-6}
        \rowcolor[gray]{0.9} \musvit{}                  & \underline{9.1}          & \textbf{1.6}   & \underline{13.3}              & \textbf{15.0}  & \underline{3.8}         \\ 
        \rowcolor[gray]{0.9} \musvitl{}           & 11.1           & \underline{2.0}      & 15.0  & \underline{16.7}     & 4.4         \\ \bottomrule[1pt]
        \end{tabular}
    }
\end{table}

\subsubsection{Music Symbol Detection}
\label{sup-subsec:detection}

The task head is a Faster R-CNN detector~\cite{Ren:NeurIPS:2015} with a Feature Pyramid Network (FPN): encoder patch embeddings are reshaped from a 1D token sequence into a 2D spatial grid and fed through the FPN, which produces multi-scale feature maps, followed by the region proposal network and classification heads.

In the \emph{linear probing} scenario, the encoder is frozen; only the task head parameters are trained. In the \emph{fine-tuning} scenario, we apply LoRA adaptation~\cite{Hu:ICLR:2022} (rank 8) to the query, key, and value projections of all self-attention modules, following the same protocol as staff-level recognition (see Section~\ref{sup-subsec:staff_line_omr}). The detection head is jointly optimized with the LoRA parameters.

Table~\ref{tab:symbol_detection_lp_all_metrics} reports the full set of COCO detection metrics under linear probing. \musvit{} leads on six of the seven metrics, with a particularly strong advantage on AP$_S$ (52.8\% vs.\ DINOv3-7B's 43.2\%), the metric that captures detection of small objects---precisely the regime of noteheads, accidentals, and fine-grained music symbols. The only metric where DINOv3-7B edges ahead is AP$_L$ (81.6\% vs.\ 79.1\%), which corresponds to larger objects that are less diagnostic for music notation.

Table~\ref{tab:symbol_detection_ft_all_metrics} reports per-class AP$_{50}$ under fine-tuning. We follow the exact evaluation protocol and class selection of the state of the art~\cite{Luo:ESWA:2024}: the seven reported classes (NoteH., Stem, Beam, Rest$_4$, Rest$_2$, Rest$_8$, Rest$_{16}$) are inherited directly from the baseline, which selected them on the basis of their small visual size; we adopt the same subset for direct comparability. \musvit{} outperforms the state of the art on every reported class, with the largest absolute gain on Stem (+17.4 points: 68.1\% $\to$ 85.5\%)---the most geometrically challenging symbol due to its thin, elongated appearance. Finally, the comparison is structurally asymmetric in two compounding respects: the state of the art~\cite{Luo:ESWA:2024} uses a Transformer-based end-to-end detection model considerably heavier than our Faster R-CNN head, and it fully fine-tunes all model parameters, whereas \musvit{} updates only the LoRA adapters (rank 8) in the attention projections, keeping the bulk of the encoder frozen. The 7-point mAP$_{50}$ advantage is therefore doubly conservative: achieved with both a lighter detection head and a parameter-efficient encoder adaptation, making the representational benefit of music-specific pre-training all the more pronounced.

\begin{table}[t]
    \centering
    \caption{Standard COCO detection metrics ($\uparrow$) on the \DeepScores{} dataset for {music symbol detection} under the \emph{linear probing} scenario (frozen encoder). Best results are in \textbf{bold}; second best are \underline{underlined}.}
    \label{tab:symbol_detection_lp_all_metrics}
    \setlength{\tabcolsep}{6pt}
    \renewcommand{\arraystretch}{1.4}
    \resizebox{0.8\columnwidth}{!}{%
        \begin{tabular}{l ccccccc}
            \toprule[1.5pt]
            & \textbf{mAP} & \textbf{w-mAP} & \textbf{AP$_{50}$} & \textbf{AP$_{75}$} & \textbf{AP$_{S}$} & \textbf{AP$_{M}$} & \textbf{AP$_{L}$} \\  \cmidrule(lr){1-8}
            PaliGemma 2~\cite{Steiner:Paligemma2:2024}  & 31.7 & 39.0 & 61.6 & 27.7 & 26.0 & 42.7 & 56.7 \\
            Kosmos-2.5~\cite{Tengchao:Kosmos25:2025}    & 42.7 & 47.4 & 71.5 & 43.3 & 36.1 & 49.5 & 54.1 \\
            Qwen3-VL~\cite{Bai:Qwen3:2025}              & 67.1 & 61.0 & 89.3 & 76.7 & 31.2 & 70.9 & 76.2 \\ 
            DINOv3-7B~\cite{Simeoni:Dinov3:2025}        & 70.4 & 62.0 & 93.0 & 83.1 & 43.2 & 73.4 & \textbf{81.6} \\  \cmidrule(lr){1-8}
            
            \rowcolor[gray]{0.95} \musvit{}         & \textbf{79.7} & \textbf{80.7} & \textbf{95.1} & \textbf{89.9} & \textbf{52.8} & \textbf{81.1} & \underline{79.1} \\ 
            \rowcolor[gray]{0.95} \musvitl{}  & \underline{79.1}          & \underline{80.4}          & \textbf{95.1} & \underline{89.0}          & \underline{51.7} & \underline{80.8} & 75.4 \\
            \bottomrule[1.5pt]
        \end{tabular}
    }
\end{table}

\begin{table}[t]
    \centering
    \caption{Per-class AP$_{50}$ (\%, $\uparrow$) on the \DeepScores{} dataset for {music symbol detection} under the \emph{fine-tuning} scenario. Best results are in \textbf{bold}; second best are \underline{underlined}.}
    \label{tab:symbol_detection_ft_all_metrics}
    \setlength{\tabcolsep}{4.5pt} 
    \renewcommand{\arraystretch}{1.5}
    \resizebox{0.8\columnwidth}{!}{%
        \begin{tabular}{l ccccccc | c}
            \toprule[1.5pt]
             & \textbf{NoteH.} & \textbf{Stem} & \textbf{Beam} & \textbf{Rest$_{4}$} & \textbf{Rest$_{2}$} & \textbf{Rest$_{8}$} & \textbf{Rest$_{16}$}  & \textbf{mAP$_{50}$} \\  \cmidrule(lr){1-9}
            State-of-the-art~\cite{Luo:ESWA:2024} & 97.2 & 68.1 & 93.7 & 91.2 & 92.0 & 93.7 & 97.4 &  90.5 \\ \cmidrule(lr){1-9}
            
            \rowcolor[gray]{0.95} \musvit{} & \textbf{99.0} & \textbf{85.5} & \underline{98.7} & \textbf{99.0} & \textbf{98.9} & \textbf{99.0} & \textbf{98.7} &  \textbf{97.0} \\ 
            \rowcolor[gray]{0.95} \musvitl{}& \textbf{99.0} & \underline{84.3} & \textbf{98.8} & \textbf{99.0} & \underline{98.7} & \underline{98.9} & \underline{97.8} &  \underline{96.6} \\
            \bottomrule[1.5pt]
        \end{tabular}
    }
\end{table}

\subsubsection{Score Difficulty Classification}
\label{sup-subsec:difficulty}

For a piece spanning $n$ pages, each page $i$ is processed independently by the encoder. Patch embeddings are mean-pooled to produce a page-level vector $\mathbf{e}_i \in \mathbb{R}^d$. The two evaluation scenarios differ in how these vectors are aggregated into a document-level prediction. In the \emph{linear probing} scenario, page embeddings are averaged to form a document representation $\bar{\mathbf{e}} = \frac{1}{n}\sum_{i=1}^{n}\mathbf{e}_i$, which is fed to an MLP classifier on top of the frozen encoder. In the \emph{fine-tuning} scenario, the sequence $(\mathbf{e}_1, \ldots, \mathbf{e}_n)$ is processed by a GRU-based recurrent network. LoRA adaptation~\cite{Hu:ICLR:2022} (rank 8) is applied to the query, key, and value projections of all self-attention modules, and the GRU head is optimized jointly with the LoRA parameters.

Under linear probing (see Table~\ref{tab:diff_datasets_lp}), per-dataset results reveal a more varied picture than the average suggests. On \PlayIt{} and \PianoStreet{}, \musvit{} variants lead on most metrics: \musvitl{} achieves the best $Acc_{0}$ on \PlayIt{} (31.2\%) and \musvit{} leads on \PianoStreet{} $Acc_{0}$ (54.3\%). On \FreeScores{}, however, PaliGemma~2 edges ahead on $Acc_{0}$ (62.9\% vs.\ \musvit{}'s 58.4\%), while \musvitl{} leads on $Acc_{1}$ (97.0\%). This variability is consistent with the holistic nature of the task: visual statistics such as notation density can serve as a partial proxy for difficulty, reducing the margin between music-specific and general-purpose encoders on some datasets.

Under fine-tuning (see Table~\ref{tab:diff_datasets_ft}), \musvit{} and \musvitl{} outperform the state of the art~\cite{Ramoneda:ISMIR:2023} on all three corpora and both metrics. The improvement is largest on \FreeScores{} (\musvit{}: 61.9\% $Acc_{0}$ vs.\ 47.3\%) and \PlayIt{} (\musvit{}: 47.5\% vs.\ 36.2\%). On \PianoStreet{}, \musvitl{} marginally outperforms \musvit{} on $Acc_{0}$ (56.3\% vs.\ 53.2\%), the only corpus where the lightweight variant exceeds the full model under fine-tuning.

\begin{table}[t]
    \centering
    \caption{Per-dataset accuracy (Acc$_0$ and Acc$_1$, $\uparrow$) for {score difficulty classification} under the \emph{linear probing} scenario (frozen encoder). Best results are in \textbf{bold}; second best are \underline{underlined}.}
    \label{tab:diff_datasets_lp}
    \setlength{\tabcolsep}{12pt} 
    \resizebox{0.8\columnwidth}{!}{%
    \renewcommand{\arraystretch}{1.4}
    \begin{tabular}{lcccccc}
        \toprule[1.5pt]
        & \multicolumn{2}{c}{\FreeScores{}} & \multicolumn{2}{c}{\PlayIt{}} & \multicolumn{2}{c}{\PianoStreet{}} \\
        & $Acc_{0}$ & $Acc_{1}$ & $Acc_{0}$ & $Acc_{1}$ & $Acc_{0}$ & $Acc_{1}$ \\ \cmidrule(lr){1-7}
        PaliGemma 2~\cite{Steiner:Paligemma2:2024}     & \textbf{62.9} & 96.5	&	26.2 &	67.2	&	51.2	& \textbf{88.0}	   \\ 
        Kosmos-2.5~\cite{Tengchao:Kosmos25:2025}       & 44.6 & 89.0	&	26.2 &	62.3	&	32.1	& 57.6	   \\
        Qwen3-VL~\cite{Bai:Qwen3:2025}                &  59.7 & 95.5	&	23.0 &	70.5	&	47.2	& 83.2	   \\
        DINOv3-7B~\cite{Simeoni:Dinov3:2025}           & \underline{61.2} & 96.2	&	24.6 &	\underline{72.1}	&	50.7	& 83.2	   \\ \cmidrule(lr){1-7}
        \rowcolor[gray]{0.95} \musvit{}         &        58.4 & \underline{96.7} & \underline{29.5}  & \textbf{78.7}      &	\textbf{54.3}    & \underline{86.0} \\
        \rowcolor[gray]{0.95} \musvitl{}        &        58.4 & \textbf{97.0} & \textbf{31.2}  & 70.5      &	\underline{52.6}    & 85.6 \\
        \bottomrule[1.5pt]
    \end{tabular}
    }
\end{table}

\begin{table}[t]
    \centering
    \caption{Per-dataset accuracy (Acc$_0$ and Acc$_1$, $\uparrow$) for {score difficulty classification} under the \emph{fine-tuning} scenario. Best results are in \textbf{bold}; second best are \underline{underlined}.}
    \label{tab:diff_datasets_ft}
    \setlength{\tabcolsep}{12pt} 
    \resizebox{0.8\columnwidth}{!}{%
    \renewcommand{\arraystretch}{1.4}
    \begin{tabular}{lcccccc}
        \toprule[1.5pt]
        & \multicolumn{2}{c}{\FreeScores{}} & \multicolumn{2}{c}{\PlayIt{}} & \multicolumn{2}{c}{\PianoStreet{}} \\
        & $Acc_{0}$ & $Acc_{1}$ & $Acc_{0}$ & $Acc_{1}$ & $Acc_{0}$ & $Acc_{1}$ \\ \cmidrule(lr){1-7}
        State-of-the-art~\cite{Ramoneda:ISMIR:2023}     & 47.3              & 92.4              & 36.2              & 81.7          & 31.8              & 78.8 \\ \cmidrule(lr){1-7}
        \rowcolor[gray]{0.95} \musvit{}                 & \textbf{61.9}     & \textbf{97.5}     & \textbf{47.5}     & \textbf{83.6} & \underline{53.2}  & \textbf{86.9} \\
        \rowcolor[gray]{0.95} \musvitl{}                & \underline{61.4}  & \underline{96.7}  & \underline{44.3}  & \textbf{83.6} & \textbf{56.3}     & \underline{86.3} \\
        \bottomrule[1.5pt]
    \end{tabular}
    }
\end{table}

\subsection{Fine-Tuning General-Purpose Models}
\label{sup-sec:ft_general}

To investigate whether task-specific adaptation of general-purpose models closes the gap with \musvit{}, we fine-tuned all four baselines on two representative tasks---full-page transcription and music symbol detection. As shown in Table~\ref{tab:ft_general}, fine-tuning improves the general-purpose encoders over their frozen versions; nevertheless, \musvit{} remains superior under the same downstream protocol. This confirms that the advantage stems from domain-specific knowledge rather than the evaluation protocol.

\begin{table}[t]
    \centering
    \setlength{\abovecaptionskip}{1pt}
    \setlength{\belowcaptionskip}{1pt}
    \caption{Fine-tuning general-purpose models on two representative downstream tasks. \musvit{} outperforms all fine-tuned general models while requiring substantially fewer parameters and GFLOPs.}
    \label{tab:ft_general}
    \setlength{\tabcolsep}{8pt}
    \renewcommand{\arraystretch}{1.5}
    \resizebox{0.8\columnwidth}{!}{%
        \begin{tabular}{lcccc}
            \toprule[1pt]
            \textbf{Model} & \textbf{Full-page} (SER$\downarrow$) & \textbf{Sym. Det.} (mAP$_{50}\uparrow$) & \textbf{Params} & \textbf{GFLOPs} \\ \cmidrule(lr){1-5}
            PaliGemma~2~\cite{Steiner:Paligemma2:2024}  & 15.7 & 66.9 & 400M & 1,685 \\
            Qwen3-VL~\cite{Bai:Qwen3:2025}              & 17.1 & 82.7 & 575M & 1,692 \\
            Kosmos-2.5~\cite{Tengchao:Kosmos25:2025}     & 34.3 & 74.2 & 410M & 2,976 \\
            DINOv3-7B~\cite{Simeoni:Dinov3:2025}         & 23.1 & 93.4 & 7,000M & 27,541 \\ \cmidrule(lr){1-5}
            \rowcolor[gray]{0.9} \musvit{}               & \textbf{10.9} & \textbf{97.0} & \textbf{85M} & \textbf{106} \\
            \bottomrule[1pt]
        \end{tabular}
    }
\end{table}

Moreover, \musvit{}'s pre-training is a one-time investment amortized across all downstream tasks, and the resulting model is remarkably compact: with 85M parameters and 106 GFLOPs per image, it is 5--82$\times$ smaller and 16--260$\times$ more efficient than the compared general-purpose encoders. Downstream adaptation further reduces cost through LoRA (rank 8), updating only ${\sim}$0.3M parameters per task.

\subsection{Sheet Music Representation Analyses}
\label{sup-sec:representation_analyses}

\subsubsection{Attention Heat Map Analysis}
\label{sup-subsec:heatmap_analysis}

To qualitatively assess which image regions \musvit{} prioritizes, we visualize patch-level representations from the final Transformer layer using Principal Component Analysis (PCA). For each input image, we extract the $d$-dimensional embedding for every patch, project all patches onto the first principal component, and map the resulting scalar values back onto the input image grid to form a spatial heat map. This reveals the dominant axis of variance in the encoder's representation space. A music-aware encoder should assign high variance---and thus strong activation---to semantically meaningful regions such as music notes, rests, clefs, or accidentals, rather than distributing it uniformly across staves or blank page areas.

\begin{figure}[ht!]
  \centering
  \begin{subfigure}{0.32\linewidth}
    \centering
    \includegraphics[width=\linewidth]{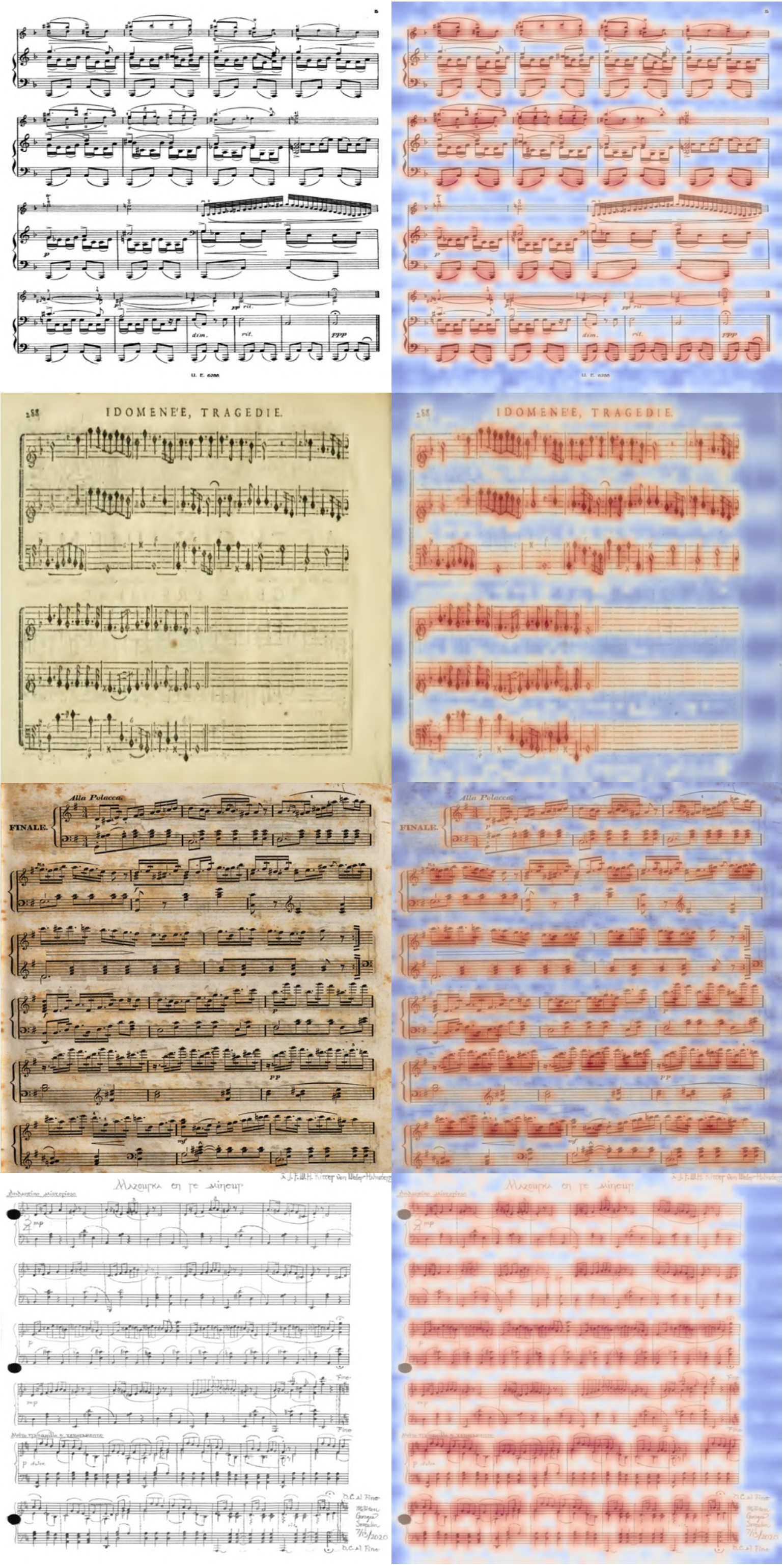}
  \end{subfigure}
  \hfill
  \begin{subfigure}{0.32\linewidth}
    \centering
    \includegraphics[width=\linewidth]{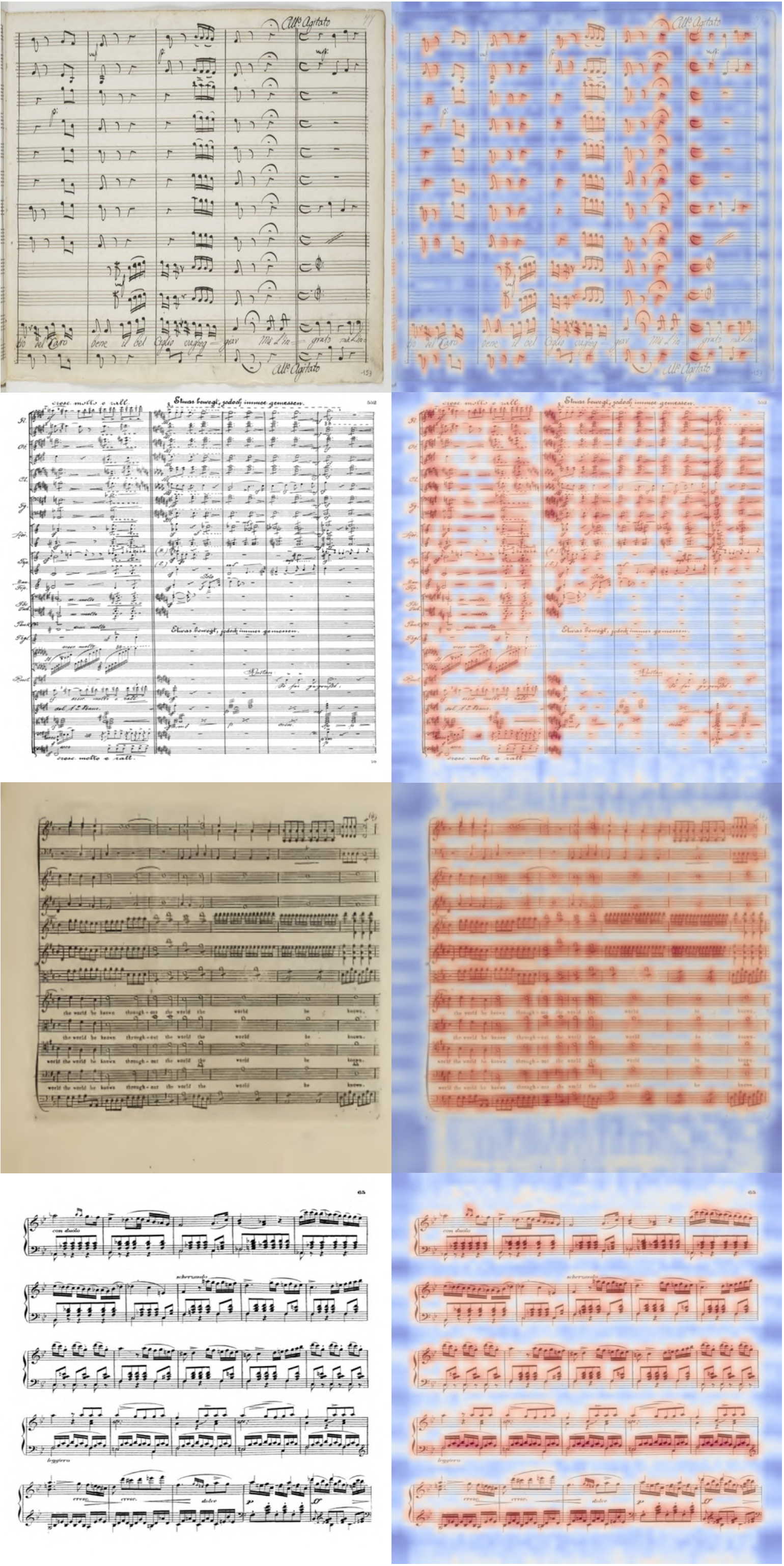}
  \end{subfigure}
  \hfill
  \begin{subfigure}{0.32\linewidth}
    \centering
    \includegraphics[width=\linewidth]{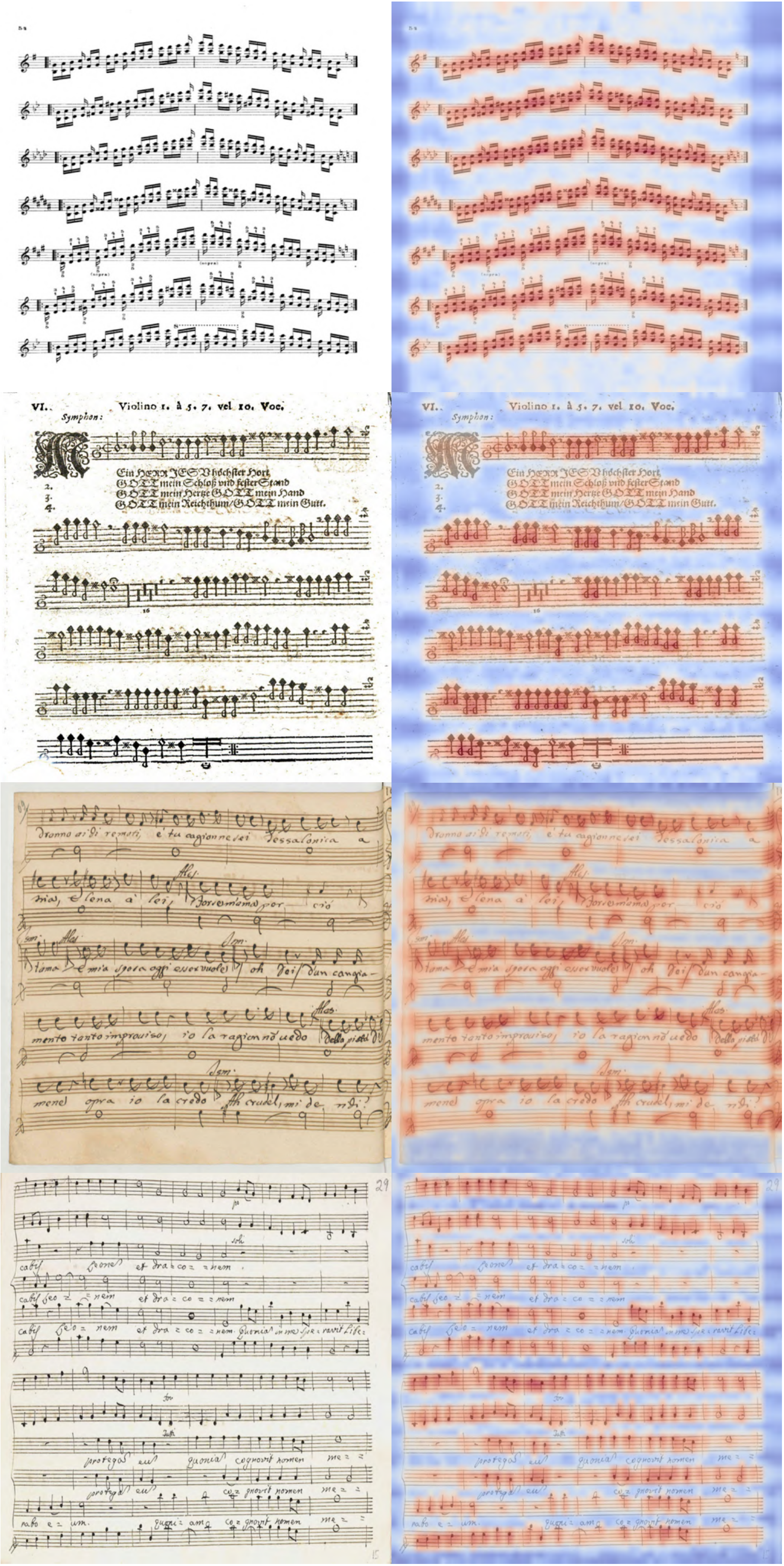}
  \end{subfigure}
  \caption{PCA activation maps for \musvit{}. Each pair shows a sheet music page (left) alongside the first-principal-component projection of patch embeddings from the final Transformer layer, rendered as a spatial heat map (right). Warm (red/orange) tones indicate high activation; cool (blue) tones indicate low activation. Twelve examples are shown across three panels, covering modern typeset scores, historical engravings, and handwritten manuscripts. In every case, \musvit{} concentrates activation along rows of musical notation while suppressing page margins and blank areas.}
  \label{fig:pca_musvit}
\end{figure}

Figure~\ref{fig:pca_musvit} shows activation maps for twelve representative sheet music pages spanning modern typeset scores, historical engravings, and handwritten manuscripts. Warm tones indicate high activation; cool tones indicate low activation. Across all examples, \musvit{} concentrates activation along rows with music notation symbols---noteheads, accidentals, rests, clefs, and rhythmic groupings---while page margins, inter-system gaps, and blank areas remain suppressed. This spatial selectivity confirms that the dominant axis of variance in \musvit{}'s representation space is anchored to the discrete musical content of the score.

\begin{figure}[ht!]
    \centering
    \includegraphics[width=1.0\linewidth]{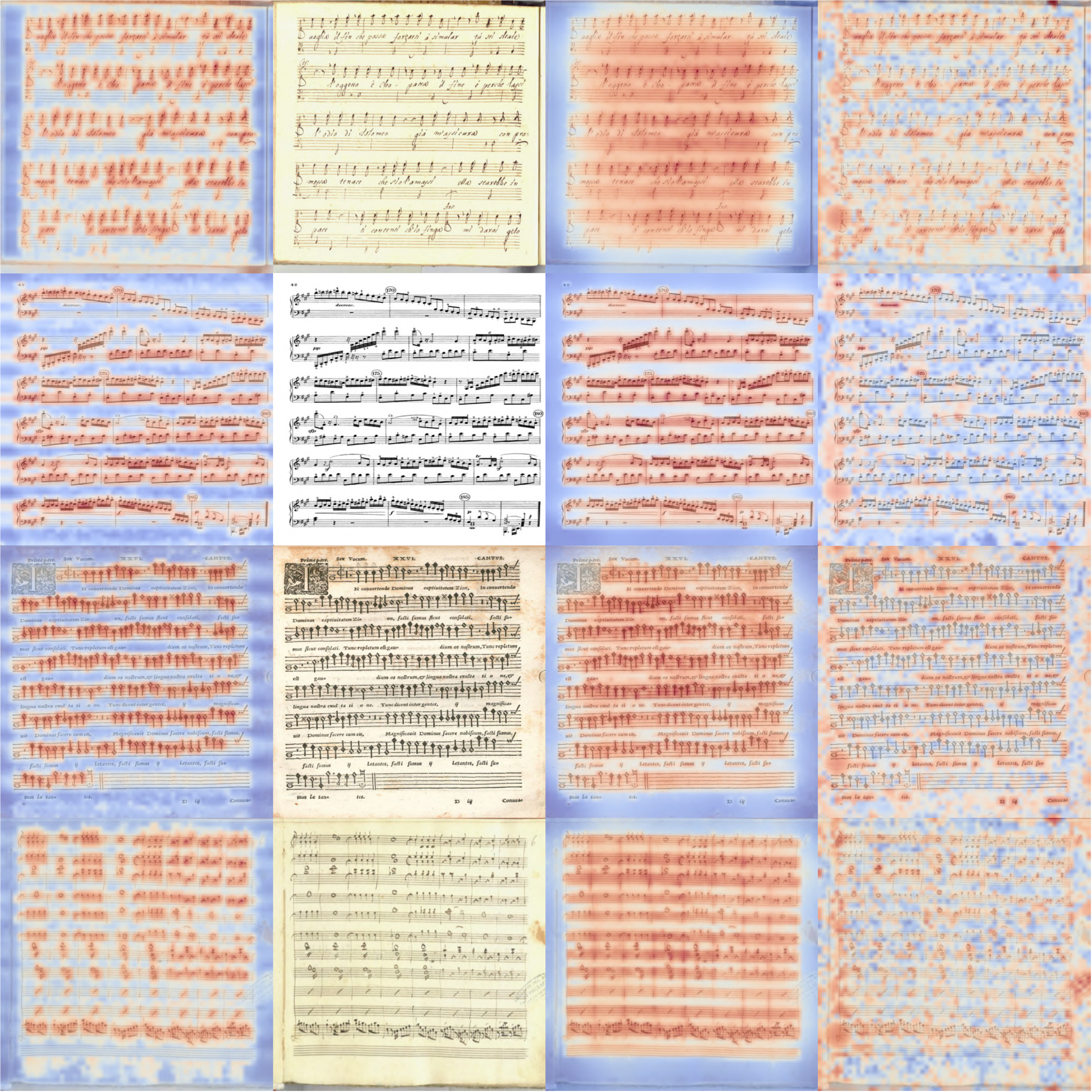}
    \caption{PCA visualization of patch embeddings for four score pages. Each row shows, from left to right: the first principal component of \musvit{} embeddings, the original score image, the first principal component of DINOv3-7B embeddings, and the first principal component of PaliGemma~2 embeddings. Warm (red/orange) tones indicate high activation; cool (blue) tones indicate low activation. \musvit{} concentrates activation tightly on notation rows, while general-purpose encoders exhibit diffuse, texture-driven activations with no alignment to musical structure.}
    \label{fig:pca_general}
\end{figure}

In contrast, Fig.~\ref{fig:pca_general} shows that general-purpose encoders produce markedly different activation patterns. For instance, DINOv3-7B distributes activation broadly across the image, capturing low-level statistics such as ink density and paper texture rather than discrete music notation content. By contrast, PaliGemma~2 exhibits scattered activations with no consistent alignment to staff regions or individual symbols. This lack of structural differentiation provides a qualitative explanation for the quantitative gap reported in Section~4 of the main paper: without representations organized around the symbolic structure of the score, these encoders cannot reliably distinguish transcriptionally similar from dissimilar pages.



\subsubsection{Nearest-Neighbor Analysis}
\label{sup-subsec:knn_analysis}

\begin{figure}[!ht]
     \centering
     \includegraphics[width=1.0\linewidth]{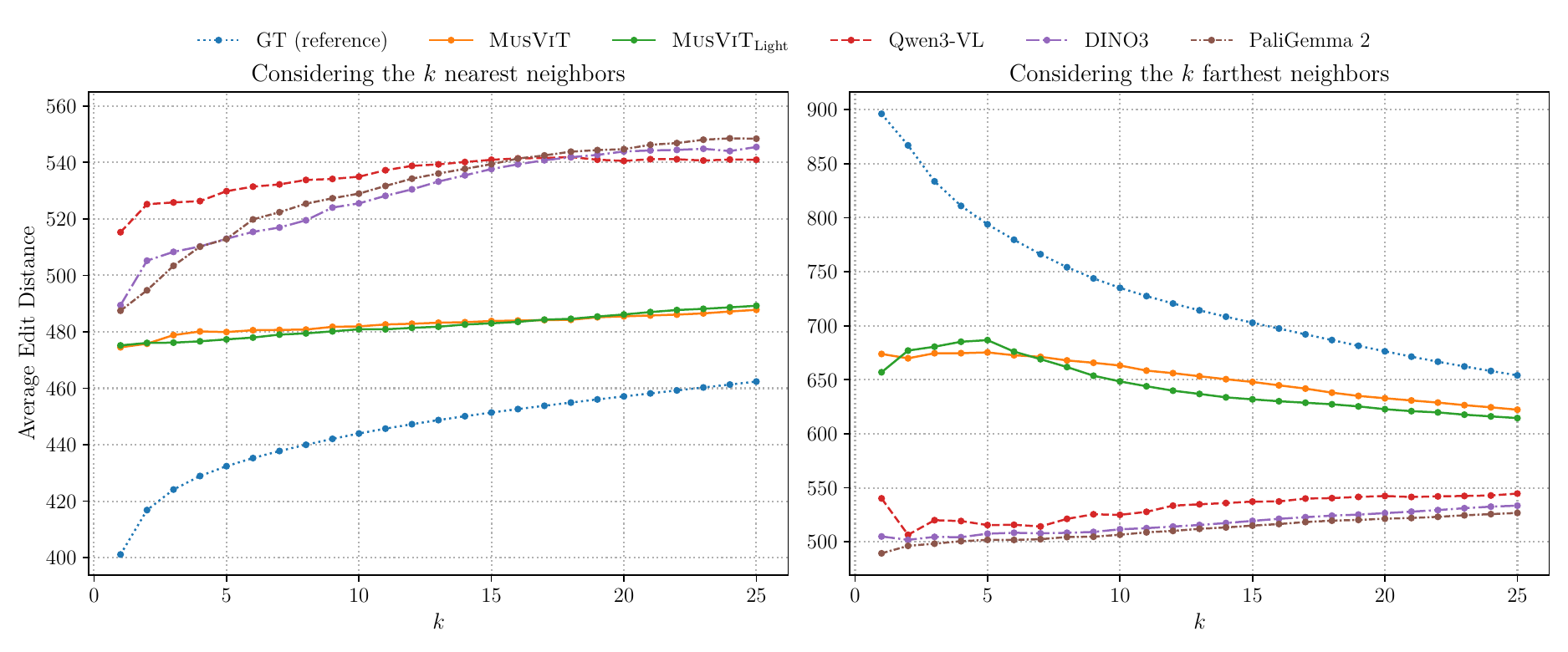}
     \caption{K-curves showing average embedding distance $\bar{d}^{\text{emb}}_k$ for the $k$ most transcriptionally similar (nearest neighbors) and most dissimilar (farthest pairs) image pairs, for $k \in \{1,\ldots,25\}$. A music-aware encoder should yield low distances for nearest neighbors and high distances for farthest pairs.}
     \label{fig:k_curves}
\end{figure}

We complement the global correlation analysis reported in Section~4 of the main paper with a targeted examination of local embedding structure. For each image $i$, we rank all other images by transcription edit distance (\Dedit{}) and retrieve the $k$ most similar (nearest neighbors) and $k$ most dissimilar (farthest pairs), for $k \in \{1, \ldots, 25\}$. We then compute the average embedding distance $\bar{d}^{\text{emb}}_k$ within each retrieved set. A music-aware encoder should produce low $\bar{d}^{\text{emb}}_k$ for nearest neighbors (transcriptionally similar images cluster together in embedding space) and high $\bar{d}^{\text{emb}}_k$ for farthest pairs (transcriptionally dissimilar images are pushed apart). The resulting k-curves, shown in Fig.~\ref{fig:k_curves}, make this local structure visible across a range of neighborhood sizes.

Figure~\ref{fig:k_curves} shows the k-curves for all encoders. General-purpose encoders show little or no separation between the nearest-neighbor and farthest-pair curves, consistent with their negative global correlations: their embedding spaces do not distinguish transcriptionally similar from dissimilar images even at large $k$. In contrast, \musvit{} and \musvitl{} exhibit a clear and consistent gap between the two curves across the full range of $k$: nearest-neighbor distances remain substantially lower ($\approx$475--490) than farthest-pair distances ($\approx$660--680), closely approaching the profile of the ground-truth reference ($\approx$400--460 and $\approx$650--900, respectively). Crucially, the general-purpose encoders show almost no separation between the two curves, with both nearest-neighbor and farthest-pair distances collapsing to the same narrow range ($\approx$500--530), confirming that their embedding spaces are globally insensitive to transcription similarity regardless of neighborhood size. This confirms that the positive global correlations reported in Section~4 reflect genuine local structure in the embedding space: \musvit{} does not only correlates with transcription similarity in aggregate, but also clusters the most similar scores tightly together and pushes the most dissimilar ones apart.

\begin{figure}[ht!]
    \centering
    \includegraphics[width=1.0\linewidth]{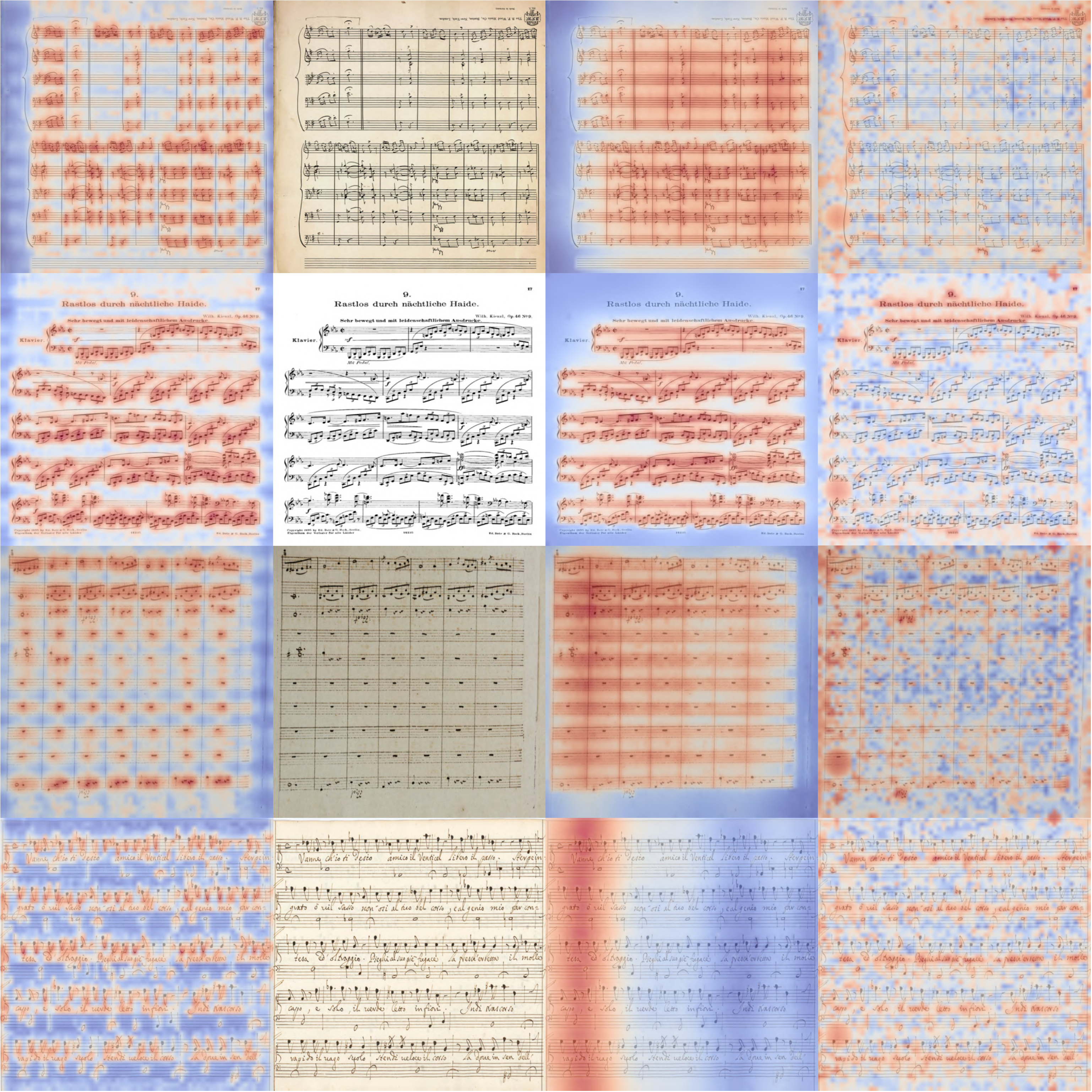}
    \caption{Additional PCA visualizations following the same layout as Fig.~\ref{fig:pca_general}. The pattern is consistent across score pages: \musvit{} activations remain tightly aligned with notation rows regardless of score style or density, while DINOv3-7B and PaliGemma~2 activations remain diffuse and unstructured.}
    \label{fig:pca_general_2}
\end{figure}

%
%
\end{appendix}

\end{document}